\documentclass[10pt,journal,compsoc]{IEEEtran}

\ifCLASSOPTIONcompsoc
  \usepackage[nocompress]{cite}
\else
  \usepackage{cite}
\fi

\ifCLASSINFOpdf
\else
\fi

\usepackage{balance}
\usepackage{amssymb}
\usepackage{cite}
\usepackage{graphicx}
\usepackage{amsmath}
\usepackage{stfloats}
\usepackage{url}
\usepackage{bm}
\usepackage[latin1]{inputenc}
\usepackage{colortbl}
\usepackage{soul}
\usepackage{multirow}
\usepackage{pifont}
\usepackage{color}
\usepackage{alltt}
\usepackage{enumerate}
\usepackage{siunitx}
\usepackage{pbox}
\usepackage{textcomp}
\usepackage{gensymb}
\usepackage{float}
\usepackage{mathtools}
\usepackage{booktabs}
\usepackage{algorithm}
\usepackage{algorithmic}
\usepackage{makecell}
\usepackage[dvipsnames]{xcolor}
\usepackage{bbm}
\usepackage{ragged2e}

\usepackage[normalem]{ulem}
\usepackage{soul}
\usepackage{enumitem}

\usepackage[colorlinks,
            linkcolor=red,
            anchorcolor=blue,
            citecolor=green
            ]{hyperref}

\newcommand{\myunderline}[1]{\underline{\smash{#1}}}

\usepackage{xcolor}
\usepackage[switch]{lineno}

\usepackage{amsthm}

\newtheorem{proposition}{Proposition}

\newtheorem{remark}{Remark}

\begin{document}

%

\title{Global Truncated Loss Minimization for Robust and Threshold-Resilient Geometric Estimation}

\author{Tianyu Huang, Liangzu Peng, Xinyue Zhang, Tongfan Guan, Jinhu Dong, Haoang Li,  \\ Laurent Kneip,~\IEEEmembership{Senior Member,~IEEE}, and Yun-Hui Liu,~\IEEEmembership{Fellow,~IEEE}
\IEEEcompsocitemizethanks{\IEEEcompsocthanksitem T. Huang, T. Guan, J. Dong, and Y.-H. Liu are with the Hong Kong Centre For Logistics Robotics, The Chinese University of Hong Kong.
\IEEEcompsocthanksitem L. Peng is with the Center for Innovation in Data Engineering and Science (IDEAS), University of Pennsylvania.
\IEEEcompsocthanksitem X. Zhang and L. Kneip are with the Mobile Perception Lab of the School of Information and Technology, ShanghaiTech University.
\IEEEcompsocthanksitem H. Li is with the Thrust of Robotics and Autonomous Systems, The Hong Kong University of Science and Technology
(Guangzhou).
\IEEEcompsocthanksitem Yun-Hui Liu is the corresponding author. E-mail: yhliu@cuhk.edu.hk
}
}

\markboth{Journal of \LaTeX\ Class Files,~Vol.~14, No.~8, August~2015}%
{Shell \MakeLowercase{\textit{et al.}}: Bare Advanced Demo of IEEEtran.cls for IEEE Computer Society Journals}

\IEEEtitleabstractindextext{%
\justifying
\begin{abstract}
To achieve outlier-robust geometric estimation, robust objective functions are generally employed to mitigate the influence of outliers.
The widely used consensus maximization~(CM) is highly robust when paired with global branch-and-bound~(BnB) search. 
However, CM relies solely on inlier counts and is sensitive to the inlier threshold. Besides, the discrete nature of CM leads to loose bounds, necessitating extensive BnB iterations and computation cost.
Truncated losses~(TL), another continuous alternative, leverage residual information more effectively and could potentially overcome these issues.
But to our knowledge, no prior work has systematically explored globally minimizing TL with BnB and its potential for enhanced threshold resilience or search efficiency.
In this work, we propose \textsf{GTM}, the first unified BnB-based framework for \myunderline{g}lobally-optimal \myunderline{T}L loss \myunderline{m}inimization across diverse geometric problems. 
GTM involves a hybrid solving design: given an $n$-dimensional problem, it performs BnB search over an $(n-1)$-dimensional subspace while the remaining 1D variable is solved by analytically bounding the objective function. 
Our hybrid design not only reduces the search space, but also enables us to derive Lipschitz-continuous bounding functions that are general, tight, and can be efficiently solved by a classic global Lipschitz solver named DIRECT, which brings further acceleration. 
We conduct a systematic evaluation on the performance of various BnB-based methods for CM and TL on the robust linear regression problem, 
showing that \textsf{GTM} enjoys remarkable threshold resilience and the highest efficiency compared to baseline methods. 
Furthermore, we apply \textsf{GTM} on different geometric estimation problems with diverse residual forms. Both simulated and real-world experiments demonstrate that \textsf{GTM} achieves state-of-the-art outlier-robustness and threshold-resilience while maintaining high efficiency across these estimation tasks.

\end{abstract}

}

\maketitle

\IEEEdisplaynontitleabstractindextext

\IEEEpeerreviewmaketitle

\ifCLASSOPTIONcompsoc
\IEEEraisesectionheading{\section{Introduction}\label{sec:intro}}
\else
\fi

\IEEEPARstart{G}{eometric} estimation is fundamental to numerous computer vision and robotics applications, e.g., scene reconstruction~\cite{Huber-IVC2003, Xu-TRO2022}, robot navigation~\cite{Geiger-IJRR2013, Royer-IJCV2007}, panoramic stitching~\cite{Brown-CVPR2007, Brown-IJCV2007}, etc.
Its core objective is fitting geometric models~(e.g., camera poses) from observed data that are often represented as feature matches~\cite{Antonante-TRO2021, Carlone-FTR2023}. 
Despite the importance, this task is challenging due to the presence of \textit{outlier} data caused by noisy measurement and imperfect feature description.
This challenge motivates the development of outlier-robust estimation techniques, particularly those grounded in robust objective functions.

\begin{figure}[!t]
    \centering
    \footnotesize
    \renewcommand{\tabcolsep}{1.2pt}
    \renewcommand\arraystretch{5}
    \begin{tabular}{cc}
        \makecell{CM} & \makecell{\includegraphics[width=0.43\textwidth]{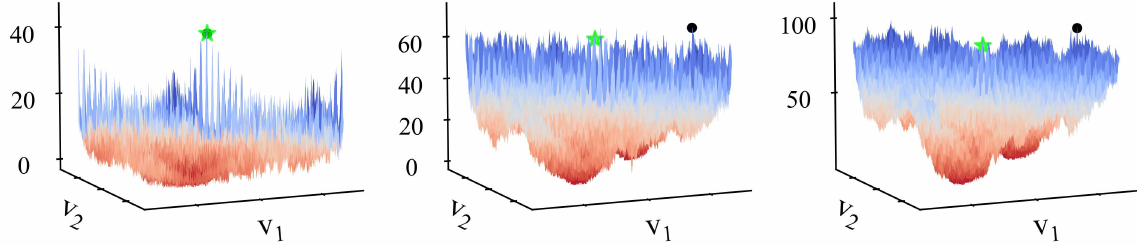}} \\
        \makecell{TL} & \makecell{\includegraphics[width=0.43\textwidth]{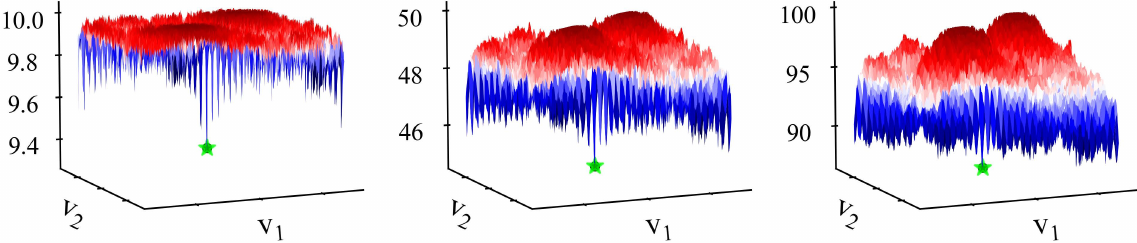}}
    \end{tabular}
    \\[0.3em]
    \makebox[0.15\textwidth]{\footnotesize \quad \quad \quad \quad (a) $\xi$\ =\ 0.02}
    \makebox[0.15\textwidth]{\footnotesize \quad \quad \quad (b) $\xi$\ =\ 0.1}
    \makebox[0.15\textwidth]{\footnotesize \quad \ \ (c) $\xi$\ =\ 0.2}
    \\[-0.8em]
    \caption{Comparison of the CM and TL objective functions in a 2D robust estimation problem~(c.f.~Section~\ref{sec:2dest}) under various thresholds $\xi$. \textcolor{green}{Stars} mark the objective function values of ground truth~(GT) and \textbf{dots} mark the global optimums. As $\xi$ increases, the GT loses its status as the global maximum in CM while maintaining the global minimum in TL.}
    \label{fig:CM_vs_TL}
\end{figure}

One prominent robust objective function is consensus maximization~(CM)~\cite{Fischler-CACM1981, Chin-CVPR2015, Huang-TPAMI2024, Huang-RAL2024}. 
For a specific estimation problem, we often have a residual function $r:\mathcal{C}\times\mathcal{D}\rightarrow[0,\infty)$~(e.g., re-projection error), where $\mathcal{C}\subset\mathbb{R}^n$ defines the feasible region of the geometric variable $\mathbf{v}$ and $\mathcal{D}$ is the data domain. Given measured data $\{\mathbf{d}_i\}_{i=1}^{M} \subset \mathcal{D}$ and a predefined threshold $\xi$, CM aims to find a $\mathbf{v} \in \mathcal{C}$ that maximizes the number of \textit{inlier} data $\mathbf{d}_i$ satisfying $r(\mathbf{v}, \mathbf{d}_i) \leq \xi$, i.e.,
\begin{align}\label{eq:CM}
    \max_{\mathbf{v}\in\mathcal{C}} \sum_{i = 1}^{M} \bm{1}(r(\mathbf{v}, \mathbf{d}_i) \leq \xi), \tag{\textcolor{red}{CM}}
\end{align}
where $\bm{1}(\cdot)$ outputs $1$ if the input statement is true, or otherwise outputs $0$. 
Despite its nonconvexity, \eqref{eq:CM} has been addressed by various powerful strategies~\cite{Torr-CVIU2000, Barath-TPAMI2021a, Hartley-IJCV2009}, particularly the \textit{branch-and-bound}~(BnB) method that can provide globally optimal solutions through systematic search~\cite{Huang-RAL2024, Huang-TPAMI2024, Campbell-ICCV2017, Bustos-TPAMI2017}. 
However, the CM objective in \eqref{eq:CM} suffers from inherent limitations. First, it 
relies solely on inlier counts and is sensitive to threshold $\xi$~(see Fig.~\ref{fig:CM_vs_TL}). In practice, selecting a suitable threshold is an art and one cannot choose it effectively without any prior knowledge~\cite{Heinrich-IVC2013, Edstedt-ARXIV2025}. Second, the discrete nature of \eqref{eq:CM} complicates the development of tight bounds for BnB, often leading to extensive iterations for convergence~\cite{Huang-CVPR2024, Zhang-TPAMI2024}.

On the other hand, the truncated losses~(TL), another classical robust objective function~\cite{Huang-CVPR2024, Huber-Book1992, Black-IJCV1996, Antonante-TRO2021}, not only mitigates the influence of outliers with residuals higher than a threshold as in \eqref{eq:CM}, but also goes further to minimize the residuals below this threshold:
\begin{equation}\label{eq:TL}
    \min_{\mathbf{v}\in\mathcal{C}} \sum_{i = 1}^{M} \min \{r(\mathbf{v}, \mathbf{d}_i),\ \xi \}. \tag{\textcolor{red}{TL}}
\end{equation}
Would the continuous objective in \eqref{eq:TL} overcome the above limitations of \eqref{eq:CM}, particularly the sensitivity to misspecified thresholds? 
Since, to the best of our knowledge, no prior work proves this intuition, in Fig.~\ref{fig:CM_vs_TL} we conduct a simple experiment to verify it and observe: 
as $\xi$ increases, the ground truth remains the global optimizer of \eqref{eq:TL} but no longer maximizes \eqref{eq:CM}, highlighting the superior threshold-resilience of \eqref{eq:TL}. 
Having verified the advantage of \eqref{eq:TL}, the next critical question arises: Can we solve \eqref{eq:TL} to global optimality? In contrast to the established success of solving \eqref{eq:CM} via global optimization techniques like BnB, TL has received significantly less exploration in this regard. Indeed, most existing solvers do not guarantee solving \eqref{eq:TL} global optimally~\cite{Black-IJCV1996, Peng-CVPR2023, Antonante-TRO2021, Yang-TPAMI2022}. To the best of our knowledge, the only attempt to solve \eqref{eq:TL} optimally is \cite{Huang-CVPR2024}, but the formulation and BnB algorithm of \cite{Huang-CVPR2024} are tailored to a specific geometric problem with a specific residual function (in the context of \textit{point cloud registration}), thus it is not applicable to \eqref{eq:TL} with diverse residuals.

In this work, we introduce the first unified BnB-based framework designed to solve \eqref{eq:TL} across a diverse range of geometric tasks characterized by various residual functions, named \myunderline{G}lobally-optimal \myunderline{T}L loss \myunderline{M}inimization~(\textsf{GTM}). 
At the heart of \textsf{GTM} is a strategic blend of two powerful and complementary global optimization mechanisms.
Instead of conducting a conventional BnB search over the full $n$-dimensional variable space, \textsf{GTM} confines the \textit{BnB procedure to an $(n-1)$-dimensional subspace}. Concurrently, the remaining 1-dimensional variable is addressed using \textit{the DIRECT method}~\cite{Jones-JOTA1993,Jones-JOP2021}, a derivative-free optimization strategy renowned for its efficiency in optimizing Lipschitz continuous functions. The rationale of employing the ($n-1$)-dimensional BnB is that it reduces the search dimension by one, leading to appreciable acceleration as evidenced in different contexts from prior studies \cite{Bustos-ICCV2015,Zhang-TPAMI2024,Huang-CVPR2024}. The rationale of employing the DIRECT method is that it perfectly addresses two fundamental challenges that arise from ($n-1$)-dimensional BnB: (1) the need to derive general bounding functions for diverse residuals; (2) globally optimizing the bounding functions efficiently. 

To deal with the first challenge, we leverage \textit{interval arithmetic}~\cite{Hansen-BOOK2003, Moore-SIAM2009} to derive innovative general bounding functions that require solving 1-dimensional optimization problems~(detailed in Sections~\ref{subsection:GTM-UB} and \ref{subsection:GTM-LB}). These problems are carefully formulated so that their objectives are provably \textit{Lipschitz continuous} (see, e.g., Proposition~\ref{prop:Lips_inher}). This is then why the DIRECT method comes into play in addressing the second challenge: It globally optimizes low-dimensional Lipschitz continuous functions. Moreover, the DIRECT method is particularly efficient for univariate problems, and its global optimality ensures tight bounds and therefore convergence in much fewer BnB iterations, which leads to the high efficiency of \textsf{GTM}. Overall, combining the two general-purpose schemes, with the DIRECT method embedded as an inner solver within the BnB framework, allows \textsf{GTM} to be applied for various geometric problems with guaranteed global optimality and improved efficiency.

To verify the effectiveness of \textsf{GTM}, extensive experiments are conducted. We perform a systematic evaluation on the performance of various BnB-based methods for solving \eqref{eq:CM} and \eqref{eq:TL} through the robust linear regression problem~(see Section~\ref{subsection:compare}). The comparison results show that 
\textsf{GTM} is much faster than all other methods~(roughly a $3\times$-$45\times$ speed-up) and achieves significantly higher threshold-resilience with fewer iteration numbers than all \eqref{eq:CM}-based methods. 
Furthermore, we apply \textsf{GTM} on three geometric estimation problems: relative pose estimation under planar motion, point cloud registration, and rotational homography estimation with unknown focal length~(see Sections \ref{sec:2dest}, \ref{sec:3dest}, and \ref{sec:4dest}). Abundant experiments show that \textsf{GTM} achieves the state-of-the-art outlier-robustness and threshold-resilience while holding high efficiency across all these applications.
We also evaluate standalone DIRECT on all above robust estimation problems since it can apply to higher dimensions.
Crucially, while 1D DIRECT within \textsf{GTM} is highly efficient, standalone DIRECT in dimensions $\geq$ 2 is as slow as vanilla BnB and can fail if the objective lacks Lipschitz continuity.
This highlights the limitations of DIRECT in high dimensions and validates the superiority of \textsf{GTM}'s hybrid design.

Overall, our main contributions are as follows:
\begin{itemize}
    \item We propose \textsf{GTM}, the first unified BnB framework for truncated loss minimization, which guarantees global optimality and applies to various geometric tasks. 
    \item We develop an innovative hybrid solving strategy in \textsf{GTM}, which involves $(n-1)$-dimensional BnB search and, more importantly, 1-dimensional Lipschitz continuous bounding functions that are general, tight, and can be efficiently solved by global Lipschitz optimization. 
    \item We apply \textsf{GTM} on different outlier-robust estimation problems with diverse residual functions. In addition, we conduct a systematic evaluation on the performance of various BnB-based methods for solving \eqref{eq:CM} and \eqref{eq:TL}. Abundant experimental results demonstrate the remarkable threshold-resilience of \textsf{GTM} and its higher efficiency compared to other BnB-based methods.
\end{itemize}

The rest of this paper is organized as follows. Section~\ref{sec:related} reviews the related robust estimation methods based on various  robust objective functions. 
Section~\ref{sec:pre} provides  necessary preliminaries of the BnB framework.
Section~\ref{sec:GTM} introduces the proposed \textsf{GTM} framework and the developed bounding functions, as well as a systematic evaluation on the robust linear regression problem. 
Next, Sections~\ref{sec:2dest}, \ref{sec:3dest}, and \ref{sec:4dest} present the applications of \textsf{GTM} on three geometric estimation problems with extensive evaluation experiments. 
Finally, the paper is concluded in Section~\ref{sec:conc}.

\section{Related Works}
\label{sec:related}
In this section, we give a brief review of robust estimation methods relying on different robust objective functions.

The \textit{Consensus Maximization} objective in~\eqref{eq:CM} has become a popular paradigm for robust geometry estimation, with many excellent solving algorithms being developed~\cite{Fischler-CACM1981, Chum-JPRS2003, Torr-CVIU2000, Barath-TPAMI2021b, Li-ICCV2009, Bustos-TPAMI2017}. 
Among these methods, the heuristic RANSAC method~\cite{Fischler-CACM1981}, thanks to its simplicity, remains a commonly used strategy in the past decades. 
RANSAC iteratively samples minimal sets of data points for candidate geometric models, and outputs the final solution that defines the largest consensus set found. While its performance has been continuously improved by new RANSAC variants~\cite{Barath-TPAMI2021a, Torr-CVIU2000, Barath-TPAMI2021b, Chum-JPRS2003}, they still lack robustness at high outlier ratios~(e.g.~$> 90\%$), due to the uncertainty of heuristic sampling~\cite{Huang-CVPR2024, Chin-CVPR2015}.
To achieve accurate estimation against huge outlier ratios, the parameter search-based methods gradually become better choices~\cite{Li-ICCV2007, Li-ICCV2009, Zheng-CVPR2011}. One representative approach is the branch-and-bound (BnB) strategy, which recursively searches the parameter space and outputs a globally optimal solution of \eqref{eq:CM}. BnB has led to impressive estimation robustness and has been widely employed in various geometric problems~\cite{Yang-ECCV2014, Campbell-ICCV2017, Campbell-CVPR2019, Peng-TPAMI2021}. However, the efficiency of standard BnB is severely affected by the dimension of the search space and tightness of the bounding functions. In recent years, plenty of methods are proposed to conduct problem decomposition by leveraging geometric constraints of specific problems~\cite{Yang-TPAMI2015, Liu-ECCV2018, Huang-TPAMI2024, Li-ECCV2020}. This decomposition usually leads to sub-problems with a dimension $\leq 3$ from the original high-dimensional problems~($\geq 6$), therefore achieving a great speedup. Moreover, some researchers propose to search the $(n-1)$-dimensional space and construct 1D bounding functions with respect to the remaining variable, which also brings a remarkable computational gain~\cite{Huang-TPAMI2024, Zhang-TPAMI2024, Bustos-ICCV2015}. In these prior works, the 1D bounding functions are formulated based on \eqref{eq:CM} and solved efficiently via \textit{interval stabbing}~\cite{Bustos-ICCV2015}. However, due to the discrete nature of \eqref{eq:CM} and interval stabbing, these works are not applicable to continuous losses such as \eqref{eq:TL}.

In contrast to the discrete objective function in \eqref{eq:CM}, there is another category of methods belonging to \textit{M-estimation}, which relates to continuous robust functions of the residuals~\cite{Huber-Book2011, Geer-Book2000, Barron-CVPR2019}.
Among these methods, a particular family of losses, which we focus on in the paper, are the \textit{Truncated Losses}, c.f.~\eqref{eq:TL}. According to the norm order of the inner residual function, there are truncate absolute residuals~($L_1$-norm), truncated least squares~($L_{2}$-norm), etc. 
To solve such kind of problems, traditional solvers involve either adaptive annealing~\cite{Sidhartha-CVPR2023, Peng-CVPR2023}, or continuation schemes such as graduated-non-convexity~\cite{Le-3DV2021, Black-IJCV1996}. While they are generally efficient and can deal with low outlier ratios, such local methods could get stuck in local optima once the initial solution is far from the ground truth due to huge outlier ratios.
In recent years, a plenty of works consider to relax the non-convex truncated least squares into a convex semidefinite program~\cite{Yang-TPAMI2022, Peng-ECCV2022, Barfoot-IJRR2025, Liao-CVPR2025}. 
However, to recover the solution of the original objective, such semidefinite relaxation could introduce quadratically many variables and constraints, leading to high computational costs~(e.g., over 7 hours to deal with 1,000 data samples in a 3-dimensional problem~\cite[Table~3]{Yang-MP2023}). Moreover, the recent work \cite{Huang-CVPR2024} employs BnB to solve the truncated entry-wise absolute residuals, where the bounding functions are designed to be tight and computationally efficient. However, this method is only applicable to a special residual formulation with truncated $L_1$-loss and lacks generality.
In contrast, in this paper, we aim to develop a unified framework that can employ BnB search on various estimation problems with diverse residual formulations.

Apart from the truncated losses that we focus on in this paper, it is worth mentioning some other popular robust objective functions belonging to the M-estimation. The classical \textit{Huber loss}~\cite{Huber-Book1992, Meyer-CVPR2021}, which combines quadratic behavior near the origin with linear tails beyond a predefined threshold, inspires a series of subsequent robust M-estimation losses, including the truncated losses. For example, the \textit{Geman-McClure loss} applies stronger suppression by saturating gradients for large residuals to improve the Huber loss~\cite{Geman-Book1986, Chen-RAL2024}.
Moreover, the \textit{Cauchy loss}~\cite{Black-CVIU1996} leverages heavy-tailed distributions to infinitely down-weight extreme errors. 
Recently, an adaptive robust loss is developed in \cite{Barron-CVPR2019} to interpolate between several classical losses, enabling learnable robustness but also leading to more complicated optimization.

\section{Preliminaries}
\label{sec:pre}
This section presents the preliminaries that form the foundation of our \textsf{GTM} framework: first, the general branch-and-bound algorithm, and second, the interval analysis basics that are essential for computing the bounds in BnB.

\subsection{Branch-and-bound}
\label{sec:pre_bnb}

Branch-and-bound~(BnB)  is a search-based solving strategy for global optimization of nonconvex problems~\cite{Lawler-OR1966, Moore-CMA1991}. 
Based on recursive search and given a prescribed error tolerance $\epsilon$, BnB guarantees finding an $\epsilon$-optimal solution, which is approximately a global solution if $\epsilon$ is sufficiently small~\cite{Boyd-Lecture2007, Olsson-TPAMI2008}.
This global optimality guarantee makes BnB a remarkable tool for solving nonconvex geometric estimation problems with increased robustness to outliers. In this section, we review its basic philosophy and identify its weaknesses for robust geometric estimation.

\noindent \textbf{Basics of BnB.} Assume we want to minimize some nonconvex function $f:\mathcal{C}\rightarrow\mathbb{R}$ over the constraint set $\mathcal{C} \subset \mathbb{R}^n$. Note that $f$ can be replaced by $-f$ if maximization is demanded. At the heart of BnB are two bounding functions $U(\cdot)$ and $L(\cdot)$, which take a subset $\mathbb{B}$ of $\mathcal{C}$ as input and output respectively an \textit{upper and lower bound} of the minimum of $f$. More precisely, we define $U(\mathbb{B})$ as an objective value $f(\mathbf{v})$ with some $\mathbf{v}\in \mathbb{B}$ to serve as an upper bound of the minimum of $f$ on the constraint set $\mathcal{C}$; we require $L(\cdot)$ to return a value, written as $L(\mathbb{B})$, which is smaller than the minimum of $f$ on the subset $\mathbb{B}$, meaning that 
\begin{equation}\label{eq:BnB_LB}
    L(\mathbb{B}) \leq \min_{\mathbf{v}\in\mathbb{B}} f(\mathbf{v}).
\end{equation}
Now, consider two subsets $\mathbb{B}_1,\mathbb{B}_2$ on which global minimizers are to be searched. Suppose the lower bound $L(\mathbb{B}_1)$ is larger than or equal to $U(\mathbb{B}_2)$. Then by definition \eqref{eq:BnB_LB}, the minimum objective values on $\mathbb{B}_1$ will not be smaller than $U(\mathbb{B}_2)$, so we can safely discard $\mathbb{B}_1$ and only search on $\mathbb{B}_2$. With the bounding functions and initial branch $\mathcal{C}$, BnB recursively divides $\mathcal{C}$ into smaller sub-branches while discarding sub-branches that cannot yield a better solution. 

We refer the readers to the appendix for the detailed BnB algorithm, while here we focus on the computation of bounds $U(\mathbb{B})$ and $L(\mathbb{B})$. For the upper bound, we could in principle set $U(\mathbb{B})\gets f(\mathbf{v})$ for an arbitrary feasible point $\mathbf{v}\in \mathbb{B}$, while in practice, $\mathbb{B}$ is often some axis-aligned rectangle, and a common choice of $\mathbf{v}$ is the center of $\mathbb{B}$. For the lower bound, the ideal case would be that we solve the minimization problem in \eqref{eq:BnB_LB} and set $L(\mathbb{B})$ to the minimum. However, this problem is often as hard as minimizing $f$ on the original constraint set $\mathcal{C}$. As a compromise, it is common to relax the original non-convex objective $f(\cdot)$, replacing it with some \textit{underestimator} $\myunderline{f}(\cdot)$. Such $\myunderline{f}(\cdot)$ satisfies that $\myunderline{f}(\mathbf{v})\leq f(\mathbf{v})$ for every $\mathbf{v}\in \mathbb{B}$ and furthermore that it has an easily computable minimum over $\mathbb{B}$. In this case, the lower bound is obtained via
\begin{equation}
    L(\mathbb{B}) \gets \min_{\mathbf{v}\in\mathbb{B}} \myunderline{f}(\mathbf{v}).
\end{equation}
There are many methods developed to find and optimize such an underestimator $\myunderline{f}$. For example, one could consider relaxing the above optimization problem into a convex program (e.g., \cite{Olsson-TPAMI2008, Zheng-CVPR2011}). In summary, the bounds $U(\mathbb{B})$ and $L(\mathbb{B})$ are generally computable.

\noindent \textbf{Weakness of Vanilla BnB.} Despite its robustness and popularity, BnB, when applied in its vanilla form, has two limitations. First, it can be slow for BnB to search in high dimensions. This issue is to some extent alleviated in outlier-robust geometric estimation, partly because the search space dimension is often small (e.g., optimization variables often have fewer than 6 degrees of freedom~(DoF)), and partly because the dimension could be furthermore reduced via DoF decomposition or relaxation based on geometric constraints~\cite{Huang-TPAMI2024, Liu-ECCV2018, Yang-TRO2020}. The other limitation is the difficulty of deriving tight bounds for losses like \eqref{eq:CM} and \eqref{eq:TL}; indeed, tighter bounds would allow BnB to discard more sub-branches and avoid unnecessary exploration. However, the aforementioned lower bounding strategies (e.g., convex relaxation) often yield imprecise bounds. In fact, the difficulty in deriving tighter lower bounds has been a major bottleneck that hinders the efficiency of BnB for outlier-robust geometric estimation.

\subsection{Interval Analysis}
\label{sec:pre_IA}

At each BnB iteration, we assume the sub-branch $\mathbb{B} \subset \mathbb{R}^n$ is a $n$-dimensional axis-aligned rectangle of the form
\begin{equation}
    \mathbb{B} = [b_1^l, b_1^u] \times [b_2^l, b_2^u] \times \dots \times [b_n^l, b_n^u],
\end{equation}
where each $[b_j^l, b_j^u]$ is the feasible interval of $j^{th}$ element in the variable $\mathbf{v}$. 
This assumption holds for many applications and is thus without loss of generality. 
For example, a 3D rotation is parameterized by three angles lying respectively in $[-\pi,\pi]$, and a 3D translation variable could have its norm well bounded above by some large enough number. The interval form offers an opportunity to perform mathematical computations with the intervals rather than a specific number contained in the interval, via a technique called \textit{interval analysis}~\cite{Hansen-BOOK2003, Moore-SIAM2009}. Our work needs two concepts in interval analysis, \textit{interval mapping} and \textit{interval arithmetic}.

\noindent \textbf{Interval Mapping.} 
Given a function $\mathcal{T}: \mathbb{R}^m \rightarrow \mathbb{R}$ and a m-dimensional interval $\mathbf{X} = [x_1^l, x_1^u] \times \dots \times [x_m^l, x_m^u] \subset \mathbb{R}^m$, the interval mapping of $\mathcal{T}$, denoted $\mathcal{T}(\mathbf{X})$, is defined as:
\begin{equation}\label{eq:interval_map}
    \mathcal{T}(\mathbf{X}) = [\min_{\mathbf{x}\in\mathbf{X}} \mathcal{T}(\mathbf{x}),\ \max_{\mathbf{x}\in\mathbf{X}} \mathcal{T}(\mathbf{x})],
\end{equation}
ensuring that $\mathcal{T}(\mathbf{X})$ contains all possible values of $\mathcal{T}(\mathbf{x})$.

\noindent \textbf{Interval Arithmetic.}
When the interval mapping consists of basic arithmetic, its range in \eqref{eq:interval_map} can be easily evaluated. For example, given intervals $X = [x^l, x^u]$ and $Y = [y^l, y^u]$, four basic interval arithmetic operations are defined as:
\begin{align}
    X + Y &= [x^l + y^l,\ x^u + y^u], \label{eq:arith-add}\\
    X - Y &= [x^l - y^u,\ x^u - y^l], \\ 
    X \cdot Y &= [\min\{x^ly^l, x^ly^u, x^uy^l, x^uy^l\}, \notag\\ 
    & \quad \quad  \quad \max\{x^ly^l, x^ly^u, x^uy^l, x^uy^l\}], \label{eq:interval_prod}\\
    X/Y &= [x^l, x^u] \cdot [1/y^u, 1/y^l] \quad \textnormal{if}\ y^l > 0.
\end{align}
With interval analysis, we can propagate the intervals of the variables to a conservative range estimate. This will greatly facilitate our development on BnB bounds.

\section{GTM: Globally-optimal TL Minimization}
\label{sec:GTM}
In this section, we introduce  \textsf{GTM} (Globally-optimal TL Minimization). It employs BnB to minimize \eqref{eq:TL} in a globally optimal manner~(Algorithm~\ref{alg:GTM}). As mentioned, the efficiency of BnB relies on a reduced search space and tight bounds. In light of this, we first describe a general technique to reduce the search space dimension by 1. Recall that in \eqref{eq:TL}, we aim to find some $\mathbf{v}= [v_1, v_2, \dots, v_n]^{\top}$ in the $n$-dimensional constrained space $\mathcal{C} \subset \mathbb{R}^n$ that globally minimizes our objective $f(\mathbf{v})$. Here, $\mathcal{C}$ is a $n$-dimensional axis-aligned rectangle, that is $\mathcal{C} = [c_1^l, c_1^u] \times [c_2^l, c_2^u] \times \dots \times [c_n^l, c_n^u]$. A key choice of \textsf{GTM} is to systematically search, in a BnB manner, the ($n-1$)-dimensional rectangle $\mathcal{C}_{2:n}:=[c_2^l, c_2^u] \times \dots \times [c_n^l, c_n^u]$, rather than the entire space $\mathcal{C}$. 

But searching over $\mathcal{C}_{2:n}$ instead of $\mathcal{C}$  brings an obstacle: With the remaining one-dimensional interval $\mathcal{C}_1:=[c_1^l, c_1^u]$ fully unexplored, how do we compute upper and lower bounds? For the case of solving \eqref{eq:CM}, this obstacle has been addressed by \cite{Zhang-TPAMI2024}. In \cite{Zhang-TPAMI2024}, the interval $\mathcal{C}_1$ unexplored by BnB is efficiently handled by a different procedure known as \textit{interval stabbing}, which furthermore computes the upper and lower bounds at the same time. Nevertheless, their bound computation is limited to \eqref{eq:CM} and does not generalize to other robust losses such as \eqref{eq:TL}. Since Fig.~\ref{fig:CM_vs_TL} implies \eqref{eq:TL} is more robust to misspecified thresholds than \eqref{eq:CM}, we are motivated to develop novel upper and lower bounding strategies for \eqref{eq:TL} with diverse residual functions, even in the absence of BnB-related information about $\mathcal{C}_1$. We detail our mathematical development of the bounds subsequently in Sections \ref{subsection:GTM-UB} and \ref{subsection:GTM-LB}, followed by Section~\ref{sec:direct} where we introduce a global Lipschitz optimization method for efficiently solving these bounds to global optimality.

\begin{algorithm}[!t]
\footnotesize
\caption{\textsf{GTM} for Globally Minimizing \eqref{eq:TL}.}
\label{alg:GTM}
\textbf{Input:} Objective function $f(v_1, \mathbf{v}_{2:n})$, prescribed error $\epsilon$;

\textbf{Output:} A global minimizer $\{\hat{\mathbf{v}}_{2:n},\ \hat{v}_1 \}$;

    \begin{algorithmic}[1]
        \STATE $\mathcal{Q} \gets $ An empty priority queue; $\hat{\mathbf{v}}_{2:n} \gets $ Center of $\mathcal{C}_{2:n}$;
        \STATE $\hat{U},\ \hat{v}_1 \gets$ Solve \eqref{eq:GTM_UB} with $\mathbf{v}_{2:n} = \hat{\mathbf{v}}_{2:n}$  (Algorithm~\ref{alg:DIRECT});
        \STATE $L(\mathcal{C}_{2:n}) \gets $ Solve \eqref{eq:GTM_LB} with $\mathbf{v}_{2:n} \in \mathcal{C}_{2:n}$  (Algorithm~\ref{alg:DIRECT});
        \STATE Insert $\mathcal{C}_{2:n}$ with priority $L(\mathcal{C}_{2:n})$ into $\mathcal{Q}$;
        \WHILE{$\mathcal{Q}$ is not empty}
            \STATE Pop a branch $\mathbb{B}_{2:n}$ with the lowest $L(\mathbb{B}_{2:n})$ from $\mathcal{Q}$;
            \IF{$\hat{U}$ - $L(\mathbb{B}_{2:n}) < \epsilon$}
            \STATE \textbf{end while}
            \ENDIF
            \STATE Divide $\mathbb{B}_{2:n}$ into $2^{n-1}$ sub-branches $\{\mathbb{B}_j\}_{j=1}^{2^{n-1}}$.
            \FOR{each sub-branch $\mathbb{B}_j$}
		  \STATE $L(\mathbb{B}_j) \gets$ Solve \eqref{eq:GTM_LB} with $\mathbf{v}_{2:n}\in\mathbb{B}_j$ by Algorithm~\ref{alg:DIRECT};
            \IF{$L(\mathbb{B}_j) > U^*$}
            \STATE \textbf{continue}
            \ELSE
            \STATE $\dot{\mathbf{v}}_{2:n}^j \gets $ Center of $\mathbb{B}_j$;
            \STATE $U(\mathbb{B}_j),\ v_{1j} \gets$ Solve \eqref{eq:GTM_UB} with $\mathbf{v}_{2:n} = \dot{\mathbf{v}}_{2:n}^j$ (Algorithm~\ref{alg:DIRECT});
            \IF{$U(\mathbb{B}_j) < \hat{U}$}
            \STATE $\hat{\mathbf{v}}_{2:n} \gets \dot{\mathbf{v}}_{2:n}^j$; $\hat{v}_1 \gets v_{1j}$, \ $\hat{U} \gets U(\mathbb{B}_j)$;
            \ENDIF
            \STATE Insert $\mathbb{B}_j$ with priority $L(\mathbb{B}_j)$ into $\mathcal{Q}$;
            \ENDIF
 		\ENDFOR 
        \ENDWHILE
    \end{algorithmic}
\end{algorithm}

\subsection{\textsf{GTM} - Upper Bound}\label{subsection:GTM-UB}
As mentioned, \textsf{GTM} performs a BnB search over the axis-aligned rectangle $\mathcal{C}_{2:n}$ of $\mathbb{R}^{n-1}$ for an $n-1$-dimensional vector $\mathbf{v}_{2:n}:=[v_2,\dots,v_n]^\top$. To derive bounds for this scheme, we first separate the variables $v_1$ and $\mathbf{v}_{2:n}$ in our notation. Write the residual function  $r(\mathbf{v}, \mathbf{d}_i)$ in \eqref{eq:TL} as $r_i(v_1, \mathbf{v}_{2:n})$; note here we suppress the presence of $\mathbf{d}_i$ and work with the more concise notation $r_i(\cdot)$. Define $f_i(v_1, \mathbf{v}_{2:n})$ to be the composition of the residual and the truncated loss, that is
\begin{align}
    f_i(v_1, \mathbf{v}_{2:n}) := \min \{ r_i(v_1, \mathbf{v}_{2:n}),\ \xi \},
\end{align}
and then our objective function $f(\mathbf{v})$ is written as
\begin{equation}
    f(v_1, \mathbf{v}_{2:n}) =  \sum_{i = 1} ^ M f_i(v_1, \mathbf{v}_{2:n}).
\label{eq:re_obj}
\end{equation}
By an upper bound for \textsf{GTM}, we mean the upper bound of the minimum value of the objective function $f(v_1, \mathbf{v}_{2:n})$ over a given feasible sub-branch $\mathbb{B}_{2:n} = [b_2^l, b_2^u] \times \dots \times [b_n^l, b_n^u] \subseteq \mathcal{C}_{2:n}$ of $\mathbf{v}_{2:n}$ and the feasible interval $\mathcal{C}_1$ of $v_1$. Similarly to Section~\ref{sec:pre_bnb}, a valid upper bound would be the function value $f(\dot{v}_1, \dot{\mathbf{v}}_{2:n} )$, where $\dot{v}_1$ and $\dot{\mathbf{v}}_{2:n}$ are respectively the centers of $\mathcal{C}_1$ and $\mathbb{B}_{2:n}$, namely
\begin{align}
    \dot{v}_1 = \frac{c_1^l+c_1^u}{2}, \quad \dot{\mathbf{v}}_{2:n} = [\frac{b_2^l + b_2^u}{2}, \dots, \frac{b_n^l + b_n^u}{2}]^{\top}.
\end{align}
However, this bound is loose for the following reason. As BnB explores, the sub-branch $\mathbb{B}_{2:n}$ shrinks in size, but $\mathcal{C}_1$ does not change and the center $\dot{v}_1$ remains the same. If all global minimizers have their first entry different from $\dot{v}_1$, which is highly likely, we will never find a global solution by this bounding strategy. Therefore, while BnB does not, our upper bounding scheme needs to search in this interval $\mathcal{C}_1$ of possibilities to localize global minimizers. To achieve so, \textsf{GTM} sets its upper bound by optimizing over $\mathcal{C}_1$: 
\begin{equation}\label{eq:GTM_UB}
    U_{\textsf{GTM}} =  \min_{v_1 \in \mathcal{C}_1} f(v_1, \dot{\mathbf{v}}_{2:n}) 
    =  \min_{v_1 \in \mathcal{C}_1}  \sum_{i = 1} ^ M f_i(v_1, \dot{\mathbf{v}}_{2:n}).
\end{equation}
Clearly, we always have $U_{\textsf{GTM}} \leq f(\dot{v}_1, \dot{\mathbf{v}}_{2:n} )$, which means that the upper bound of \textsf{GTM} is always tighter.

It remains to efficiently solve \eqref{eq:GTM_UB} and actually compute $U_{\textsf{GTM}}$. While problem \eqref{eq:GTM_UB} has only a single variable $v_1$, finding its minimum $U_{\textsf{GTM}}$ is not easy, as the objective consists of diverse residual function $r(\cdot)$ composed with the truncated loss, which is non-smooth and non-convex. We postpone the computational procedure of solving \eqref{eq:GTM_UB} to Section~\ref{sec:direct}, as our lower bound computation, which we detail in Section~\ref{subsection:GTM-LB}, will need to solve a similar program.

\subsection{\textsf{GTM} - Lower Bound}\label{subsection:GTM-LB}
Similarly to BnB (Section~\ref{sec:pre_bnb}), an ideal lower bound for \textsf{GTM} would be the minimum of the program
\begin{align}\label{eq:GTM-ideal-LB}
    \min_{v_1\in \mathcal{C}_1, \mathbf{v}_{2:n}\in \mathbb{B}_{2:n}} f(v_1, \mathbf{v}_{2:n}),
\end{align}
where $f$ is defined as in \eqref{eq:re_obj}. Since this program is as hard as the original problem \eqref{eq:TL}, we will find an underestimator $\myunderline{f}$ of $f$ and set the \textsf{GTM} lower bound to the minimum of $\myunderline{f}$. However, the choice of  $\myunderline{f}$ and the lower bound derivation is not as straightforward as deriving an upper bound. Therefore, for clarity, in Section~\ref{subsubsection:lb-regression} we exemplify the derivation for \textit{robust linear regression} where the residual is simply the absolute value of a linear function. This builds intuition for and paves the way to Section~\ref{subsubsection:lb-general}, where we derive lower bounds for \textsf{GTM} with diverse residual functions.

\subsubsection{Lower Bounds for Robust Linear Regression}\label{subsubsection:lb-regression}

In robust linear regression, the $i$-th sample is of the form $(\mathbf{a}_i,y_i)$, where $\mathbf{a}_i=[a_{i,1},\dots,a_{i,n}]^\top$ is a $n$-dimensional vector and $y_i$ a scalar. The residual $r_i(\mathbf{v})$ is of the form
\begin{align}\label{eq:regression-loss}
    r_i(\mathbf{v}) = |\mathbf{v}^\top \mathbf{a}_i - y_i| = | v_1 a_{i,1} + \mathbf{v}_{2:n}^\top \mathbf{a}_{i,2:n}  - y_i|.
\end{align}
Here we recall that $\mathbf{v}_{2:n}$ and $\mathbf{a}_{i, 2:n}$ are ($n-1$)-dimensional vectors consisting of last $n-1$ entries of $\mathbf{v}$ and $\mathbf{a}_i$, respectively. To obtain a lower bound of \eqref{eq:GTM-ideal-LB}, we recall that $f$ is a sum of truncated residuals $\sum_{i}^M f_i$, where each $f_i$ is defined as $f_i(v_1, \mathbf{v}_{2:n}) = \min\{r_i(v_1, \mathbf{v}_{2:n}), \xi\}$, thus it suffices to find an underestimator $\myunderline{f}_i$ for each $f_i$. Indeed, by summing over  $\myunderline{f}_i$  we obtain that $\myunderline{f} = \sum_{i=1}^M \myunderline{f}_i$ is an underestimator of $f$.

We proceed to construct $\myunderline{f}_i$ through the following steps:
\begin{enumerate}[label=(\alph*)]
    \item Write $r_i(\mathbf{v})=|h_i(v_1)+ g_i(\mathbf{v}_{2:n})|$, where $h_i,g_i$ are defined as $h_i(v_1):= v_1 a_{i,1}-y_i$ and $g_i(\mathbf{v}_{2:n}) := \mathbf{v}_{2:n}^\top \mathbf{a}_{i,2:n}$. Compute the maximum and minimum of $g_i(\mathbf{v}_{2:n})$ over the sub-branch $\mathbb{B}_{2:n}$ for every $i=1,\dots, M$, that is 
    \begin{align}
        s_i^l=\min_{\mathbf{v}_{2:n}\in \mathbb{B}_{2:n}} g_i(\mathbf{v}_{2:n}),\ \  s_i^u = \max_{\mathbf{v}_{2:n}\in \mathbb{B}_{2:n}} g_i(\mathbf{v}_{2:n}).
    \end{align}
    Since $g$ here is the linear function $\mathbf{v}_{2:n}^\top \mathbf{a}_{i,2:n}$, it attains its optimum when every  variable $v_i$ attains either endpoint of the interval $[b_i^l, b_i^u]$. Exhaustively searching for such an optimum takes linear time in $n$, and doing so for every sample entails linear dependency in $M$. Thus, computing $s_i^l,s_i^u$ for every $i=1,\dots,M$ takes $O(nM)$ time. In all cases we consider, $n$ is small, e.g., $n<6$. 
    \begin{remark}\label{remark:interval-linear-reg}
        An alternative way to compute the bounds $s_i^l,s_i^u$ is through the lens of interval arithmetic. For example, since we know $v_{j}$ lies in an interval for every $j=2,\dots,n$, so does the product $v_j\cdot a_{i,j}$ (see the product rule \eqref{eq:interval_prod}), and so does the inner product $g_i(\mathbf{v}_{2:n}) = \mathbf{v}_{2:n}^\top \mathbf{a}_{i,2:n}$ (see the summation rule \eqref{eq:arith-add}). One verifies that the interval that $g_i(\mathbf{v}_{2:n})$ lies in, calculated in this way, is precisely $[s_i^l, s_i^u]$. 
    \end{remark}
    \item With the bounds $s_i^l,s_i^u$, we now see that the inner product $\mathbf{v}^\top \mathbf{a}_i$ lies in $[v_1 a_{i,1} + s_i^l, v_1 a_{i,1} + s_i^u]$ and therefore the residual $r_i(\mathbf{v})$ in \eqref{eq:regression-loss} is lower bounded by:
    \begin{align}
        \myunderline{r}_i(v_1) = \begin{cases}
            h_i(v_1) + s_i^l\ &\textnormal{if}\ h_i(v_1) + s_i^l \geq 0; \\
            - h_i(v_1) - s_i^u\ &\textnormal{if}\ h_i(v_1) + s_i^u \leq 0; \\ 
            0 &\textnormal{otherwise}. 
        \end{cases}
    \end{align}
    \item We then take the following $\myunderline{f}_i$ to lower bound $f_i$:
    \begin{align}
        \myunderline{f}_i(v_1) = \min\{\myunderline{r}_i(v_1) , \xi \}. 
    \end{align}
    Summing $\myunderline{f}_i$ over $i$ gives $\myunderline{f}$, our underestimator for $f$. Given $\myunderline{f}$, we optimize, instead of the $n$-dimensional objective $f$ in \eqref{eq:GTM-ideal-LB}, the following problem to compute a lower bound on $\mathcal{C}_1 \times \mathbb{B}_{2:n}$:
    \begin{align}\label{eq:GTM_LB}
         L_{\textsf{GTM}} = \min_{v_1\in \mathcal{C}_1} \myunderline{f}(v_1) = \min_{v_1 \in \mathcal{C}_1} \sum_{i = 1} ^ M \myunderline{f}_i(v_1).
    \end{align}
    Note that \eqref{eq:GTM_LB} is a one-dimensional problem, similar to  \eqref{eq:GTM_UB}. Here, for robust linear regression, we do not exclude the possibility of tackling \eqref{eq:GTM_LB} via a customized solver, similarly to interval stabbing or that of \cite{Huang-CVPR2024}. However, even if such a solver exists, it would not be trivial to generalize it for more vision applications. Thus, we do not derive a tailored algorithm for the case here, but will instead pursue a general-purpose scheme in Section~\ref{sec:direct} that is able to solve both \eqref{eq:GTM_LB} and \eqref{eq:GTM_UB} globally optimally under mild assumptions.
\end{enumerate}

\subsubsection{Lower Bounds for General Residuals}\label{subsubsection:lb-general}
In deriving the lower bound for robust linear regression (Section~\ref{subsubsection:lb-regression}), step (a) eliminates the presence of variables $\mathbf{v}_{2:n}$, which allows us to derive a residual lower bound in step (b) and then formulate the single-variable optimization problem in step (c). We now generalize these steps for diverse residual functions under suitable assumptions:
\begin{enumerate}[label=(\alph*)]
    \item At high level, we could leverage again interval arithmetic to eliminate $\mathbf{v}_{2:n}$, as in Remark \ref{remark:interval-linear-reg}. Specifically, we assume the residual function $r_i$ consists of the compositions of basic arithmetic operations (e.g., sum, product, difference, division), start with the interval where $v_j$ lies in, inspect arithmetic operations acting on it and map the interval based on them (see Section~\ref{sec:pre_IA}). This eventually leads to a lower bound residual function in variable $v_1$. While intuitive, it is not easy to write a mathematical derivation down, without assumptions on the specific form of the residuals. To be mathematically precise, we assume our residual $r_i$ is of the form
    \begin{align}\label{eq:general-r_i}
        r_i(\mathbf{v}) = \|h_i(v_1) + g_i(\mathbf{v}_{2:n}) \|,
    \end{align}
    where $h_i=[h_i^{(1)},\dots,h_i^{(W)}]^\top$ and $g_i=[g_i^{(1)},\dots,g_i^{(W)}]^\top$ are now vector-valued functions that output $W$-dimensional vectors and $\| \cdot \|$ represents any vector norm; we will show the residuals for geometric problems in Sections~\ref{sec:2dest}-\ref{sec:4dest} can all be transformed into the above form. We furthermore assume that, given $\mathbf{v}_{2:n}\in \mathbb{B}_{2:n}$, we can  bound $g_i(\mathbf{v}_{2:n})$ entry-wise with
    \begin{align}
        \mathbf{s}^l_i \leq g_i(\mathbf{v}_{2:n})\leq \mathbf{s}^u_i,
    \end{align}
    where $\mathbf{s}^l_i, \mathbf{s}^u_i$ are $W$-dimensional vectors written as
    \begin{align}\label{eq:LB-general-s}
        \mathbf{s}^l_i=[s^{(1)l}_i,\dots, s^{(W)l}_i]^\top, \mathbf{s}^u_i=[s^{(1)u}_i,\dots, s^{(W)u}_i]^\top.
    \end{align}
    
    \item Applying step (b) in Section~\ref{subsubsection:lb-regression} in an entry-wise fashion to our residual $r_i(\mathbf{v})$ gives lower bound  $\myunderline{r}_i(v_1)=\| \myunderline{r}_i^{(1)}(v_1),\dots, \myunderline{r}_i^{(W)}(v_1) \|$, where  $\myunderline{r}_i^{(j)}(v_1)$ is equal to 
    \begin{align}\label{eq:general-r_ij}
        \begin{cases}
            h_i^{(j)}(v_1) + s_i^{(j)l}\ &\textnormal{if}\ h_i^{(j)}(v_1) + s_i^{(j)l} \geq 0; \\
            - h_i^{(j)}(v_1) - s_i^{(j)u}\ &\textnormal{if}\ h_i^{(j)}(v_1) + s_i^{(j)u} \leq 0; \\ 
            0 &\textnormal{otherwise}. 
        \end{cases}
    \end{align}
    \item Our final step to calculate an lower bound is identical to step (c) in Section~\ref{subsubsection:lb-regression}. In particular, see \eqref{eq:GTM_LB}. 
\end{enumerate}

\begin{algorithm}[!t]
\footnotesize
\caption{DIRECT Method for 1D Global Optimization}
\label{alg:DIRECT}
\textbf{Input:} Objective function $\chi(x)$ with interval $x\in[a^l, a^u]$; Tolerance $\varphi$; Minimal resolution $\varepsilon'$;

\textbf{Output:} A global minimizer $\hat{x}$;

    \begin{algorithmic}[1]
    \STATE $\mathcal{P} \gets$ An interval set $\{[a^l, a^u]\}$;
    \STATE $\hat{x} = \frac{a^l+a^u}{2}$;\ $\chi_{min} \gets \chi(\hat{x})$;
    \WHILE{$\mathcal{P}$ is not empty}
      \FOR{each $[a_i^l, a_i^u]$ with center $\dot{a}_i = \frac{a_i^l + a_i^u}{2}$ in $\mathcal{P}$}
      \STATE \textbf{if} $\exists$ $\tilde{K}_i > 0$ \textbf{such that}  \quad \textcolor{gray}{$\#$ Lipschitz constant check}
      \STATE \quad \textbf{for} all $[a_j^l, a_j^u]$ with $\dot{a}_j = \frac{a_j^l + a_j^u}{2}$ in $\mathcal{P}$:
      \STATE \quad \quad $\chi(\dot{a}_i) - \chi(\dot{a}_j) \leq \tilde{K}_i\cdot(\frac{a_i^u-a_i^l}{2} - \frac{a_j^u-a_j^l}{2})$,
      \STATE \quad \quad $\chi(\dot{a}_i) - \tilde{K}_i\cdot\frac{a_i^u-a_i^l}{2} \leq \chi_{min} - \varphi\cdot|\chi_{min}|$;
      \STATE \quad \textbf{end for}
      \STATE \textbf{then} Divide $[a_i^l, a_i^u]$ to three equal sub-intervals:
      \STATE \quad \textbf{for} each sub-interval with center $\dot{a}_i'$ \textbf{do}
      \STATE \quad \quad \textbf{if} $\chi(\dot{a}_i') < \chi_{min}$ \textbf{then} 
      \STATE \quad \quad \quad $\chi_{min} \gets \chi(\dot{a}_i')$, $\hat{x} \gets \dot{a}_i'$; 
      \STATE \quad \quad \textbf{end if}
      \STATE \quad \textbf{end for}
      \STATE \quad  \textbf{if} the sub-interval length $< \varepsilon$ \textbf{then} 
      \STATE \quad \quad \textbf{end while}
      \STATE \quad \textbf{end if}
      \STATE \quad Replace $[a_j^l, a_j^u]$ with its sub-intervals in $\mathcal{P}$;
      \STATE \textbf{end if}
      \ENDFOR
    \ENDWHILE
    \end{algorithmic}
\end{algorithm}

\subsection{DIRECT: Inner Solver for Upper and Lower Bounds }
\label{sec:direct}
We have derived the upper and lower bounds related to two 1-dimensional optimization problems as in \eqref{eq:GTM_UB} and  \eqref{eq:GTM_LB}.
To compute these bounds, an intuitive idea would be directly applying BnB again to these 1D problems. This gives a \textit{nested} BnB algorithm that has actually been shown to perform very well in several geometric problems~\cite{Yang-TPAMI2015, Campbell-ICCV2017, Campbell-TPAMI2018, Liu-TIP2018}. 
However, we argue that this nested approach can be sub-optimal as it does not fully leverage problem properties. Indeed, the vector function $h_i(v_1)$ in the general-form residual \eqref{eq:general-r_i} is often \textit{Lipschitz continuous} or can be reformulated as so; recall that a function $\chi$ is Lipschitz continuous if there is some constant $K>0$ such that $ \|\chi(\mathbf{x}_1) - \chi(\mathbf{x}_2)\| \leq K\|\mathbf{x}_1 - \mathbf{x}_2\|$ for all  $\mathbf{x}_1, \mathbf{x}_2 \in \textnormal{dom}(\chi)$. Furthermore, the Lipschitz continuity of $h_i(v_i)$ implies that our bounding functions are Lipschitz continuous as well:
\begin{proposition}\label{prop:Lips_inher}
    If the function $h_i(v_1)$ in \eqref{eq:general-r_i} is Lipschitz continuous, then the upper and lower bounding functions of \textsf{GTM} defined in \eqref{eq:GTM_UB} and \eqref{eq:GTM_LB} inherit its Lipschitz continuity.
\end{proposition}
The proof is in Appendix~\ref{sec:proof_prop1}.
Since vanilla 1D BnB search is unable to explicitly make use of Lipschitz continuity of our bounding functions, we instead consider existing algorithms based on \textit{global Lipschitz optimization}~\cite{Horst-BOOK2013, Pinter-BOOK1995, Bubeck-FTML2015} for computing our bounds in \eqref{eq:GTM_UB} and \eqref{eq:GTM_LB}. Our method of choice here is a classical one: the DIRECT method~\cite{Jones-JOTA1993, Jones-JOP2021}. In our context it has several advantages over 1D BnB: It performs Lipschitz optimization without the need to know Lipschitz constants; it is fast enough for low-dimensional and in particular 1D problems; it is general enough to be used as a block box and support our \textsf{GTM} method that is designed for diverse geometric residuals.

\begin{figure}[!t]
    \centering
    \includegraphics[width=0.45\textwidth]{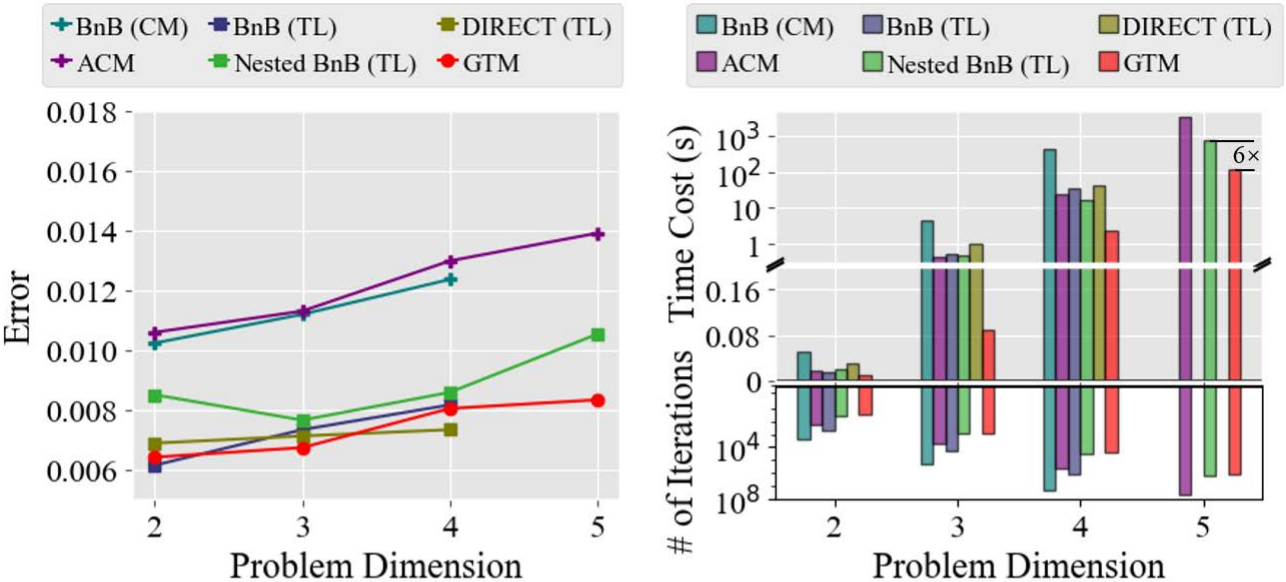}
    \\[-0.3em]
    \makebox[0.24\textwidth]{\footnotesize \quad \quad \quad \quad (a-1)}
    \makebox[0.24\textwidth]{\footnotesize \quad  \ (a-2)}
    \\[0.2em]
    \includegraphics[width=0.45\textwidth]{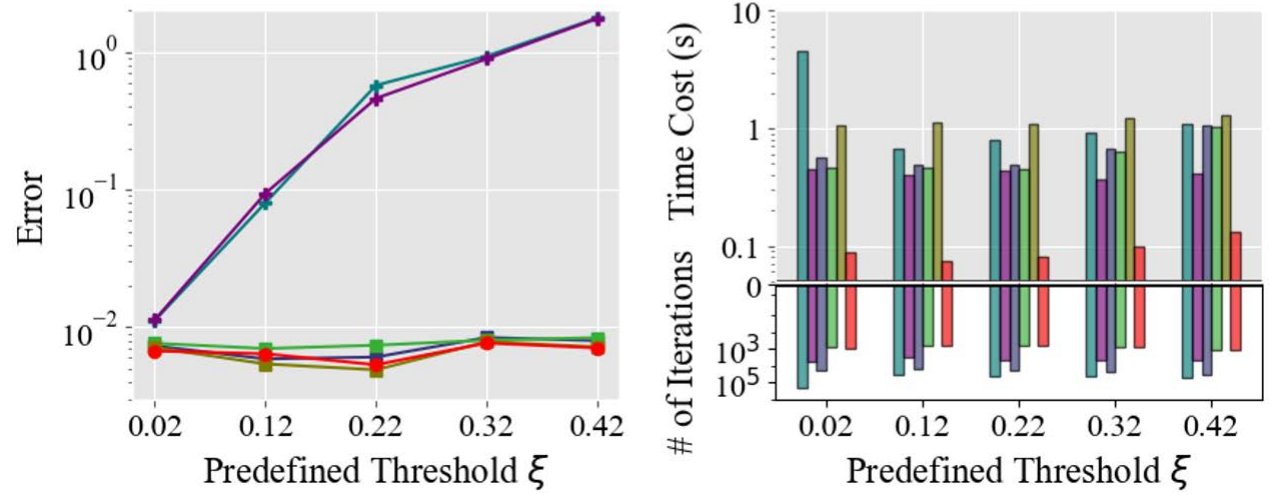}
    \\[-0.4em]
    \makebox[0.24\textwidth]{\footnotesize \quad \quad \quad \quad (b-1)}
    \makebox[0.24\textwidth]{\footnotesize \quad \ (b-2)}
    \\
    \vspace{-0.8em}
    \caption{Experiments on robust linear regression~(c.f.~Section~\ref{subsection:compare}). (a) Increasing the problem dimension with fixed threshold $\xi = 0.02$; (b) Increasing $\xi$ on the 3-dimensional problem. TL-based methods exhibit higher accuracy and threshold robustness than CM-based methods. Among TL-based methods, our \textsf{GTM} achieves a $3\times$-$45\times$ speed-up over standalone BnB or DIRECT.~($M = 500$, 90$\%$ outliers, 50 trials)}
    \label{fig:linear_eval}
\end{figure}

\textbf{DIRECT Method.} Algorithm~\ref{alg:DIRECT} outlines the DIRECT method for 1-dimensional problems. Similar to BnB, it iteratively divides the search space into smaller regions until finding a globally optimal solution up to a minimal resolution $\epsilon'$. Different from BnB, DIRECT does not compute any bounds, and instead chooses "potentially optimal" sub-branches based on the Lipschitz continuity of the objective (Lines 4-8). Giving up computing the bounds does not mean giving up global optimality: It can be shown that DIRECT is globally optimal for Lipschitz continuous functions and specifically for our bounding functions (Proposition~\ref{prop:Lips_inher}). Furthermore, sidestepping the bound computation brings two more benefits. First, it makes DIRECT several times faster than BnB for 1-dimensional problems~(see Table~\ref{tab:linear_eval_1Dtime}). Second, it does not require customized implementation for computing the bounds and only needs access to function values, which makes DIRECT applicable as a black box to our geometric problems, no matter how sophisticated their objective functions are. For these reasons, we employ the DIRECT method as our inner solver for \eqref{eq:GTM_UB} and \eqref{eq:GTM_LB}.

\noindent \textbf{Can we use standalone DIRECT?} Note that the DIRECT method can be also applied to problems with higher dimensions~(i.e., $\geq 2$)~\cite{Jones-JOTA1993}, therefore one may wonder if we can simply apply the standalone DIRECT method to solve \eqref{eq:TL} with various residuals.
While no previous works have made such an attempt to our knowledge, in the following we conduct validation experiments and find that DIRECT method loses its higher efficiency over vanilla BnB when the search dimension exceeds 2~(see Fig.~\ref{fig:linear_eval}(a-2)), which may results from the exponential increasing of potentially optimal regions along with the increasing problem dimension. 
Also, as a Lipschitz optimization method, DIRECT cannot guarantee the optimality for non-Lipschitz continuous functions, leading to inaccurate results even at $50\%$ outlier ratios~(see Figs.~\ref{fig:app3_simu_eval}(a)-(b) on rotational homography estimation). 

In summary, the proposed \textsf{GTM} framework for globally minimizing \eqref{eq:TL} involves $(n-1)$-dimensional BnB search with an embedded 1-dimensional DIRECT solver for bound computation.  
This hybrid design is our strategic blend that exploits the complementary strengths of BnB and DIRECT: first, the ($n-1$)-dimensional BnB search largely decreases the computational complexity of vanilla BnB; second, by fully leveraging the Lipschitz continuity of bounding functions, DIRECT can efficiently handle the remaining 1D optimization globally optimally. In the next section, we conduct evaluation experiments on robust linear regression and demonstrate that \textsf{GTM} is substantially more efficient than either vanilla BnB or DIRECT alone, while maintaining high robustness with global optimality guarantees.

\subsection{Comparing (Nested) BnB, DIRECT, and GTM}\label{subsection:compare}

Here we validate our design of \textsf{GTM} through robust linear regression experiments. 
To our knowledge, no prior work has systematically explored employing BnB or DIRECT solvers for the truncated loss problem in \eqref{eq:TL}. 
To address this gap, we implement related methods for the basic robust linear regression problem and compare their performance with \textsf{GTM}. We also evaluate the threshold robustness of \eqref{eq:TL} and \eqref{eq:CM}. The compared methods include: 
(1) \textsf{BnB (TL)} and \textsf{Nested BnB~(TL)}, vanilla and nested BnB solving \eqref{eq:TL} with residual \eqref{eq:regression-loss} as \textsf{GTM}; 
(2) \textsf{DIRECT~(TL)}, standalone DIRECT also solving \eqref{eq:TL};
(3) \textsf{BnB (CM)} and \textsf{ACM}~\cite{Zhang-TPAMI2024}, vanilla BnB and a recent $(n-1)-$dimeniosnal BnB method, both solving the \eqref{eq:CM} problem with residual \eqref{eq:regression-loss}. 
The prescribed error $\epsilon$ in BnB-based methods for \eqref{eq:TL}, the minimal resolution $\varepsilon$ and the tolerance $\rho$ in the DIRECT method are all set to $10^{-4}$ empirically, unless specified otherwise. More implementation details are presented in the appendix.

\begin{table}[!t]
\centering
    \caption{Average time~(ms) of the inner solvers for lower bound computation at each iteration on the experiments of Fig.~\ref{fig:linear_eval}; the inner solvers are 1D BnB in \textsf{Nested BnB~(TL)} and  1D DIRECT in \textsf{GTM}~(c.f.~Section~\ref{subsection:compare}).}
    \vspace{-0.8em}
    \label{tab:linear_eval_1Dtime}
    \footnotesize
    \renewcommand{\tabcolsep}{3.4pt} 
    \renewcommand\arraystretch{1.1}
    \begin{tabular}{l|cccc|cccc}
    \Xhline{1pt}
     \multirow{2}{*}{Method} &  \multicolumn{4}{c|}{Dimension~($\xi=0.02$)} & \multicolumn{4}{c}{$\xi$~(Dimension: 3)} \\
     \cline{2-9}
     & 2 & 3 & 4 & 5 & 0.12 & 0.22 & 0.32 & 0.42\\
     \Xhline{0.5pt}
     1D BnB & 0.256 & 0.307 & 0.388 & 0.453 & 0.312 & 0.309 & 0.321 & 0.320\\
     1D DIRECT & 0.053 & 0.058 & 0.064 & 0.077 & 0.048 & 0.056 & 0.061 & 0.057\\
     \Xhline{1pt}
    \end{tabular}
\end{table}

\noindent \textbf{Data Generation.}
The simulated data for robust linear regression are generated as follows. First, we randomly sample a ground-truth $\mathbf{v}^*$ from $\mathcal{N}(0, 3\mathbf{I}_n)$ and the vectors $\{\mathbf{a}_i\}_{i=1}^M$ from $\mathcal{N}(0, \mathbf{I}_n)$. Next, each $y_i$ is computed through $y_i = \mathbf{v}^{*\top}\mathbf{a}_i$ and then added Gaussian noise with radius $0.01$. To generate outliers, we replace a fraction of $y_i$'s with random Gaussian points sampled from $\mathcal{N}(0, 2)$.

\noindent \textbf{Results.} We conducted controlled experiments with varying problem dimension $n$ and threshold $\xi$: (a) $n$ is increased from $2$ to $5$ with $\xi = 0.02$; (b) $\xi$ is increased from $0.02$ to $0.42$ with $n = 3$.
Fig.~\ref{fig:linear_eval} presents the accuracy and timing results of various methods. For $n=5$, we do not show the results of \textsf{BnB~(CM)}, \textsf{BnB~(TL)}, and \textsf{DIRECT}, as they generally take more than several hours. Given the evaluation results, we can draw the following conclusions:
\begin{enumerate}
\item The \eqref{eq:TL}-based methods can always provide higher estimation accuracy and require much fewer iteration numbers than \eqref{eq:CM}-based methods across various problem dimensions~(see Fig.~\ref{fig:linear_eval}(a-1)); this strongly supports our motivation of applying BnB to solve \eqref{eq:TL}.
\item As $\xi$ increases, \eqref{eq:CM}-based methods easily fail, while \eqref{eq:TL}-based methods maintain high accuracy; this demonstrates the higher threshold-resilience of the \eqref{eq:TL} problem formulation~(see Fig.~\ref{fig:linear_eval}(b-1)).
\item Among various problem dimensions and threshold selections, our \textsf{GTM} always achieves the highest efficiency~(see Figs.~\ref{fig:linear_eval}(a-2) and (b-2)). Significantly, while all rely on a nested scheme, \textsf{GTM} achieves a $2\times$ - $30\times$ speed-up over \textsf{ACM} and an average $6\times$ speed-up over \textsf{Nested BnB~(TL)}, which verifies the effectiveness of the bounding functions in \textsf{GTM}.
\item The methods relying on $(n-1)$-dimensional BnB usually demand much fewer branching iteration numbers than methods relying on vanilla $n$-dimensional BnB, leading to much lower time cost~(see Figs.~\ref{fig:linear_eval}(a-2) and (b-2)). In addition, while \textsf{Nested BnB~(TL)} and \textsf{GTM} enjoy comparable iteration numbers, \textsf{GTM} is $5\times$ - $7\times$ times faster since the 1D DIRECT method is faster than 1D BnB search by this factor~(see Table~\ref{tab:linear_eval_1Dtime}).
\end{enumerate}

In the subsequent three sections, we apply the proposed \textsf{GTM} framework to three specific geometric estimation problems with different dimensions and diverse-form residuals. We will find that the derivation of general bounds in \textsf{GTM} greatly facilitates developing bounds for special cases in a concise manner. 
On each application, we compare \textsf{GTM} with related baseline methods on both simulated and real-world data to further verify the performance of \textsf{GTM}.

\section{Application~1: Relative Pose Estimation under Planar Motion}
\label{sec:2dest}
Two-view relative pose estimation of a moving camera is an essential geometric problem in many autonomous robot tasks, such as visual odometry and map-free visual re-localization~\cite{Mur-TRO2017, Arnold-ECCV2022}.
To solve the relative pose~(generally represented as the 5-DoF essential matrix with unit translation length), we usually obtain a set of outlier-corrupted 2D-2D keypoint matches and then conduct the outlier-robust estimation.
Note that in many applications, the camera is always attached to a planar-ground mobile robot or vehicle, leading to a planar motion constraint that can be employed to reduce the problem dimension.

As shown in Fig.~\ref{fig:app1_ill}, given the planar constraint, the motion of the camera is constrained to a 1-dimensional rotation and a 2-dimensional translation. 
While the problem is simplified, the existence of outlier matches and unsuitable threshold selection could still limit the performance of related methods~\cite{Choi-ICV2018, Liu-FN2022, Zhang-TPAMI2024}.
In the following, we will reparameterize the problem as 2-dimensional truncated loss minimization ~\eqref{eq:TL-1}, which we then solve to global optimality via our \textsf{GTM} solver that combines the 1-dimensional BnB search and 1-dimensional DIRECT method.

\subsection{Problem Formulation}
\label{sec:2dest_prop}
\begin{figure}[!t]
    \centering
    \includegraphics[width=0.36\textwidth]{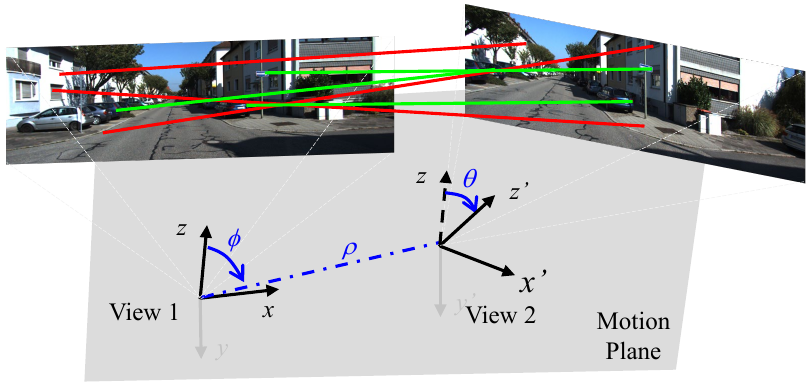}
    \\[-1em]
    \caption{Illustration of the planar motion constraint in the relative pose estimation problem~(c.f.~Section~\ref{sec:2dest}). The camera motion is constrained to 1-dimensional rotation parameterized by an angle \textcolor{blue}{$\theta$}, and 2-dimensional translation parameterized by the length \textcolor{blue}{$\rho$}~(usually set to unit due to monocular ambiguity) and angle \textcolor{blue}{$\phi$}. Given a set of 2D-2D matches consisting of both \textcolor{green}{inliers} and \textcolor{red}{outliers}, we aim to recover the \textcolor{blue}{$\theta$} and \textcolor{blue}{$\phi$}.} 
    \label{fig:app1_ill}
\end{figure}

Consider a set of 2D-2D keypoint matches $\{(\mathbf{u}_i,\ \mathbf{u}'_i)\}_{i=1}^M$ between two images captured by a forward-moving camera with planar motion~(Fig.~\ref{fig:app1_ill}), where $\mathbf{u}_i = [u_{1i},\ u_{2i}]^{\top} \in \mathbb{R}^2$ and $\mathbf{u}'_i = [u'_{1i},\ u'_{2i})]^{\top} \in \mathbb{R}^2$ are the normalized pixel coordinates in the respective images. Let $\mathbf{R}^*\in$ SO(3) and $\mathbf{t}^* = [t_1^*, t_2^*, t_3^*]^{\top} \in \mathbf{R}^3$ define the relative rotation and translation from view 2 to view 1. 
Given homogeneous coordinates $\bar{\mathbf{u}}_i = [\mathbf{u}_i^{\top},\ 1]^{\top} \in \mathbb{R}^3$ and $\bar{\mathbf{u}}_i' = [\mathbf{u}_i'^{\top},\ 1]^{\top} \in \mathbb{R}^3$, the ground-truth matches in the noiseless case should satisfy the well-known epipolar constraint
\begin{align} \label{eq:epi}
   (\bar{\mathbf{u}}_i)^{\top}\mathbf{E}^*\bar{\mathbf{u}}_i' = 0,
\end{align}
where $\mathbf{E}^* = [\mathbf{t}^*]_{\times}\mathbf{R}^* \in \mathbb{R}^{3\times3}$ defines the essential matrix and $[\cdot]_{\times}$ is the skew-symmetric matrix of the given vector.

Now, we take the planar motion constraint into consideration (Fig.~\ref{fig:app1_ill}). Without loss of generality, we assume the $y-$axis of the camera points downwards. In this case we can parameterize the camera rotation from view 2 to view 1 by a rotation angle $\theta^*$, which is uniquely identified with some 1D rotation around the $y-$axis. 
To represent the 2D translation vector, we let $\rho^*$ be the length and $\phi^*$ the motion angle of the translation with respect to the $z$-axis of view 1.

With auxiliary variables $\theta_1^* := \theta^* - \phi^* \in [-\pi, \pi]$ and $\theta_2^* := \phi^* \in [-\pi, \pi]$, we can now reformulate the epipolar constraint in \eqref{eq:epi} into (with some calculation omitted)
\begin{align}\label{eq:plane_con}
    A_{1i}\sin(\theta_1^*+\phi_{1i}) + A_{2i}\sin(\theta_2^*+\phi_{2i}) = 0,
\end{align}
where $A_{1i} = u_{2i}\sqrt{1+u_{1i}'^2}$, $\phi_{1i} = \arctan(u_{1i}')$, $A_{2i} = u_{2i}'\sqrt{1+u_{1i}^2}$, and $\phi_{2i} = - \arctan(u_{1i})$ are all constant. This motivates us to define the residual function $r_i(\theta_1, \theta_2)$ as 
\begin{align}
    r_i(\theta_1, \theta_2) = | A_{1i}\sin(\theta_1+\phi_{1i}) + A_{2i}\sin(\theta_2+\phi_{2i}) |
\end{align}
for $i$-th data sample. 
To apply truncated loss minimization, we take a predefined threshold $\xi$ and formulate:
\begin{equation}\label{eq:TL-1}
\begin{aligned}
    \min_{\substack{\theta_1\in[-\pi, \pi] \\ \theta_2 \in [-\pi, \pi]}} & \sum_{i=1}^{M} \min \{|h_i(\theta_1) + g_i(\theta_2)|,\ \xi\} \\
        \textnormal{s.t.}\ & {h_i}(\theta_1) = A_{1i}\sin(\theta_1+\phi_{1i}), \\
        & g_i(\theta_2) = A_{2i}\sin(\theta_2+\phi_{2i}),
\end{aligned} \tag{\textcolor{red}{TL-1}}
\end{equation}
Note that the residual $|h_i(\theta_1) + g_i(\theta_2)|$ is a special case of our general consideration in \eqref{eq:general-r_i}. 
Below we present the developed bounding functions of \eqref{eq:TL-1}, based on which we can apply \textsf{GTM} to solve the problem with global optimality.

\subsection{Bounding Functions for Solving \eqref{eq:TL-1}}
\label{sec:2d_bound}
Note that \eqref{eq:TL-1} is a 2-dimensional problem with variables $\theta_1 \in [-\pi, \pi]$ and $\theta_2 \in [-\pi, \pi]$. We employ the scheme of \textsf{GTM}~(c.f., Algorithm~\ref{alg:GTM}) to conduct BnB search over the 1D space of $\theta_2$. During the search, for an arbitrary given sub-branch $\mathbb{B}_2 = [\theta_2^l, \theta_2^u] \subseteq [-\pi, \pi]$ of $\theta_2$, we construct the upper and lower bounding functions in $\theta_1 \in [-\pi, \pi]$ as follows:

\noindent (\textbf{\textit{Upper Bound}}) In sub-branch $\mathbb{B}_2$, we follow \eqref{eq:GTM_UB} and provide an upper bound $U_{\textsf{GTM}-1}$ at the center $\dot{\theta}_2 = \frac{\theta_2^l + \theta_2^u}{2}$:
\begin{equation}\label{eq:U_app1}
    \begin{aligned}
        U_{\textsf{GTM}-1} &= \min_{\theta_1 \in [-\pi, \pi]} \sum_{i=1}^M \min\{|{h_i}(\theta_1) + g_i(\dot{\theta}_2)|,\ \xi\}. 
    \end{aligned}
\end{equation}
As $\theta_2$ lies in $[\theta_2^l, \theta_2^u]$, we can compute the range of $g_i(\theta_2)$ in \eqref{eq:TL-1} with constant time~(see Appendix~\ref{app:app1_range} for details).

\noindent (\textbf{\textit{Lower Bound}}) Again, we let $[{s_i^l}, {s_i^u}]$ denote the computable range of $g_i(\theta_2)$, then a lower bound $\myunderline{f}_i(\theta_1)$ of the truncated loss $f_i(\theta_1, \theta_2) := \min \{|{h_i}(\theta_1) + {g_i}(\theta_2)|,\ \xi\}$ can be derived as in steps (b) and (c) of Sections~\ref{subsubsection:lb-general}.
Next, based on $\myunderline{f}_i(\theta_1)$, a lower bound of \eqref{eq:TL-1} can be derived as in \eqref{eq:GTM_LB} by
\begin{equation}\label{eq:L_app1}
    \begin{aligned}
        L_{\textsf{GTM}-1} = \min_{\theta_1 \in [-\pi,\pi]} \sum_{i=1}^M {\myunderline{f}_i}(\theta_1).
    \end{aligned}
\end{equation}

Now, since $h_i(\theta_1)  = A_{1i}\sin(\theta_1+\phi_{1i})$ is Lipschitz continuous with the constant $|A_{1i}|$ for all $1\leq i \leq M$, we rely on Proposition~\ref{prop:Lips_inher} to obtain that, the bounding functions in \eqref{eq:U_app1} and \eqref{eq:L_app1} are Lipschitz continuous. As a consequence, solving them via DIRECT guarantees global optimality.

\subsection{Evaluation Experiments}
\label{sec:2dest-exper}
In the following, we compare our \textsf{GTM} with several baseline methods regarding relative pose estimation under planar motion on both simulated and real-world datasets.
The compared methods include: 
(1)~\textsf{RANSAC}~\cite{Fischler-CACM1981} and \textsf{MLESAC}~\cite{Torr-CVIU2000}, two heuristic methods implemented with minimal solvers~\cite{Choi-ICV2018}, note that \textsf{RANSAC} solves the CM version of \eqref{eq:TL-1} and \textsf{MLESAC} further considers the maximum likelihood loss;
(2)~\textsf{BnB~(CM)}~\cite{Liu-FN2022} and \textsf{ACM}~\cite{Zhang-TPAMI2024}, two search-based methods that employ BnB with $n-$dimensional and $(n-1)$-dimensional search respectively and also solve the CM problem; 
(3)~\textsf{DIRECT-2D}~\cite{Jones-JOTA1993}, which solves \eqref{eq:TL-1} via the standalone DIRECT method. For accuracy evaluation, we compute the absolute estimation errors of $\theta_1$ and $\theta_2$, and choose the maximum one for comparison~(denoted as max. angular error).
The maximum iteration numbers of \textsf{RANSAC} and \textsf{MLESAC} are set to $5k$.

\subsubsection{Experiments on Simulated Data}
\label{sec:2dest-sim}
\noindent \textbf{Data Generation}.
We project $M=2000$ randomly sampled 3D points distributed between 4 and 8 from the world frame origin to two camera frames with the same focal length of 800. 
A fraction of points are replaced by randomly sampled points before being projected to the second-view camera, thus they are considered outliers.
We add Gaussian distributed noise with radius 2 on each projected pixel before normalization.
To simulate the planar motion, we constrain the yaw angle of both cameras within $[-\pi/3, \pi/3]$ while the roll and pitch angles are set to 0. 
As to the camera translation, the angle $\phi$ is randomly sampled from $[-\pi/3, \pi/3]$ and the length $\rho$ from $[-2, 2]$. 

\begin{figure}[!t]
    \centering
    \includegraphics[width=0.44\textwidth]{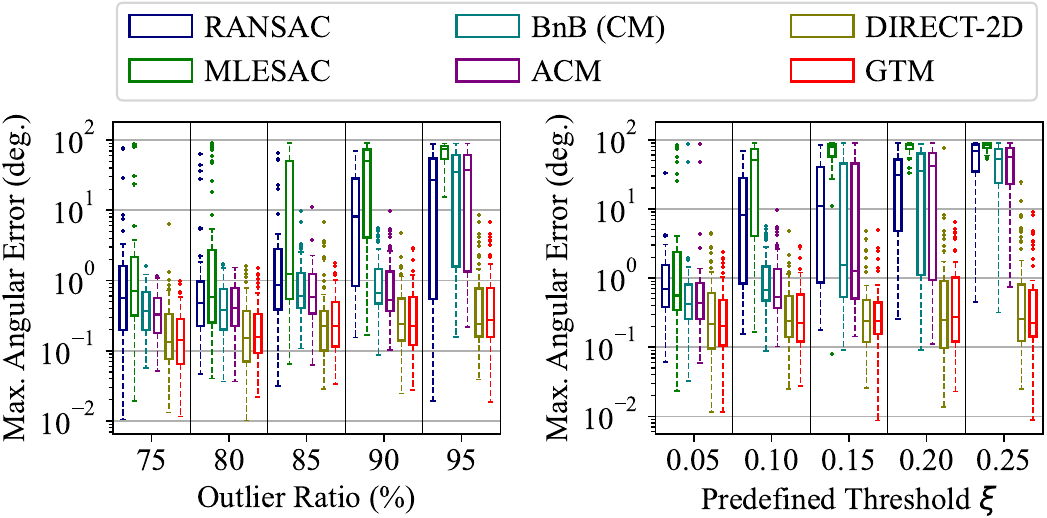}
    \\[-0.5em]
    \makebox[0.24\textwidth]{\footnotesize \quad \quad \quad \quad (a-1)}
    \makebox[0.24\textwidth]{\footnotesize \quad \quad \ (b-1)}
    \\[0em]
    \includegraphics[width=0.44\textwidth]{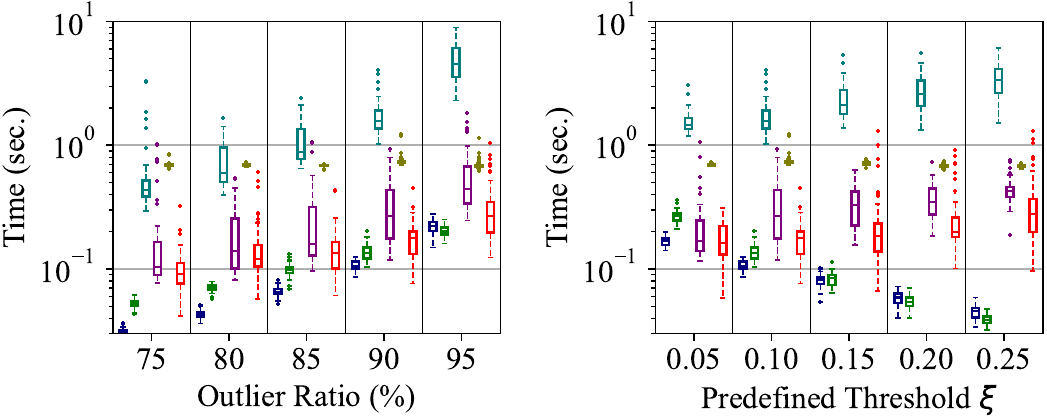}
    \\[-0.4em]
    \makebox[0.24\textwidth]{\footnotesize \quad \quad \quad \quad  (a-2)}
    \makebox[0.24\textwidth]{\footnotesize \quad \quad \ (b-2)}
    \\[0.2em]
    \includegraphics[width=0.44\textwidth]{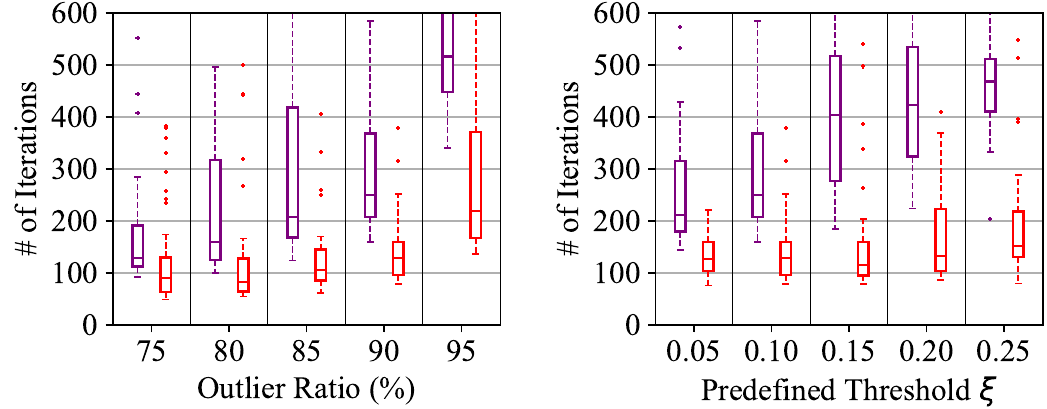}
    \\[-0.5em]
    \makebox[0.24\textwidth]{\footnotesize \quad \quad \quad \quad (a-3)}
    \makebox[0.24\textwidth]{\footnotesize \quad \quad \ (b-3)}
    \vspace{-2em}
    \caption{Evaluation results of application~1 on simulated datasets~(c.f.~Section~\ref{sec:2dest-sim}). (a) Increasing outlier ratio with fixed $\xi = 0.1$; (b) Increasing $\xi$ with fixed $90\%$ outlier ratio. Only our \textsf{GTM} and the \textsf{DIRECT-2D} can survive all cases, with \textsf{GTM} being several times faster; Moreover, while both relying on BnB, \textsf{GTM} usually runs several times faster than \textsf{ACM} as \textsf{GTM} requires much fewer branching iterations.~($M = 2000$, 50 trials)}
    \label{fig:app1_simu_eval}
\end{figure}

\noindent \textbf{Results}.
We compare the performance of various methods under different outlier ratios or predefined thresholds~(see Fig.~\ref{fig:app1_simu_eval}).
In most cases, heuristic methods like \textsf{RANSAC} and \textsf{MLESAC} fail despite their high efficiency. 
Under the extreme outlier ratio of 95\%, even the search-based \textsf{BnB~(CM)} and \textsf{ACM} fail. 
In addition, all the above methods easily fails as the threshold $\xi$ increases, due largely to the sensitivity of CM to the thresholds (see Fig.~\ref{fig:app1_simu_eval}(b-1)). In contrast, our \textsf{GTM} and the \textsf{DIRECT-2D} maintain high robustness across various setups, demonstrating the superiority of the \eqref{eq:TL-1} formulation and stability of its solution. 
Also, our \textsf{GTM} based on BnB\&DIRECT is several times faster than \textsf{DIRECT-2D}.
Moreover, while both conducting $(n-1)$-dimensional BnB, \textsf{GTM} usually takes a lot fewer iterations than \textsf{ACM}~(see Figs.~\ref{fig:app1_simu_eval}(a-3) and (b-3)). This comparison highlights the tightness of the bounds solved by the 1D DIRECT method.

\begin{table*}[!t]
\centering
    \caption{Evaluation results for planar motion estimation on the real-world KITTI dataset~(c.f.~Section~\ref{sec:2dest-real}). Max. angular errors are averaged results over all successful image pairs of each method. 
    Our \textsf{GTM} is more robust threshold $\xi$ than \textsf{ACM} and is several times faster than \textsf{DIRECT-2D}.}
    \vspace{-1em}
    \label{tab:app1_real_eval}
    \footnotesize
    \renewcommand{\tabcolsep}{3.2pt} 
    \renewcommand\arraystretch{1.1}
    \begin{tabular}{c|l|cccc|cccc}
    \Xhline{1pt}
     \multirow{2}{*}{Threshold $\xi$} & \multirow{2}{*}{Method \textbackslash \ Seq.} & \multicolumn{4}{c|}{00-05} & \multicolumn{4}{c}{06-10} \\
     \cline{3-10}
     & & \makecell{\textit{SR}~($\%$)$\uparrow$} & \makecell{Max. Angular\\Error~(deg.)$\downarrow$} & Time~(s)$\downarrow$ & \makecell{\# of\\Iterations$\downarrow$} & \makecell{\textit{SR}~($\%$)$\uparrow$} & \makecell{Max. Angular\\Error~(deg.)$\downarrow$} & Time~(s)$\downarrow$ & \makecell{\# of\\Iterations$\downarrow$} \\
     \Xhline{0.5pt}
     \multirow{3}{*}{$10^{-4}$} & \textsf{ACM}~\cite{Zhang-TPAMI2024} &  79.17 & 4.13 & 0.86 & 3779 & 78.91 & 3.11 & 0.82 & 4049\\
     & \textsf{DIRECT-2D}~\cite{Jones-JOTA1993} & 79.11 & \textbf{3.07} & 1.32 & - & \textbf{80.20} & 3.15 & 1.41 & -\\
     & \textsf{GTM} & \textbf{80.49} & 3.11 & \textbf{0.45} & \textbf{467} & 80.11 & \textbf{3.01} & \textbf{0.46} & \textbf{459} \\
     \Xhline{0.5pt}
     \multirow{3}{*}{$10^{-3}$} & \textsf{ACM}~\cite{Zhang-TPAMI2024} & 71.10 & 4.77 & 0.88 & 3829 & 74.07 & 3.41 & 0.79 & 3751\\
     & \textsf{DIRECT-2D}~\cite{Jones-JOTA1993} & 84.61 & \textbf{2.95} & 1.27 & - & 84.22 & \textbf{2.96} & 1.21 & - \\
     & \textsf{GTM} &  \textbf{84.65} & 2.96 & \textbf{0.25} & \textbf{311} & \textbf{84.56} & 2.99 & \textbf{0.27} & \textbf{320} \\
     \Xhline{0.5pt}
     \multirow{3}{*}{$10^{-2}$} & \textsf{ACM}~\cite{Zhang-TPAMI2024} &  44.15 & 4.98 & 0.76 & 3254 & 45.16 & 5.12 & 0.88 & 3805 \\
     & \textsf{DIRECT-2D}~\cite{Jones-JOTA1993} & 84.98 & 2.73 & 1.35 & - & \textbf{83.31} & 3.04 & 1.23 & -\\
     & \textsf{GTM} & \textbf{85.04} & \textbf{2.71} & \textbf{0.28} & \textbf{273} & 83.25 & \textbf{2.79} & \textbf{0.28} & \textbf{291}\\
     \Xhline{1pt}
    \end{tabular}
\end{table*}

\subsubsection{Experiments on Real-world Data}
\label{sec:2dest-real}
To further evaluate the performance of our method, we conduct experiments on the real-world KITTI dataset~\cite{Geiger-IJRR2013}.
We follow \cite{Liu-FN2022, Zhang-TPAMI2024} to process the first 11 sequences~(00-10) containing ground-truth.
The putative correspondences are obtained by matching SIFT features~\cite{Lowe-IJCV2004} for each pair of consecutive frames.
The number of correspondences $M$ is set to 1000. Then the obtained correspondences are normalized based on the calibrated intrinsic camera matrix. The parameters $\theta_1^*$ and $\theta_2^*$ are computed using the ground-truth relative pose of each image pair. We consider three threshold setups: $\xi = 10^{-4}$~(employed by~\cite{Zhang-TPAMI2024}), $10^{-3}$, and $10^{-2}$. 
Apart from the angular error, we also compare the success rates~(denoted as \textit{SR}) of various methods, i.e., the percentage of successful estimation where the max. angular error is below 10 degrees.

\noindent \textbf{Results}.
Table~\ref{tab:app1_real_eval} presents the evaluation results that again demonstrate the strength of \textsf{GTM}. We divide the 11 sequences into 2 groups for better presentation. Here, we do not compare with heuristic methods and \textsf{BnB~(CM)}, as they have been shown to be either unreliable or much slower than \textsf{ACM}~(see ~\cite[Table~1]{Liu-FN2022} and~\cite[Table~2]{Zhang-TPAMI2024}). In Table~\ref{tab:app1_real_eval}, when $\xi = 10^{-4}$, the three methods have comparable accuracy; as we increase $\xi$ to $10^{-2}$, the accuracy of \textsf{ACM} gets seriously affected, with the success rates decrease by more than $30\%$, which exposes the sensitivity of \eqref{eq:CM} to thresholds. 
In contrast, by solving the \eqref{eq:TL-1}, our \textsf{GTM} and the \textsf{DIRECT-2D} maintains high robustness, and even present higher success rates when $\xi$ is increased to $10^{-3}$ and $10^{-2}$.
These results verify the threshold robustness of \eqref{eq:TL-1} and our preference of it over \eqref{eq:CM}. Finally, we note that \textsf{GTM} is always several times faster than \textsf{ACM} and \textsf{DIRECT-2D} in different setups, despite the fact that, in some sense, the TL objective of \textsf{GTM} is harder to solve than that of \textsf{ACM}.

\section{Application~2: Point Cloud Registration}
\label{sec:3dest}
\begin{figure}[!t]
    \centering
    \includegraphics[width=0.3\textwidth]{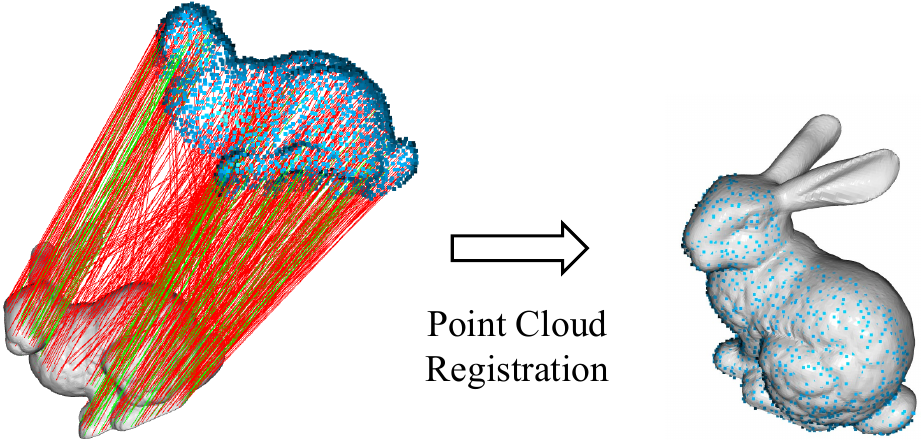}
    \\[-1em]
    \caption{Illustration of the outlier-robust point cloud registration problem~(c.f.~Section~\ref{sec:3dest}). Given a set of \textcolor{red}{outlier}-corrupted 3D-3D correspondences between the source model and the target point set, we aim to estimate a rigid transformation that best aligns them.} 
    \label{fig:app2_ill}
\end{figure}

In this section, we consider another geometric estimation task, i.e., rigid point cloud registration, which relates to finding a 6-DoF Euclidean transformation that best aligns two 3D point clouds. This is a fundamental problem in computer vision and robotics, playing an important role in 3D reconstruction and object pose estimation~\cite{Choi-ICV2018, Drost-CVPR2010}. Similar to other geometry problems, a commonly used pipeline for point cloud registration consists of correspondence identification and outlier-robust estimation~\cite{Huang-TPAMI2024,Bustos-TPAMI2017}. We focus on applying \textsf{GTM} to the outlier-robust estimation when a set of outlier-corrupted 3D-3D matches is given~(see Fig.~\ref{fig:app2_ill}).

Below, we will show that by algebraic operations, the 6-DoF constraint for rigid registration can be reduced to a 3-DoF constraint that involves the 3D translation only. Accordingly, we construct a 3-dimensional truncated loss minimization problem~\eqref{eq:TL-2} to estimate the translation, and we solve it to global optimality by \textsf{GTM} that blends the 2D BnB and 1D DIRECT method. 

\subsection{Problem Formulation}
Consider a set of 3D-3D correspondences $\{(\mathbf{p}_i, \mathbf{q}_i)\}_{i=1}^M$, where $\mathbf{p}_i = [p_{i1}, p_{i2}, p_{i3}]^\top$ belongs to the source cloud and $\mathbf{q}_i = [q_{i1}, q_{i2}, q_{i3}]^\top$  to the target cloud. Given the ground-truth rotation $\mathbf{R}^*\in$ SO(3) and translation $\mathbf{t}^* \in \mathbb{R}^3$, we assume the inlier correspondence satisfies the following 6-DoF constraint:
\begin{align}\label{eq:rigid_const}
    \mathbf{q}_i = \mathbf{R}^*(\mathbf{p}_i + \mathbf{t}^*) + \mathbf{\zeta}_i,\ \|\mathbf{\zeta}_i\|_2^2 \leq\xi^*,\ \mathbf{\zeta}_i \in \mathbf{R}^3,
\end{align}
where $\xi^*$ is some inlier threshold, $\mathbf{\zeta}_i$ the measurement noise, and $\|\cdot\|_2$ the $L_2$-norm. The above constraint implies 
\begin{align}
    & \mathbf{q}_i = \mathbf{R}^*(\mathbf{p}_i + \mathbf{t}^*) + \mathbf{\zeta}_i \notag \\
    \Rightarrow \ & \| \mathbf{q}_i - \mathbf{\zeta}_i \|_2^2 = \|\mathbf{R}^*(\mathbf{p}_i + \mathbf{t}^*) \|_2^2 = \|\mathbf{p}_i + \mathbf{t}^*\|_2^2 \label{eq:R_ele}\\
    \Rightarrow \ & \|\mathbf{q}_i\|_2^2 - \|\zeta_i\|_2^2 \leq \|\mathbf{p}_i + \mathbf{t}^*\|_2^2 \leq \|\mathbf{q}_i\|_2^2 + \|\zeta_i\|_2^2 \\
    \Rightarrow \ & \big| \|\mathbf{p}_i + \mathbf{t}^*\|_2^2 - \|\mathbf{q}_i\|_2^2 \big| \leq \|\zeta_i\|_2^2 \\
    \Rightarrow \ & \big| \|\mathbf{p}_i + \mathbf{t}^*\|_2^2 - \|\mathbf{q}_i\|_2^2 \big| \leq \xi^*, \label{eq:t_const}
\end{align}
where \eqref{eq:R_ele} holds as $\|\mathbf{R}\mathbf{a}\|_2 = \|\mathbf{a}\|_2$ for any $\mathbf{R} \in $ SO(3) and $\mathbf{a}\in \mathbb{R}^3$. 
In the above, we have relaxed the 6-DoF transformation constraint in \eqref{eq:rigid_const} to the constraint defined in \eqref{eq:t_const} which involves the 3-DoF translation $\mathbf{t}^*$. Accordingly, given an arbitrary $\mathbf{t} = [t_1, t_2, t_3]^{\top} \in \mathbb{R}^3$, we define the residual function $r_i(\mathbf{t})$ for every $i=1,\dots,M$ by 
\begin{align}\label{eq:r_app2}
     r_i(\mathbf{t}) &= \big| \|\mathbf{p}_i + \mathbf{t}\|_2^2 - \|\mathbf{q}_i\|_2^2 \big| \\
    &= \big| (t_1 + p_{i1})^2 + (t_2 + p_{i2})^2 + (t_3 + p_{i3})^2 - \|\mathbf{q}_i\|_2^2 \big|. \notag
\end{align}
With some predefined threshold $\xi>0$, we proceed by formulating the following optimization problem:
\begin{equation}\label{eq:TL-2}
    \begin{aligned}
        \min_{\mathbf{t}\in\mathcal{C}} & \sum_{i=1}^M \min\{ | h_i(t_1) + g_i(t_2, t_3) |,\ \xi\} \\
        \textnormal{s.t.}\  &h_i(t_1)=(t_1 + p_{i1})^2, \\
        & g_i(t_2, t_3)=(t_2 + p_{i2})^2 + (t_3 + p_{i3})^2 - \|\mathbf{q}_i\|_2^2, 
    \end{aligned} \tag{\textcolor{red}{TL-2}}
\end{equation}
Here, $\mathcal{C}$ is the initial 3-dimensional rectangle $\mathcal{C} = [c_1^l, c_1^u] \times [c_2^l, c_2^u] \times [c_3^l, c_3^u] \in \mathbb{R}^3$ that we assume contains the ground-truth translation $\mathbf{t}^*$. This assumption is without loss of generality as $\mathcal{C}$ can always be set to be large enough. Again, it is readily seen that the residual here is a special case of \eqref{eq:general-r_i}. Thus, in what follows, we specialize our general bounding functions developed earlier to the case here.

\subsection{Bounding Functions for Solving \eqref{eq:TL-2}}
\label{sec:3d_bound}
Note that \eqref{eq:TL-2} is a 3-dimensional problem with variables $t_1 \in [c_1^l, c_1^u]$, $t_2 \in [c_2^l, c_2^u]$, and $t_3 \in [c_3^l, c_3^u]$. Again, according to the scheme of our \textsf{GTM}, we conduct the BnB search over the two variables $t_2$ and $t_3$; and during the 2D search, given a sub-branch $\mathbb{B}_{2:3} = [t_2^l,t_2^u]\times[t_3^l, t_3^u] \subseteq [c_2^l, c_2^u]\times[c_3^l, c_3^u]$, we construct the upper and lower bounding functions in the left variable $t_1 \in [c_1^l, c_1^u]$ as follows:

\noindent (\textbf{\textit{Upper Bound}}) As in \eqref{eq:GTM_UB}, we take the center in $\mathbb{B}_{2:3}$,  $\dot{t}_2 = \frac{t_2^l+t_2^u}{2}$ and $\dot{t}_3 = \frac{t_3^l+t_3^u}{2}$, to compute upper bound $U_{\textsf{GTM}-2}$:
\begin{align}\label{eq:U_app2}
    U_{\textsf{GTM}-2} &= \min_{t_1 \in [c_1^l, c_1^u]} \sum_{i=1}^M\min\{|h_i(t_1) + g_i(\dot{t}_2, \dot{t}_3)|,\ \xi\}.
\end{align}

\noindent (\textbf{\textit{Lower Bound}}) Given $[t_2, t_3]^{\top} \in \mathbb{B}_{2:3}$, we  compute the range $[s_i^l, s_i^u]$ of $g_i(t_2, t_3)$ with constant time~(see Appendix~\ref{app:app2_range} for details). Then an underestimator $\myunderline{f}_i(t_1)$ of $f_i(\mathbf{t}) := \min\{ | {h_i}(t_1) + {g_i}(t_2, t_3) |,\ \xi\}$ can be computed as in steps (b) and (c) of Section~\ref{subsubsection:lb-general}.
Accordingly, a lower bound of \eqref{eq:TL-2} can be derived as in \eqref{eq:GTM_LB} by
\begin{align}\label{eq:L_app2}
 L_{\textsf{GTM}-2} = \min_{t_1\in[c_1^l, c_1^u]} \sum_{i=1}^M {\myunderline{f}_i(t_1)}.
\end{align}

Similar to the bounds $U_{\textsf{GTM}-1}$ and $L_{\textsf{GTM}-1}$ of the first application in Section~\ref{sec:2d_bound}, the bounds $U_{\textsf{GTM}-2}$ and $L_{\textsf{GTM}-2}$ here also both relate to solving 1-dimensional problems. While $h_i(t_1) = (t_1 + p_{i1})^2$ is not Lipschitz continuous in the full domain $\mathbb{R}$, it is indeed Lipschitz continuous on the interval $[c_1^l, c_1^u]$. Thus, based on Proposition~\ref{prop:Lips_inher} we confirm the Lipschitz continuity of the two bounding functions in \eqref{eq:U_app2} and \eqref{eq:L_app2}. With these, the DIRECT method is guaranteed to yield globally optimal solutions for computing the bounds.

\subsection{Solving the 6-DoF Transformation}
\label{sec:3dest-rigid}

Once a globally optimal 3-DoF translation $\hat{\mathbf{t}}$ is solved from~\eqref{eq:TL-2} via \textsf{GTM}, the remaining rotation can be naturally computed by robust rotation estimation methods introduced by~\cite{Peng-CVPR2022} or \cite{Bustos-ICCV2015}. However, we argue that a much easier and faster method can be employed. 
Actually, after solving~\eqref{eq:TL-2}, apart from the translation $\hat{\mathbf{t}}$, we can calculate the residual $r_i(\hat{\mathbf{t}})$ of each correspondence; note that the inliers typically hold lower residuals. By selecting the point pairs with the lowest residuals, we form a subset of correspondences with a high inlier ratio. While there might be a few outliers left in the subset, we employ Fast Global Registration (FGR)~\cite{Zhou-ECCV2016}, a fast and reliable 3D registration method designed for low outlier ratios, on the new subset to compute the 6-DoF transformation as the final estimation result.
Note that FGR also avoids explicit threshold selection via a graduated non-convexity procedure, which makes it ideal for this refinement stage.

\begin{figure}[!t]
    \centering
    \includegraphics[width=0.48\textwidth]{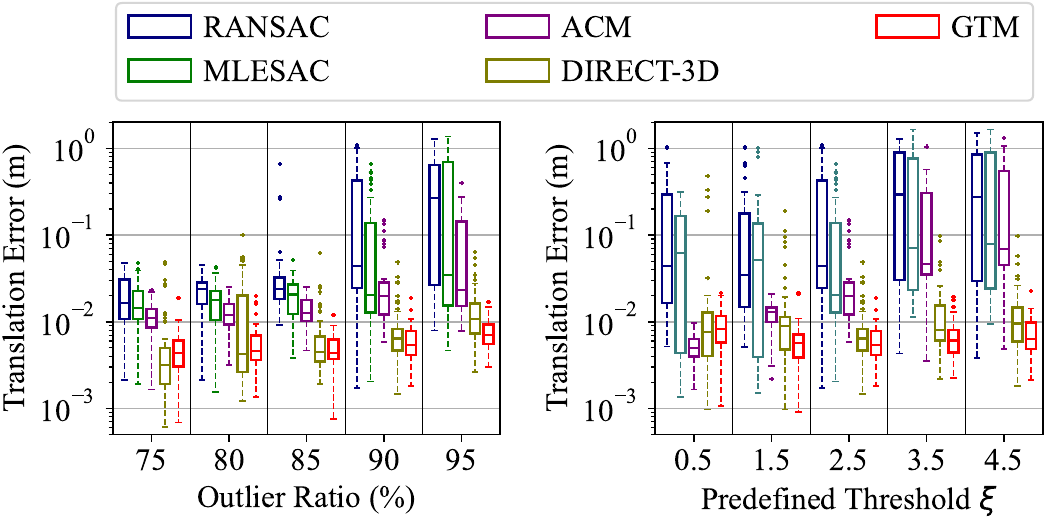}
    \\[-0.5em]
    \makebox[0.24\textwidth]{\footnotesize \quad \quad \quad (a-1)}
    \makebox[0.24\textwidth]{\footnotesize \quad \quad \quad \ \ (b-1)}
    \\[0.2em]
    \includegraphics[width=0.48\textwidth]{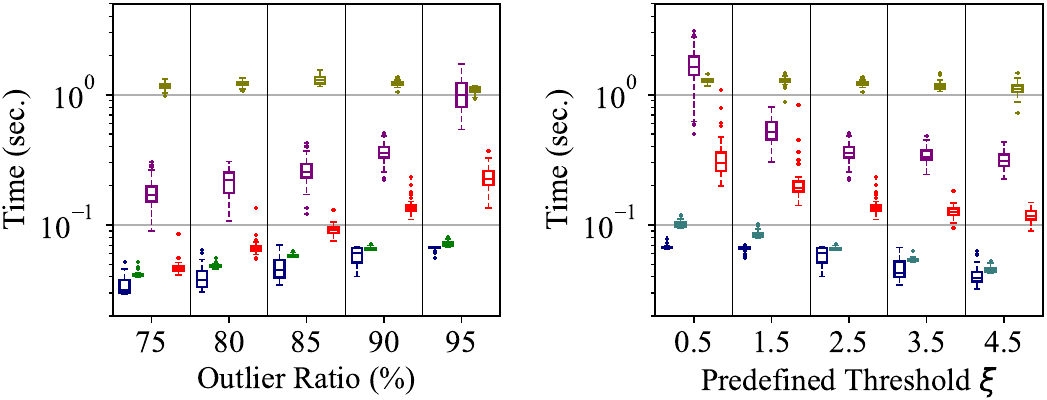}
    \\[-0.4em]
    \makebox[0.24\textwidth]{\footnotesize \quad \quad \quad (a-2)}
    \makebox[0.24\textwidth]{\footnotesize \quad \quad \quad \ \ (b-2)}
    \\[0.1em]
    \includegraphics[width=0.48\textwidth]{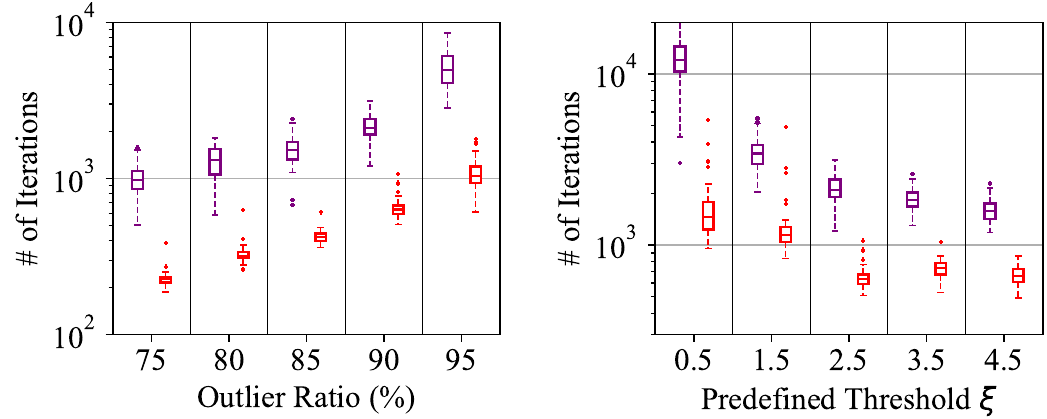}
    \\[-0.5em]
    \makebox[0.24\textwidth]{\footnotesize \quad \quad \quad (a-3)}
    \makebox[0.24\textwidth]{\footnotesize \quad \quad \quad \ \ (b-3)}
    \vspace{-2em}
    \caption{Evaluation results for 3-DoF translation estimation on simulated data~(c.f.~Section~\ref{sec:3dest-sim}). (a) Varying outlier rates with fixed $\xi = 2.5$; (b) Varying $\xi$ with fixed $90\%$ outlier rates. In various setups, only \textsf{GTM} maintains high robustness. Moreover, 
    \textsf{GTM} usually runs 3-15 times faster than \textsf{ACM} and \textsf{DIRECT-3D}.~($M = 1000$, 50 trials)}
    \label{fig:app2_simu_eval}
\end{figure}

\subsection{Evaluation Experiments}
\label{sec:3dest-exper}
In the following, we compare our \textsf{GTM} with several baselines and state-of-the-arts~(SoTAs) regarding point cloud registration on simulated and real-world datasets.
Recall that our approach involves estimating the 3-DoF translation by solving \eqref{eq:TL-2} based on \textsf{GTM} and estimating the 6-DoF rigid transformation via post-processing with FGR.
For clarity, in simulated experiments, we focus on analyzing the performance of our \textsf{GTM} part for translation estimation.
Then in real-world experiments, we analyze both rotation and translation errors. 
To evaluate the estimation accuracy, the translation error is computed by $\textit{E}_\mathbf{t} = \|\mathbf{\hat{t}}-\mathbf{t}^{*}\|$ and the rotation error by $\textit{E}_\mathbf{R} = \arccos{\frac{\textnormal{Tr}(\mathbf{\hat{R}}^{\top}\mathbf{R}^{*})-1}{2}}$, where $\mathbf{\hat{t}}$ and $\mathbf{\hat{R}}$ denote the estimated rotation and translation, respectively; $\mathbf{t}^{*}$ and $\mathbf{R}^{*}$ are the ground-truth rotation and translation, respectively; $\textnormal{Tr}(\cdot)$ represents the trace of a matrix.

\begin{table*}[!t]
\centering
    \caption{Evaluation results for 6-DoF rigid pose estimation on the real-world ETH dataset~(c.f.~Section~\ref{sec:3dest-real}). The best results are \textbf{boldfaced}. The second best are \myunderline{underlined}. Running times are marked as \textcolor{gray}{gray} for methods with \textit{SR} lower than $50\%$. \textsf{GTM} is the most robust to threshold variations.}
    \vspace{-0.8em}
    \label{tab:app2_real_eval}
    \footnotesize
    \renewcommand{\tabcolsep}{5pt} 
    \renewcommand\arraystretch{1}
    \begin{tabular}{c|l|ccc|ccc|ccc}
        \Xhline{1pt}
         \multirow{2}{*}{Threshold $\xi$} &  \multirow{2}{*}{Method \textbackslash \ Scene} & \multicolumn{3}{c|}{Arch \& Facade} & \multicolumn{3}{c|}{Courtyard} & \multicolumn{3}{c}{Office \& Trees}\\
        \cline{3-11}
         & & \makecell{\textit{SR}~$\uparrow$} & \makecell{$\textit{E}_\mathbf{R}/\textit{E}_\mathbf{t}$\\($^{\circ}$/cm)} $\downarrow$ & Time~(s)~$\downarrow$ 
         & \makecell{\textit{SR}~$\uparrow$} & \makecell{$\textit{E}_\mathbf{R}/\textit{E}_\mathbf{t}$\\($^{\circ}$/cm)} $\downarrow$ & Time~(s)~$\downarrow$ & \makecell{\textit{SR}~$\uparrow$}  & \makecell{$\textit{E}_\mathbf{R}/\textit{E}_\mathbf{t}$\\($^{\circ}$/cm)} $\downarrow$ & Time~(s)~$\downarrow$ \\
        \Xhline{0.5pt}
        \multirow{7}{*}{$0.04$} & \textsf{RANSAC}~\cite{Fischler-CACM1981} & 11/31 & 1.04/15.37 & \textcolor{gray}{0.19} & 13/28 & 0.47/13.26 & \textcolor{gray}{0.41} & 2/25 & 0.64/10.11 & \textcolor{gray}{0.17} \\
        & \textsf{MLESAC}~\cite{Torr-CVIU2000} & 14/31 & 1.10/21.96 & \textcolor{gray}{0.24} & 9/28 & 0.55/10.84 & \textcolor{gray}{0.53} & 2/25 & 0.61/9.96 & \textcolor{gray}{0.32}\\
        & \textsf{FGR}~\cite{Zhou-ECCV2016} & 9/31 & 0.95/19.98 & \textcolor{gray}{0.21} & 2/28 & 1.40/30.01 & \textcolor{gray}{0.46} & 0/25 & -/- & \textcolor{gray}{0.23}\\
        & \textsf{TEASER++}~\cite{Yang-TRO2020} & \myunderline{29/31} & 0.25/\myunderline{5.87} & \myunderline{0.42} & \textbf{28/28} & 0.09/3.87 & 1.43 & \textbf{23/25} & 0.52/7.23 & 2.08 \\
        & \textsf{MAC}~\cite{Yang-TPAMI2024} & 27/31 & \textbf{0.17}/6.28 & 4.66 & \textbf{28/28} & 0.11/4.01 & 18.29 & 20/25 & \myunderline{0.31}/\textbf{5.62} & 25.30 \\
        & \textsf{HERE}~\cite{Huang-TPAMI2024} & \textbf{30/31} & 0.27/5.96 & \textbf{0.27} & \textbf{28/28} & \myunderline{0.06}/\myunderline{3.83} & \textbf{0.71} & \myunderline{22/25} & 0.54/7.11 & \textbf{0.78}\\
        & \textsf{GTM} & \myunderline{29/31} & \myunderline{0.22}/\textbf{5.71} & 0.98 & \textbf{28/28} & \textbf{0.05}/\textbf{3.77} & \myunderline{1.31} & \textbf{23/25} & \textbf{0.30}/\myunderline{6.94} & \myunderline{1.89}\\
        \Xhline{0.5pt}
        \multirow{7}{*}{$4$} & \textsf{RANSAC}~\cite{Fischler-CACM1981} & 9/31 & 1.38/34.87 & \textcolor{gray}{0.20} & 10/28 & 0.52/24.07 & \textcolor{gray}{0.38} & 0/25& -/- & \textcolor{gray}{0.18} \\
        & \textsf{MLESAC}~\cite{Torr-CVIU2000} & 7/31 & 1.42/27.95 & \textcolor{gray}{0.22} & 7/28 & 0.60/19.14 & \textcolor{gray}{0.46} & 1/25 & 1.74/41.13 & \textcolor{gray}{0.35} \\
        & \textsf{FGR}~\cite{Zhou-ECCV2016} & 12/31 & 0.88/21.54 & \textcolor{gray}{0.20} & 9/28 & 0.28/17.24 & \textcolor{gray}{0.41} & 4/25 & 1.28/18.77 & \textcolor{gray}{0.38} \\
        & \textsf{TEASER++}~\cite{Yang-TRO2020} & 23/31 & 1.15/\myunderline{11.44} & 3.27 & \textbf{28/28} & 0.27/10.15 & 6.06 & 16/25 & 1.03/\textbf{11.73} & 84.14\\
        & \textsf{MAC}~\cite{Yang-TRO2020} & \myunderline{26/31} & \myunderline{0.41}/11.83 & 8.83 & \textbf{28/28} & 0.21/9.22 & 21.86 & \myunderline{20/25} & \textbf{0.74}/16.27 & \myunderline{48.09}\\
        & \textsf{HERE}~\cite{Huang-TPAMI2024} & 21/31 & 0.57/14.46 & \textbf{0.31} & \myunderline{26/28} & \textbf{0.10}/\myunderline{5.57} & \textbf{1.75} & 7/25 & 1.15/13.77 & \textcolor{gray}{2.86}\\
        & \textsf{GTM} & \textbf{27/31} & \textbf{0.38}/\textbf{9.21} & \myunderline{1.76} & \textbf{28/28} & \myunderline{0.17}/\textbf{5.33} & \myunderline{2.33} & \textbf{22/25} & \textbf{0.66}/\myunderline{11.89} & \textbf{3.18}\\
        \Xhline{1pt}
    \end{tabular}
\end{table*}

\subsubsection{Experiments on Simulated Data}
\label{sec:3dest-sim}
To investigate the effectiveness of our \textsf{GTM} for solving~\eqref{eq:TL-2}, we compare with the  baselines that could be used to solve the translation: 
(1)~\textsf{RANSAC}~\cite{Fischler-CACM1981} and its variant \textsf{MLESAC}~\cite{Torr-CVIU2000};
(2)~\textsf{ACM}~\cite{Zhang-TPAMI2024} that solves the CM version of \eqref{eq:TL-2};
(3)~\textsf{DIRECT-3D}~\cite{Jones-JOTA1993} that solves \eqref{eq:TL-2} via the standalone DIRECT method. The maximum iteration numbers of \textsf{RANSAC} and \textsf{MLESAC} are set to $5k$.

\noindent \textbf{Data Generation}.
First, we follow the data generation in Section~\ref{sec:2dest-sim} to generate $M = 1000$ randomly sampled 3D points and treat them as the source cloud. 
Then, we apply a random rotation and translation sampled in a unit cube to each point in the source cloud, and add noise from a Gaussian distribution with zero mean and 0.02 variance to each point, leading to the target cloud.
Next, to generate outliers, we replace a fraction of points in the target cloud with randomly sampled points ranging between 4 and 8 from the origin of the world frame.

\noindent \textbf{Results}. Fig.~\ref{fig:app2_simu_eval} presents the performance of various methods under different setups. The two heuristic methods, \textsf{RANSAC} and \textsf{MLESAC}, easily fail at high outlier ratios~(i.e.,~$\geq 90\%$). Methods based on global search, such as \textsf{ACM} and \textsf{DIRECT-3D}, sometimes fail at high outlier ratios or with large, misspecified thresholds. On the other hand, \textsf{GTM} remains robust in all setups with higher accuracy. In particular, \textsf{GTM} achieves a 5 to 20 times speed gain over \textsf{ACM} and \textsf{DIRECT-3D}. 
While \textsf{ACM} also conducts BnB in the ($n-1$)-dimensional space, \textsf{GTM} converges in much fewer iterations~(see Figs.~\ref{fig:app2_simu_eval}(a-3) and (b-3)).
These demonstrate the effectiveness of our combination of the ($n-1$)-dimensional BnB and 1-dimensional DIRECT method.



\begin{figure*}
    \centering
    \footnotesize
    \renewcommand{\tabcolsep}{4pt}
    \begin{tabular}{ccccc}
      \makecell{\includegraphics[width=0.10\textwidth]{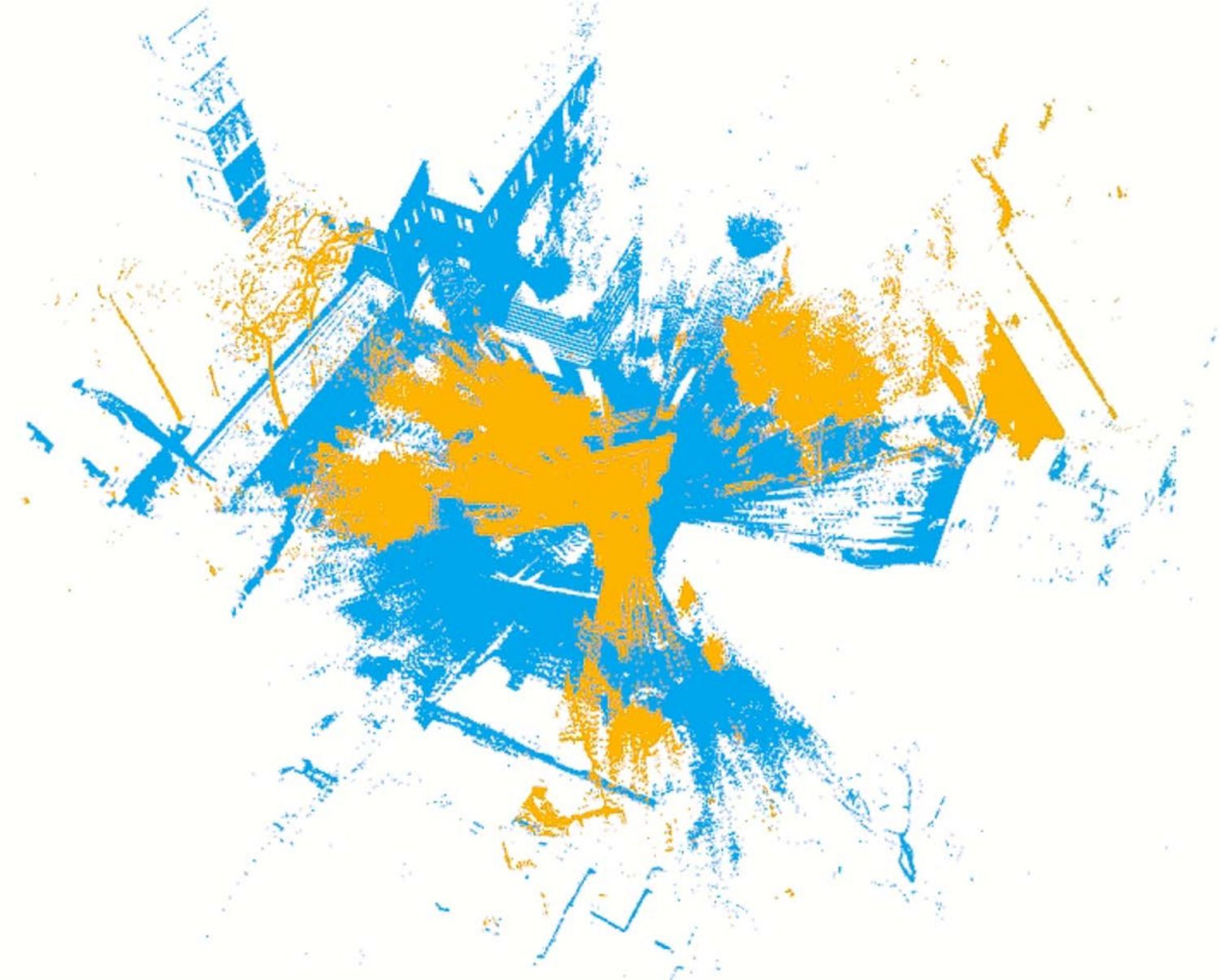} \\ (a) Inputs: \\ Point Pairs - 8953 \\ \# of Inliers - 189} &
      \makecell{\includegraphics[width=0.10\textwidth]{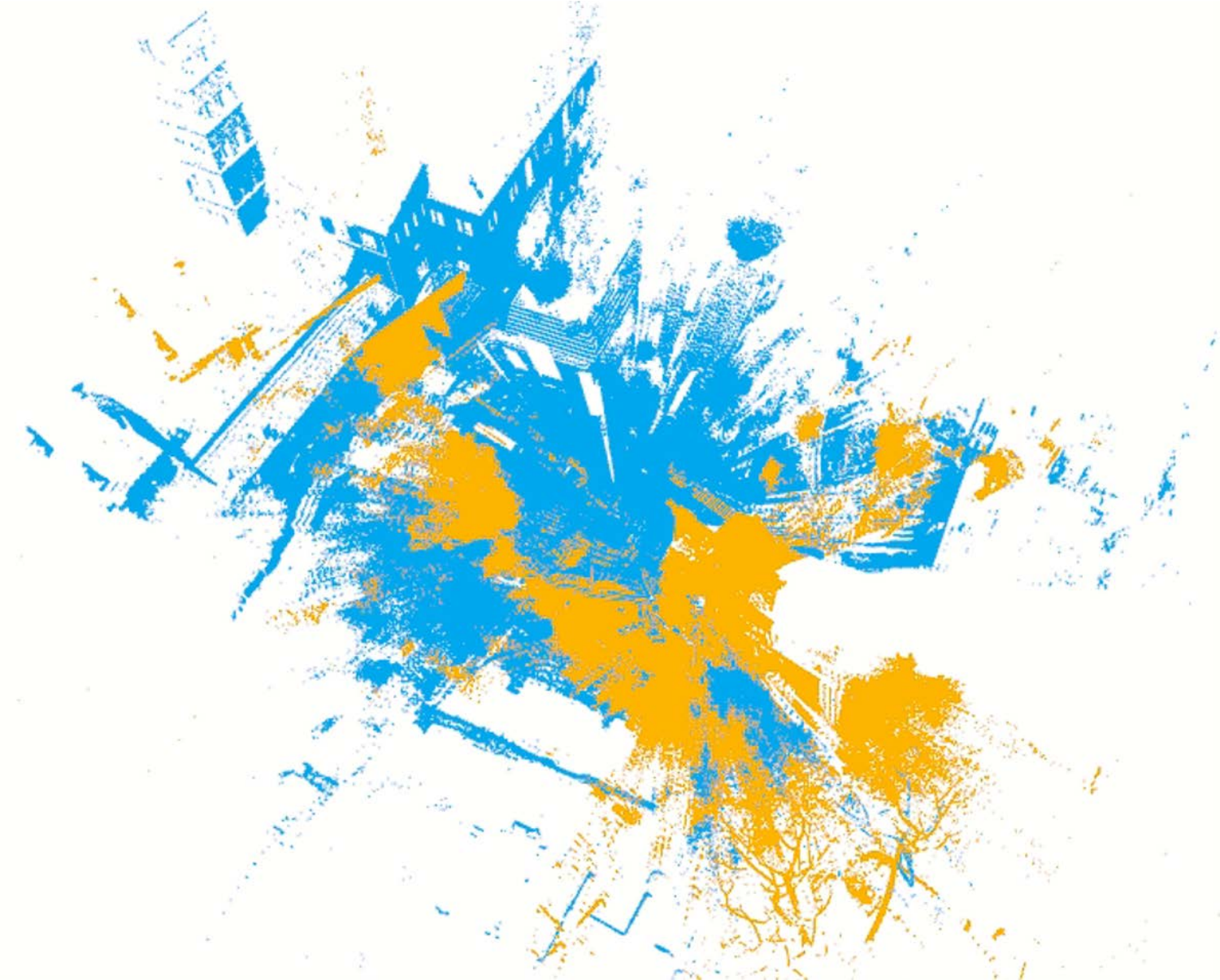} \\ (b-1) \textsf{RANSAC}~($\xi = 0.04$) \\ $\textit{E}_\mathbf{R}$ = 16.98$^{\circ}$, \ $\textit{E}_\mathbf{t}$ = 357.76cm \\ t = 0.38s~(\textcolor{red}{Fail}) \\[0.5em] \includegraphics[width=0.10\textwidth]{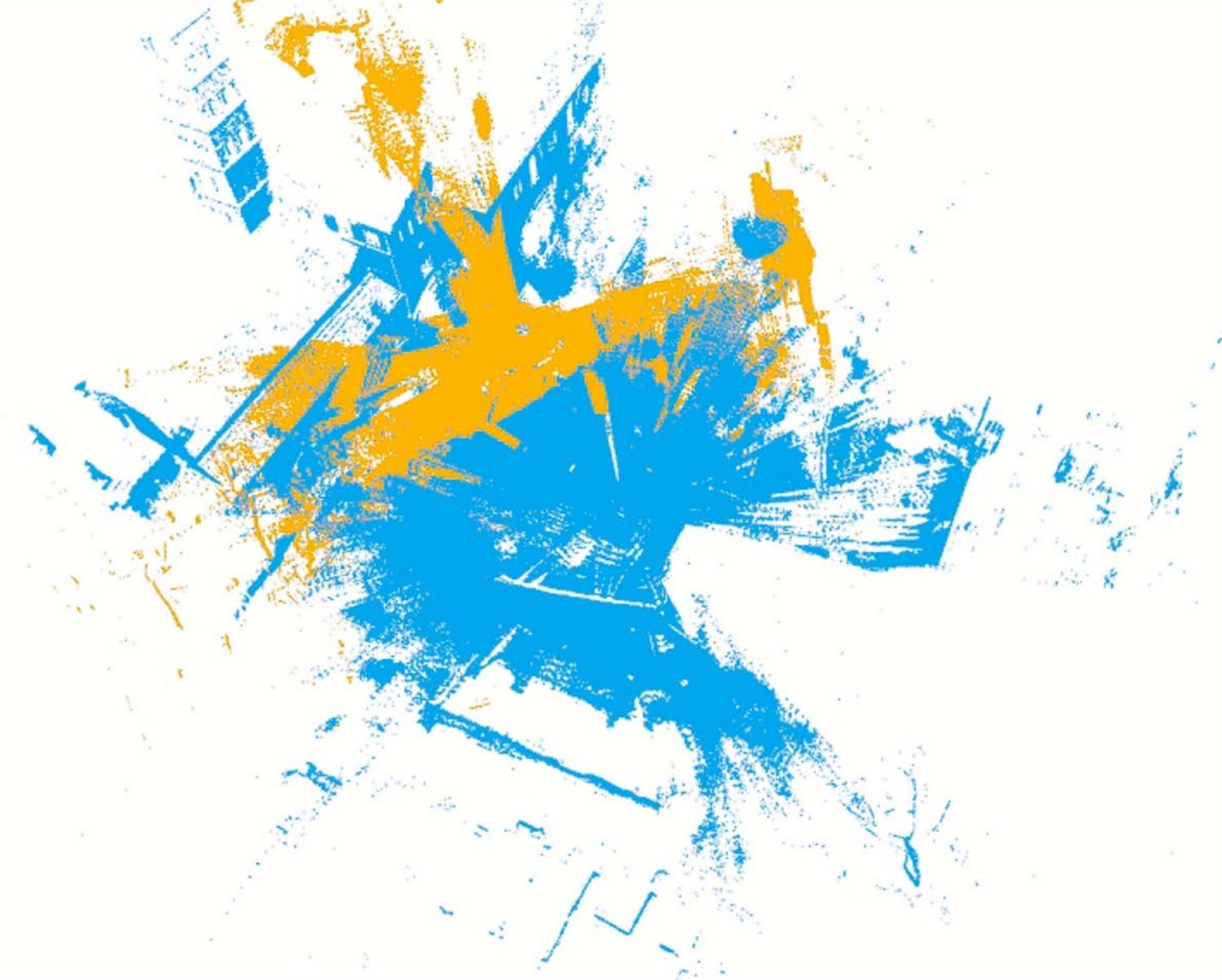} \\ (b-2) \textsf{RANSAC}~($\xi = 4$) \\ $\textit{E}_\mathbf{R}$ = 144.14$^{\circ}$, \ $\textit{E}_\mathbf{t}$ = 1877.96cm \\ t = 0.42s~(\textcolor{red}{Fail}) } &
      \makecell{\includegraphics[width=0.10\textwidth]{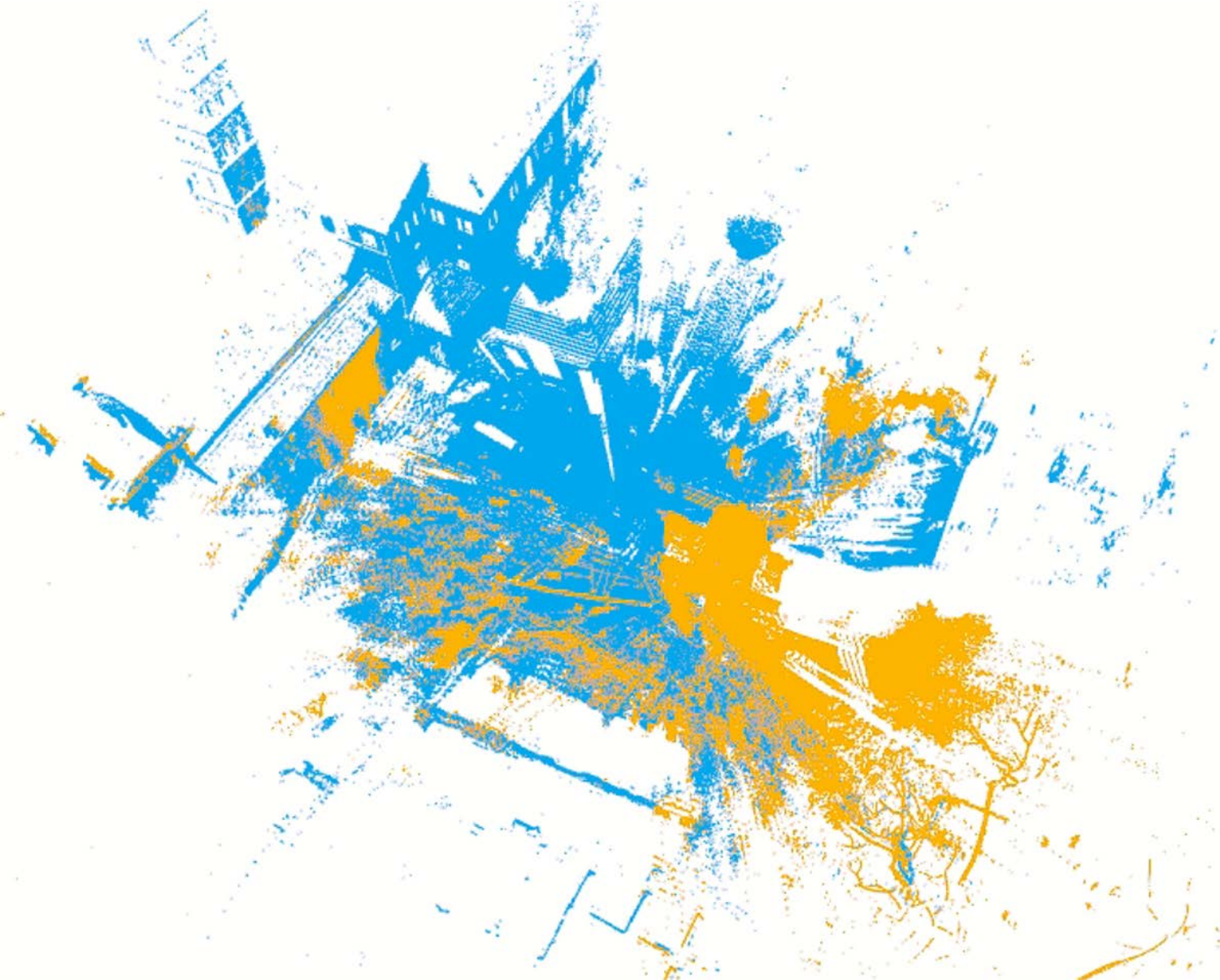} \\ (c-1) \textsf{TEASER++}~($\xi = 0.04$) \\ $\textit{E}_\mathbf{R}$ = 0.22$^{\circ}$, \ $\textit{E}_\mathbf{t}$ = 5.41cm \\ t = 0.93s~(\textcolor{blue}{Success}) \\[0.5em]  \includegraphics[width=0.10\textwidth]{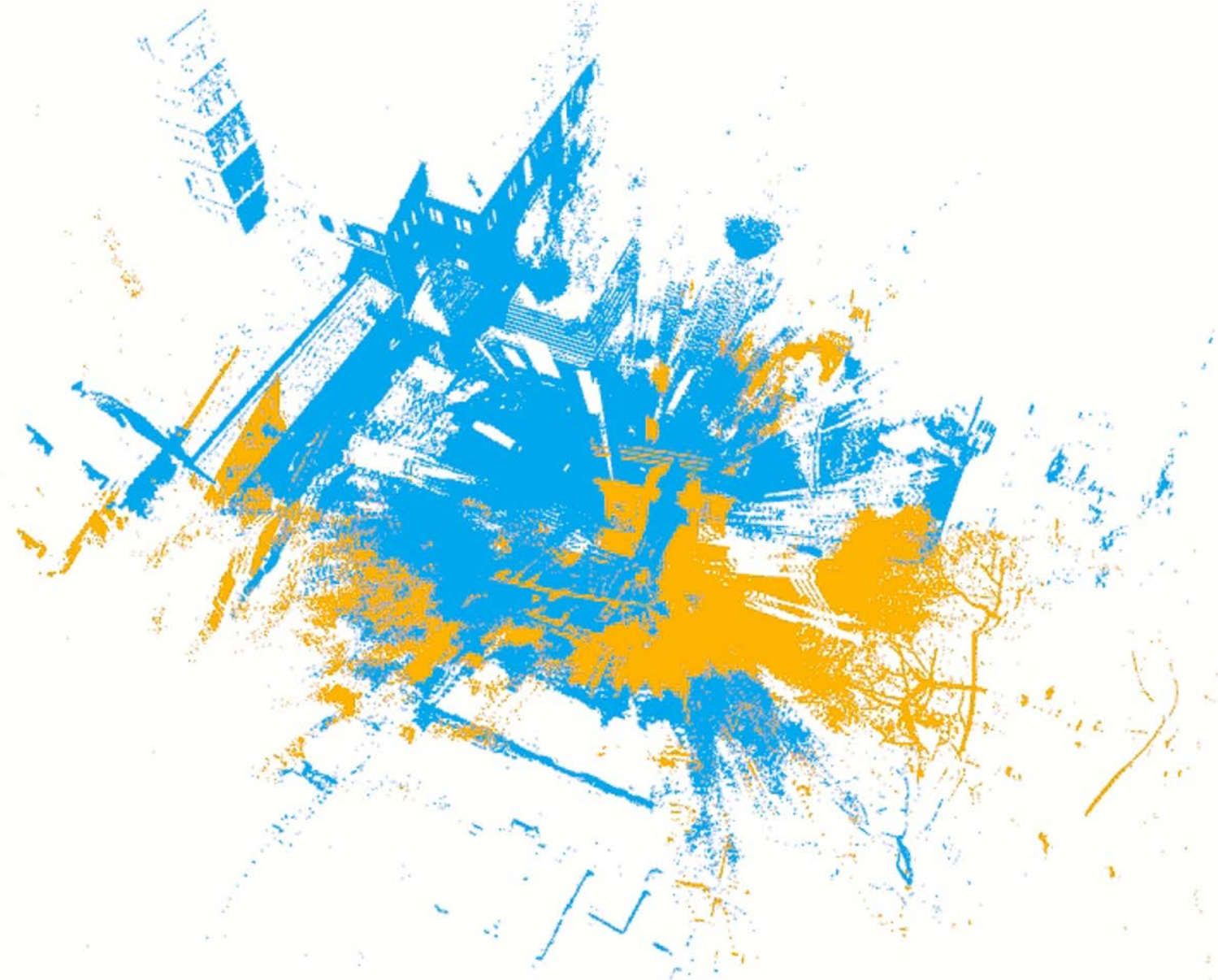} \\ (c-2) \textsf{TEASER++}~($\xi = 4$) \\ $\textit{E}_\mathbf{R}$ = 14.06$^{\circ}$, \ $\textit{E}_\mathbf{t}$ = 367.77cm \\ t = 8.73s~(\textcolor{red}{Fail}) } &
      \makecell{\includegraphics[width=0.10\textwidth]{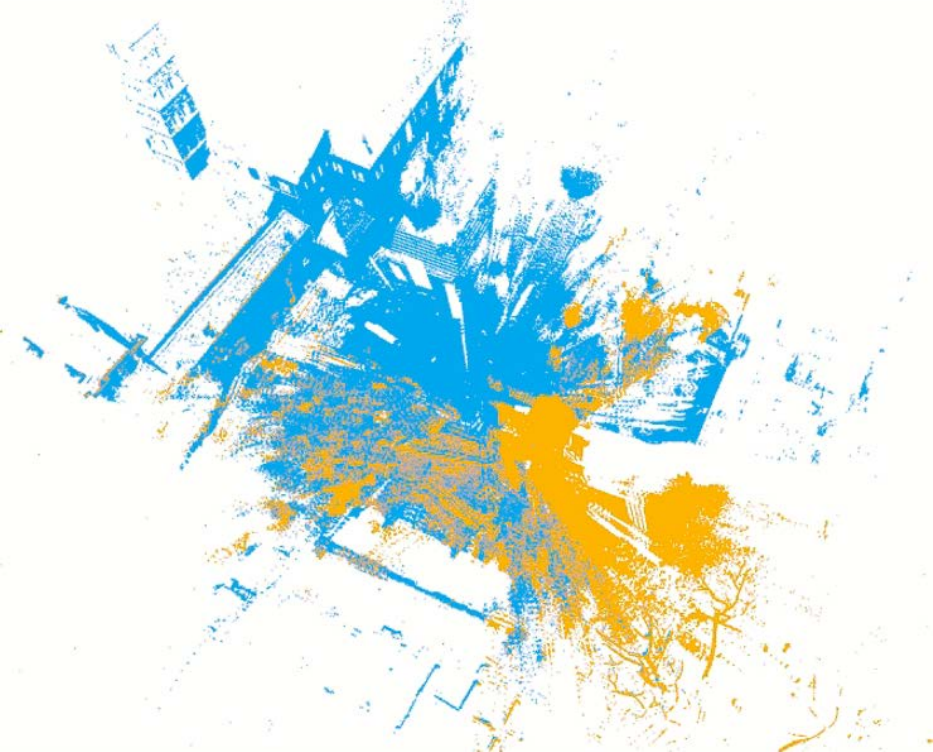}  \\ (d-1) \textsf{MAC}~($\xi = 0.04$) \\ $\textit{E}_\mathbf{R}$ = 0.18$^{\circ}$, \ $\textit{E}_\mathbf{t}$ = 6.32cm \\ t = 21.37s~(\textcolor{blue}{Success}) \\[0.5em] \includegraphics[width=0.10\textwidth]{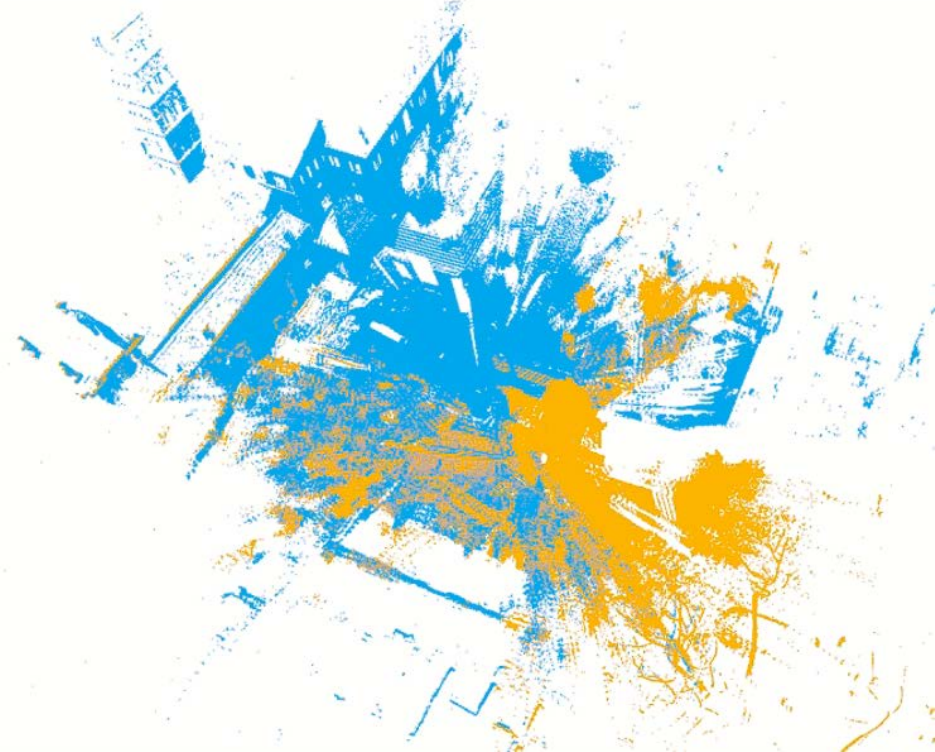} \\ (d-2) \textsf{MAC}~($\xi = 4$) \\ $\textit{E}_\mathbf{R}$ = 0.84$^{\circ}$, \ $\textit{E}_\mathbf{t}$ = 16.71cm \\ t = 38.42s~(\textcolor{blue}{Success}) } &
      \makecell{\includegraphics[width=0.10\textwidth]{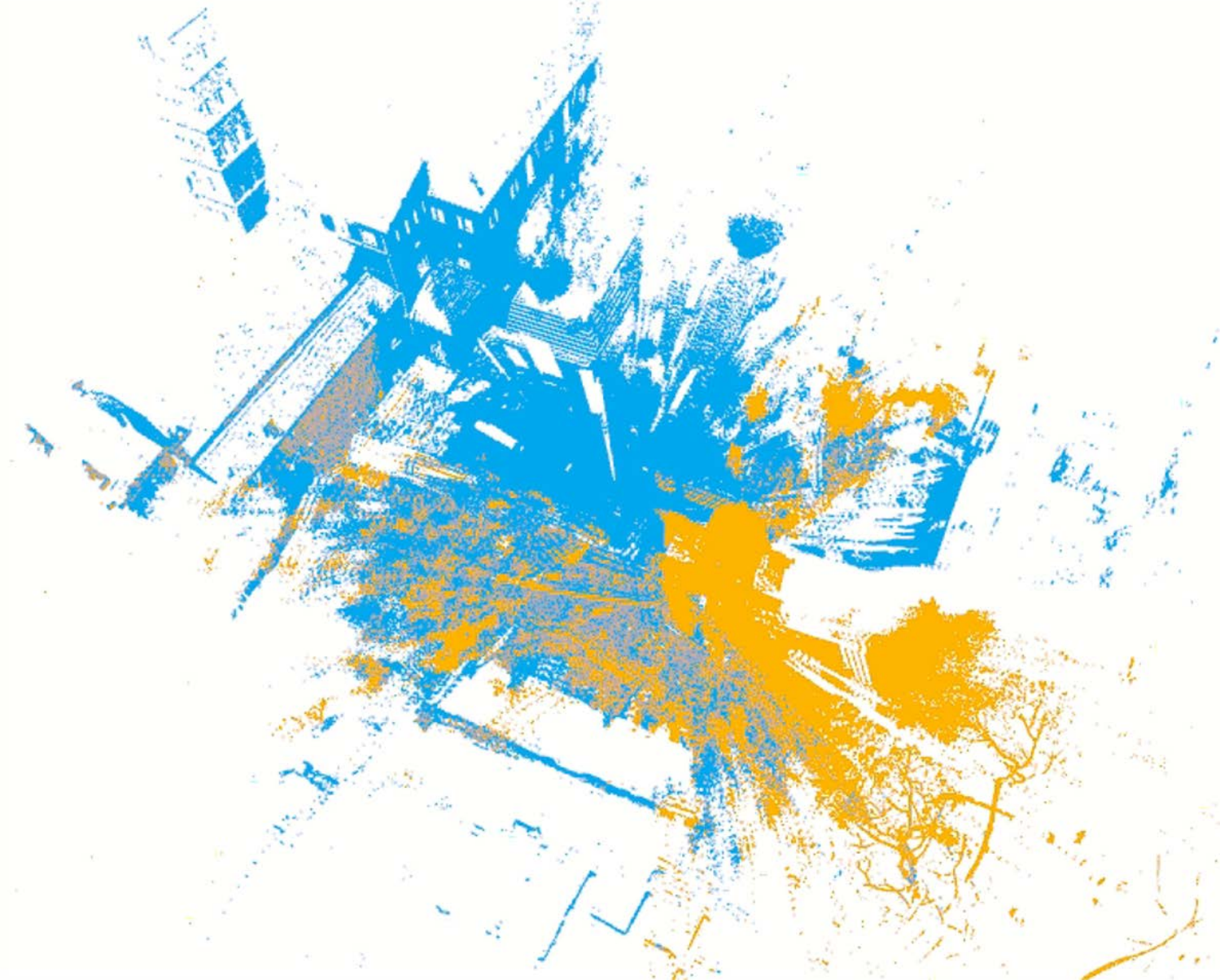}  \\ (e-1) \textsf{GTM}~($\xi = 0.04$) \\ $\textit{E}_\mathbf{R}$ = 0.25$^{\circ}$, \ $\textit{E}_\mathbf{t}$ = 4.96cm \\ t = 1.86s~(\textcolor{blue}{Success}) \\[0.5em] \includegraphics[width=0.10\textwidth]{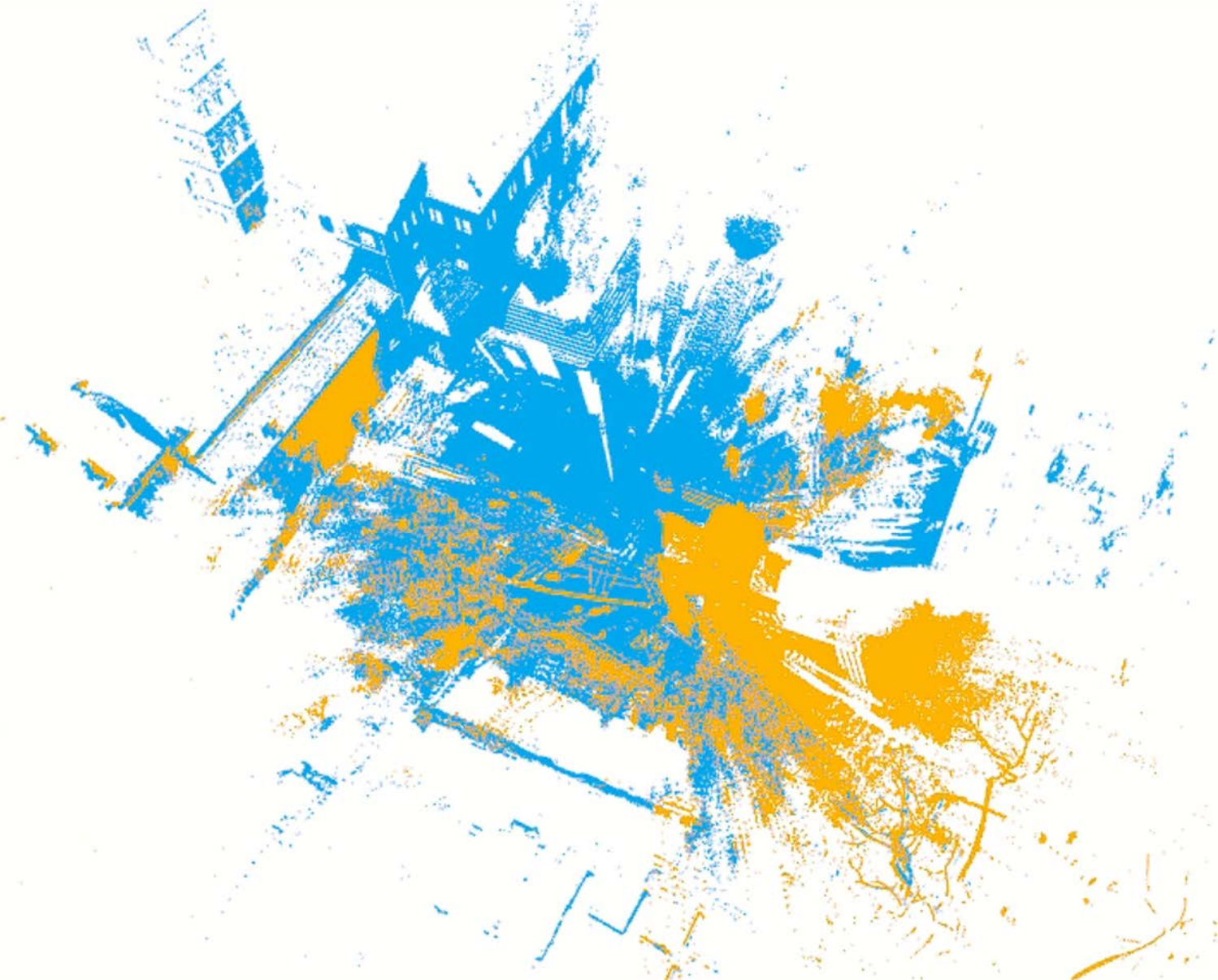} \\ (e-2) \textsf{GTM}~($\xi = 4$) \\ $\textit{E}_\mathbf{R}$ = 0.42$^{\circ}$, \ $\textit{E}_\mathbf{t}$ = 11.03cm \\ t = 2.77s~(\textcolor{blue}{Success}) } \\ 
      [-0.5em]
      \\
      \makecell{\includegraphics[width=0.08\textwidth]{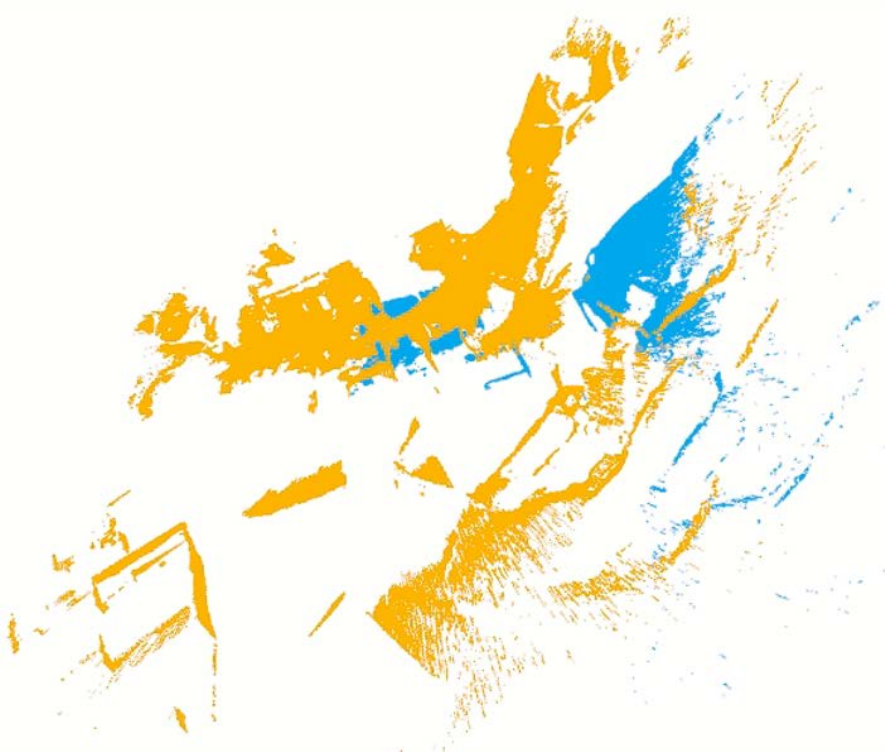} \\ (f) Inputs: \\ Point Pairs - 5247 \\ \# of Inliers - 168} &
      \makecell{\includegraphics[width=0.08\textwidth]{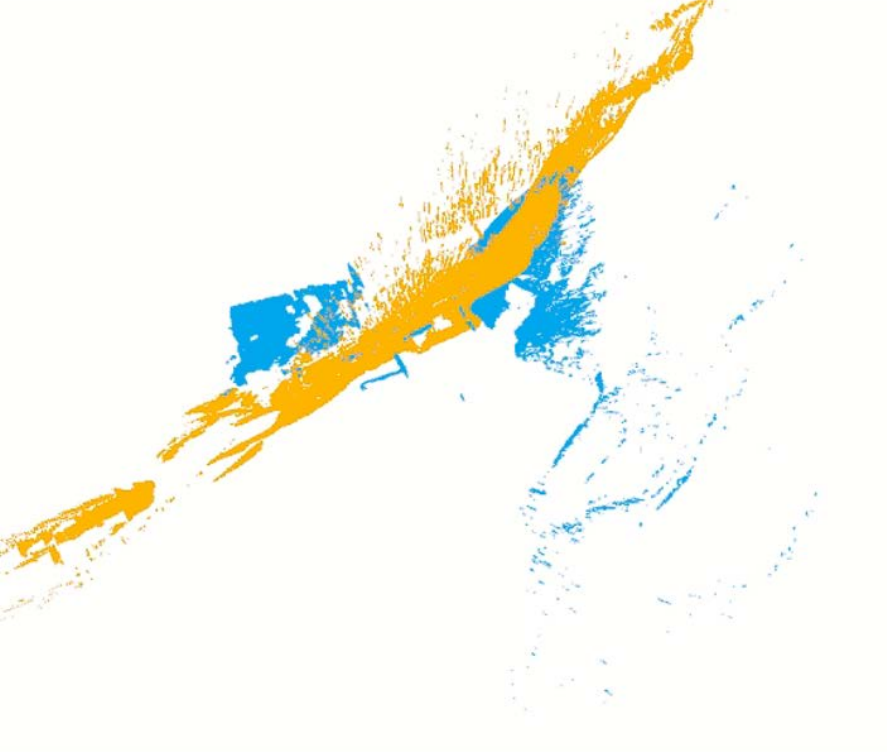} \\ (g-1) \textsf{MLESAC}~($\xi = 0.04$) \\ $\textit{E}_\mathbf{R}$ = 39.76$^{\circ}$, \ $\textit{E}_\mathbf{t}$ = 949.68cm \\ t = 0.16s~(\textcolor{red}{Fail}) \\[0.5em] \includegraphics[width=0.08\textwidth]{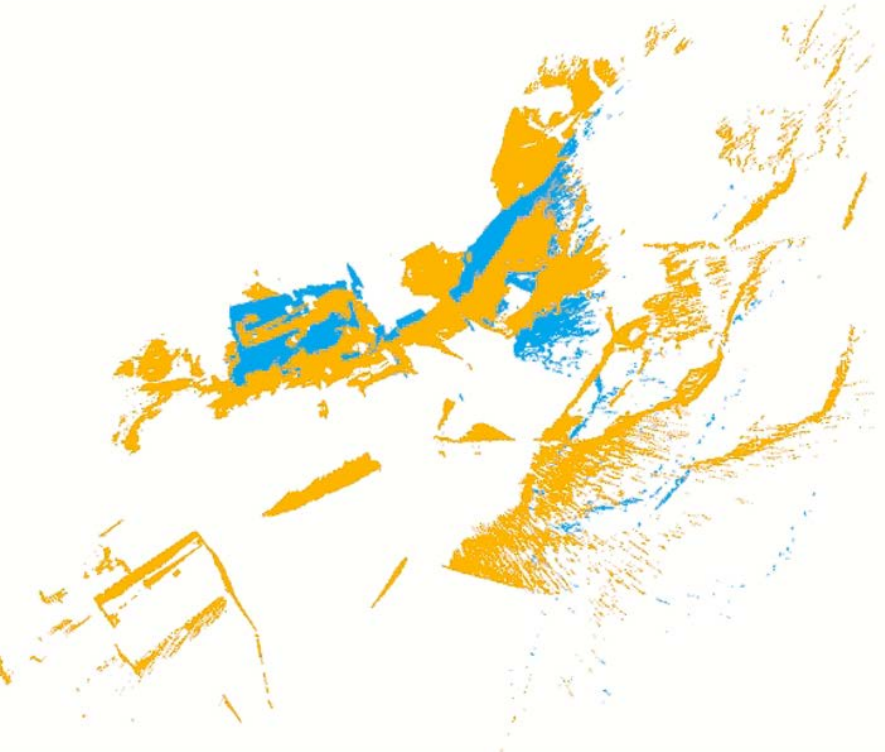} \\ (g-2) \textsf{MLESAC}~($\xi = 4$) \\ $\textit{E}_\mathbf{R}$ = 14.78$^{\circ}$, \ $\textit{E}_\mathbf{t}$ = 391.34cm \\ t = 0.15s~(\textcolor{red}{Fail}) } &
      \makecell{\includegraphics[width=0.08\textwidth]{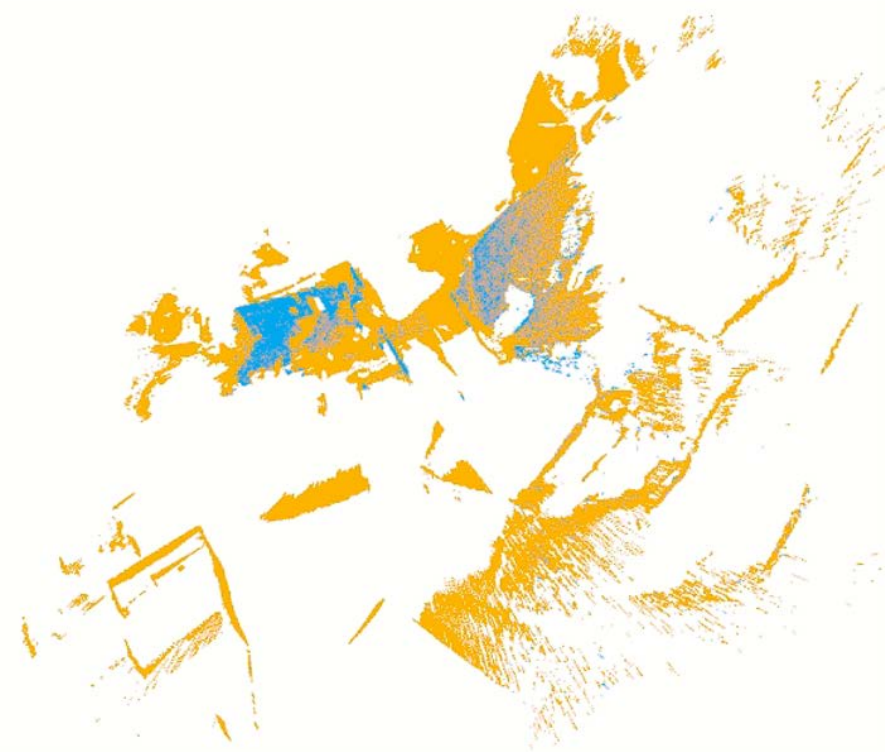} \\ (h-1) \textsf{MAC}~($\xi = 0.04$) \\ $\textit{E}_\mathbf{R}$ = 0.14$^{\circ}$, \ $\textit{E}_\mathbf{t}$ = 6.97cm \\ t = 17.31s~(\textcolor{blue}{Success}) \\[0.5em] \includegraphics[width=0.08\textwidth]{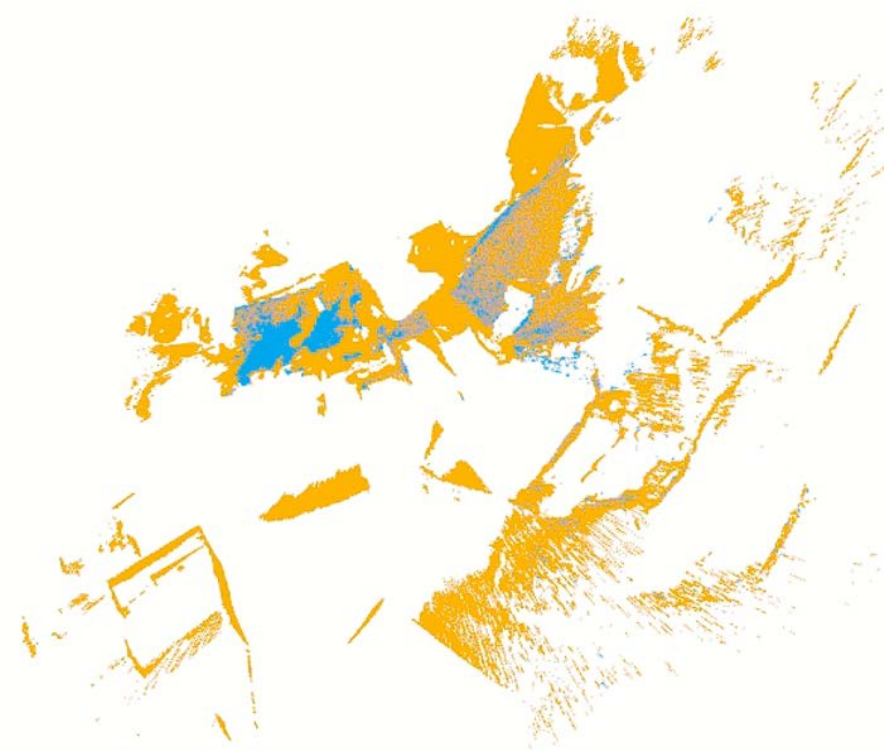} \\ (h-2) \textsf{MAC}~($\xi = 4$) \\ $\textit{E}_\mathbf{R}$ = 0.33$^{\circ}$, \ $\textit{E}_\mathbf{t}$ = 11.82cm \\ t = 22.65s~(\textcolor{blue}{Success}) } &
      \makecell{\includegraphics[width=0.08\textwidth]{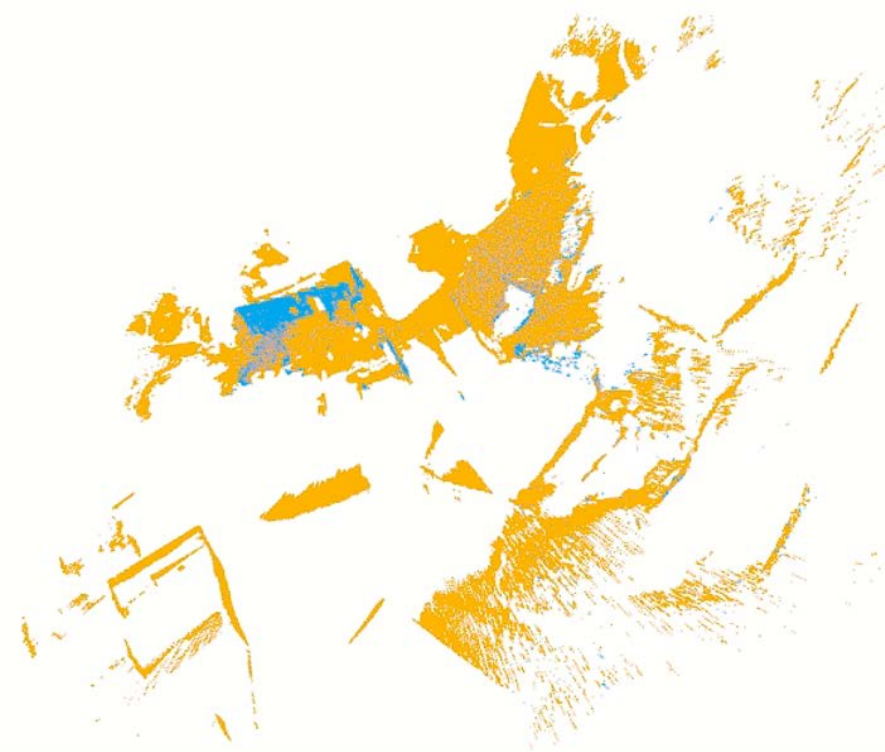} \\ (i-1) \textsf{HERE}~($\xi = 0.04$) \\ $\textit{E}_\mathbf{R}$ = 0.10$^{\circ}$, \ $\textit{E}_\mathbf{t}$ = 5.98cm \\ t = 1.09s~(\textcolor{blue}{Success}) \\[0.5em]  \includegraphics[width=0.08\textwidth]{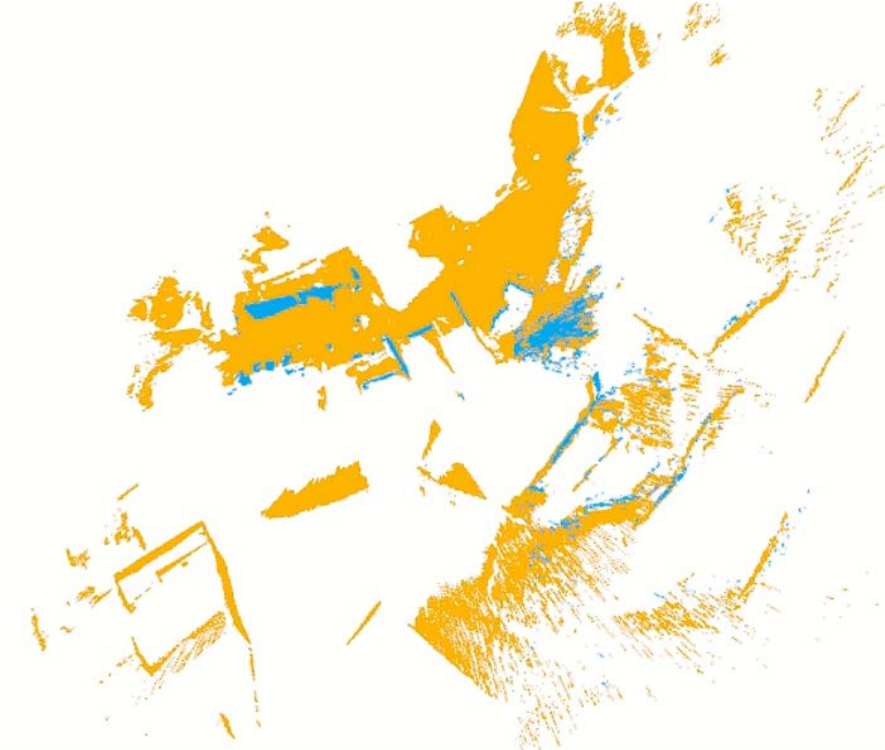} \\ (i-2) \textsf{HERE}~($\xi = 4$) \\ $\textit{E}_\mathbf{R}$ = 8.73$^{\circ}$, \ $\textit{E}_\mathbf{t}$ = 265.91cm \\ t = 1.54s~(\textcolor{red}{Fail}) } &
      \makecell{\includegraphics[width=0.08\textwidth]{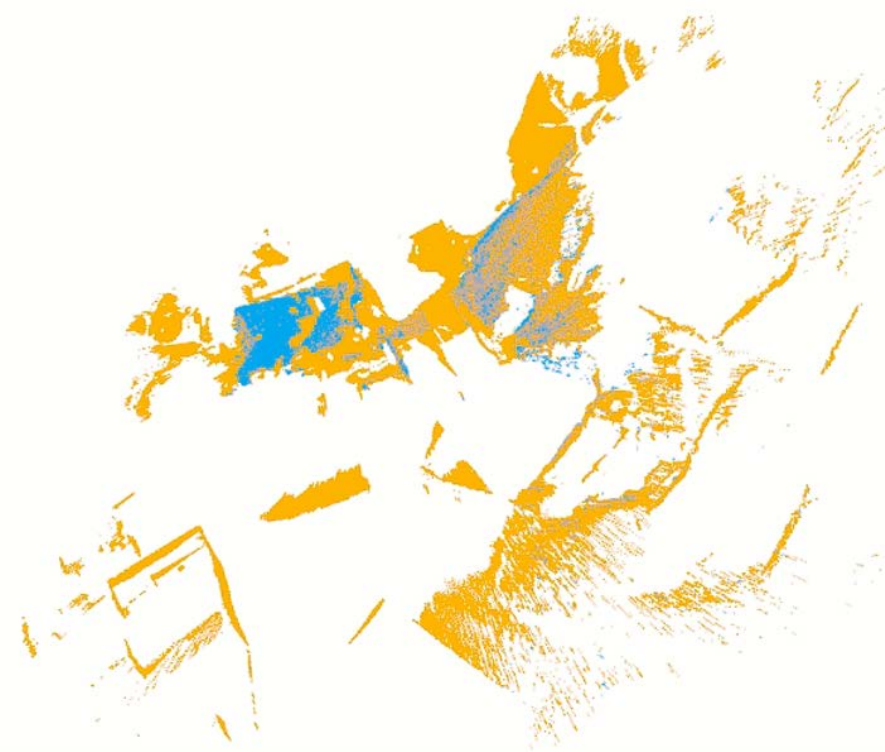}  \\ (j-1) \textsf{GTM}~($\xi = 0.04$) \\ $\textit{E}_\mathbf{R}$ = 0.07$^{\circ}$, \ $\textit{E}_\mathbf{t}$ = 6.69cm \\ t = 1.24s~(\textcolor{blue}{Success}) \\[0.5em] \includegraphics[width=0.08\textwidth]{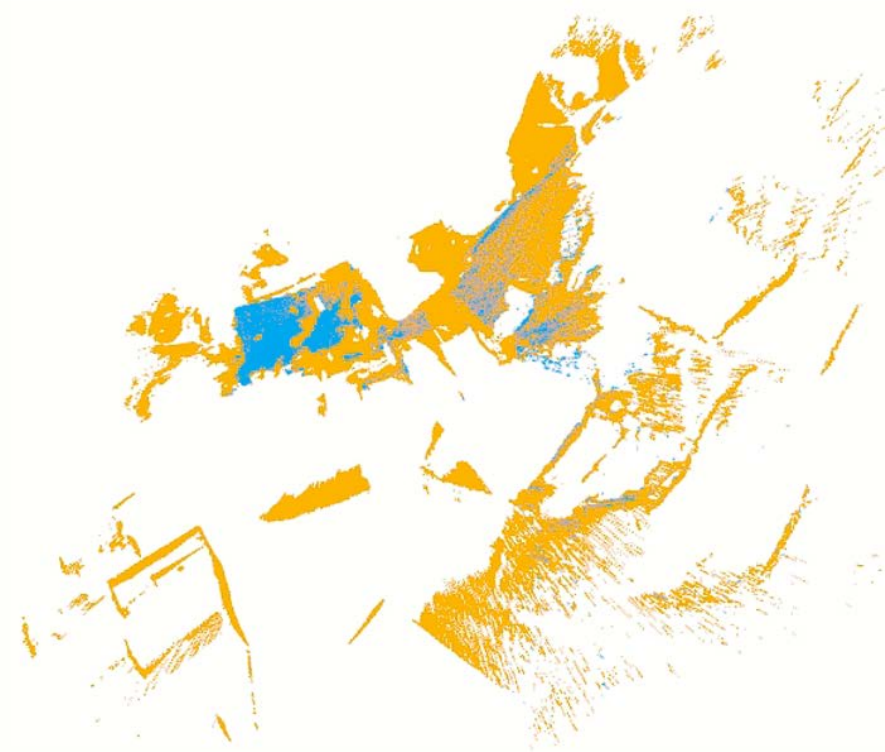} \\ (j-2) \textsf{GTM}~($\xi = 4$) \\ $\textit{E}_\mathbf{R}$ = 0.26$^{\circ}$, \ $\textit{E}_\mathbf{t}$ = 8.31cm \\ t = 1.78s~(\textcolor{blue}{Success}) } \\
    \end{tabular}
    \vspace{-0.6em}
    \caption{Two representative visual cases of 6-DoF rigid pose estimation on the real-world ETH dataset~(c.f.~Section~\ref{sec:3dest-real}). The first column presents the input point pairs and the other columns show the outputs of various methods. Only \textsf{MAC} and our \textsf{GTM} succeed in all cases, and \textsf{GTM} is $10\times$ faster than \textsf{MAC} with comparable accuracy.}
    \label{fig:app2_real_rep}
\end{figure*}

\subsubsection{Experiments on Real-world Data}
\label{sec:3dest-real}
To evaluate the performance of estimating the whole 6-DoF rigid transformation, we conduct experiments on the real-world ETH dataset~\cite{Theiler-ISPRS2014} containing large-scale point clouds~(\textit{exceeding hundreds of meters}) on five outdoor scenes~(see the appendix for details). Here we do not compare \textsf{ACM} and \textsf{DIRECT-3D} as they do not estimate the rotation. Instead, we benchmark \textsf{GTM} together with three SoTA methods for robust rigid point cloud registration, i.e., \textsf{TEASER++}~\cite{Yang-TRO2020}, \textsf{MAC}~\cite{Yang-TPAMI2024}, and \textsf{HERE}~\cite{Huang-TPAMI2024}. 
\textsf{TEASER++} first invokes a maximum clique algorithm that approximates the CM solution and then runs an iterative refinement to minimize a truncated least-squares loss. \textsf{MAC} extends the maximum clique formulation into computing multiple maximal cliques. \textsf{HERE} relies on combining BnB and heuristic sampling to efficiently optimize the CM. As all three above methods are accelerated by shared-memory parallelism, in this experiment we parallelize the BnB search of \textsf{GTM}. 
We set the subset size for 6-DoF pose as $2\%$ of the original point pair set in \textsf{GTM}. 
Moreover, \textsf{RANSAC}, \textsf{MLESAC}, and the standalone \textsf{FGR} method are also compared. 
We consider two threshold setups: $\xi=0.04$m$^2$ and $\xi=4$m$^2$~(we observe that most methods perform the best with $\xi=0.04$m$^2$). 
Notably, the residual functions of various methods may vary and result in slightly different threshold definition, for example, some methods employ $\sqrt{\xi}$ as the threshold; this does not influence the evaluation as the basic rigid constraint in \eqref{eq:rigid_const} is the same among all the methods. 
Apart from the estimation error, we evaluate the success rate of various methods, i.e., the percentage of successful estimation with rotation error $\textit{E}_\mathbf{R} < 3^{\circ}$ and translation error $\textit{E}_\mathbf{t} < 0.5$m.

\noindent \textbf{Data Pre-processing}.
We follow~\cite{Theiler-ISPRS2014} to process the raw point clouds in the dataset for putative 3D-3D correspondences. First, the dense point clouds are down-sampled with a 0.1m voxel grid filter. Next, ISS~\cite{Zhong-ICCVW2009} and FPFH~\cite{Rusu-ICRA2009} are adopted to extract the point feature descriptors. Then, we establish the correspondence set based on the top-5 nearest neighbor search in the space of feature descriptors. Due to low overlapping ratios between each pair of point clouds, the average outlier ratio of this dataset is higher than 95$\%$. 

\noindent \textbf{Results}.
Table~\ref{tab:app2_real_eval} presents the success rate, rotation error~($\textit{E}_\mathbf{R}$), translation error~($\textit{E}_\mathbf{t}$), and timing results of each method with different threshold setups tested on different scenes in the ETH dataset. When $\xi = 0.04$m$^2$, \textsf{TEASER++}, \textsf{HERE}, and  \textsf{GTM} achieve the highest robustness with comparable accuracy. But when $\xi$ is increased to $4$m$^2$, both \textsf{TEASER++} and \textsf{HERE} become unreliable with lower success rates, while \textsf{GTM} still maintains high accuracy, demonstrating its remarkable threshold-resilience. The accuracy of \textsf{MAC} remains stable in light of the increasing threshold as well, thanks to its use of multiple maximal cliques. However, the number of maximal cliques grows with the thresholds, which drastically increases the running time of \textsf{MAC} and makes it 5-15 times slower than \textsf{GTM}. Fig.~\ref{fig:app2_real_rep} presents a visual comparison, where only \textsf{GTM} and \textsf{MAC} succeed in all cases, with \textsf{GTM} 10 times faster on average.


\section{Application~3: Rotational Homography Estimation}
\label{sec:4dest}
Next, we focus on another interesting geometry estimation task, i.e., panoramic image stitching based on rotational homography estimation. This task holds great significance and can find applications in diverse domains, particularly in virtual reality~\cite{Anderson-TOG2016}. As illustrated in Fig.~\ref{fig:app3_ill}, given a pair of images captured by a camera rotating around its optical centre, the relative rotation and focal length of the camera are required to perform panoramic image stitching. Like the aforementioned tasks, the estimation often requires a set of outlier-contaminated 2D-2D pixel matches, while differently, here we need to estimate the 3D camera rotation and the unknown focal length, 4-DoF in total. In what follows, we show that such estimation can be achieved by constructing a 4-dimensional truncated loss minimization problem~\eqref{eq:TL-3} and solving it to global optimality by our $\textsf{GTM}$ approach, which involves applying the 3D BnB search and 1D DIRECT solver.

\subsection{Problem Formulation}
\begin{figure}[!t]
    \centering
    \includegraphics[width=0.35\textwidth]{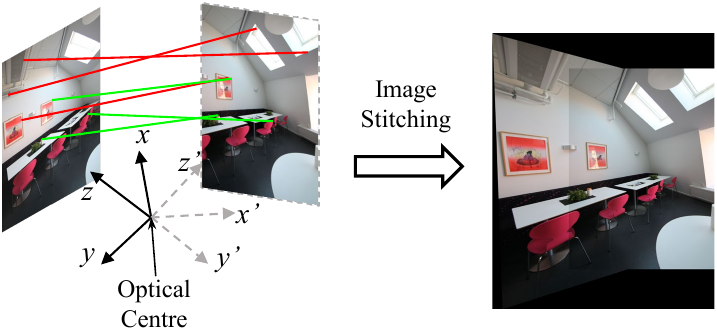}
    \\[-0.7em]
    \caption{Illustration of the rotational homography estimation problem for panoramic image stitching~(c.f.~Section~\ref{sec:4dest}). Given a set of \textcolor{red}{outlier}-corrupted 2D-2D pixel matches between two images captured by a camera rotating around its optical centre, we aim to estimate the 3-DoF relative rotation and the 1-DoF focal length of the camera.} 
    \label{fig:app3_ill}
\end{figure}

Consider a set of 2D-2D pixel matches $\{(\mathbf{z}_i, \mathbf{z}_i')\}_{i=1}^M$ between two images captured as Fig.~\ref{fig:app3_ill}, where $\mathbf{z}_i = [z_{1i},\ z_{2i}]^{\top}$ and $\mathbf{z}_i' = [z_{1i}',\ z_{2i}']^{\top}$ denote the \textit{centered} 2D pixel coordinates~\cite{Bazin-ECCV2014}. Let $\bar{\mathbf{z}}_i$,  $\bar{\mathbf{z}}_i'$ be the corresponding homogeneous coordinates as in Section.~\ref{sec:2dest_prop}.
As the camera rotates around its optical centre, there is no translation involved, and we denote the unknown rotation by $\mathbf{R}^* \in$ SO(3) and the unknown focal length by  $F^*\in\mathbb{R}$. If $(\mathbf{z}_i, \mathbf{z}_i')$ is a noiseless inlier match, then
\begin{align}\label{eq:homo_const}
    \bar{\mathbf{z}}_i' \propto \mathbf{K}^* \mathbf{R}^* {\mathbf{K}^*}^{-1} \bar{\mathbf{z}}_i,
\end{align}
where the \textit{intrinsic calibration matrix} $\mathbf{K}^*$ is of the form $\mathbf{K}^* = \textnormal{diag}([F^*, F^*, 1])$, as the image pixels are centered. 

We now proceed by deriving a residual function that is a special case of our general formulation in \eqref{eq:general-r_i}. To do so, we first represent the rotation matrix in its angular form:
\begin{align}
    \mathbf{R}^* = \mathbf{R}_{z}(\alpha^*) \mathbf{R}_{y}(\beta^*)\mathbf{R}_z(\gamma^*).
\end{align}
Here $\mathbf{R}_z(\alpha^*)$ and $\mathbf{R}_z(\gamma^*)$ are the rotations around the $z$-axis by angles $\alpha^*$ and $\gamma^*$ respectively, and $\mathbf{R}_y(\beta^*)$ the rotation around the $y-$axis by angle $\beta^*$.

Let $\omega^* := F^*\tan(\beta^*)$ and $k^* := 1/F^*$, we then follow~\cite{Bazin-ECCV2014, Zhang-TPAMI2024} to simplify the constraint in~\eqref{eq:homo_const} to:
\begin{align*}
    \mathbf{R}_z(\alpha^*)^{\top}\bar{\mathbf{z}}_i' \propto \begin{bmatrix}
        1 & 0 & \omega^* \\
        0 & \sqrt{1+\omega^{*2}k^{*2}} & 0 \\
        -\omega^*k^{*2} & 0 & 1
    \end{bmatrix} \mathbf{R}_z(\gamma^*) \bar{\mathbf{z}}_i.
\end{align*}
By homogenizing both sides of the above constraint with basic algebraic calculation (omitted here), we obtain for every $i=1,\dots,M$ the following residual $r_i(\alpha, \beta, \gamma, F) $:
\begin{align}
    r_i(\alpha, \beta, \gamma, F) = \|h_i(\alpha) + g_i(\beta, \gamma, F)\|_\infty,
\end{align}
where $\|\cdot\|_{\infty}$ denotes the $L_{\infty}$-norm; $h_i(\alpha)$, $g_i(\beta, \gamma, F)$ are now 2-dimensional vector-valued functions defined as
\begin{align*}
    h_i(\alpha) &= \begin{bmatrix}
     h_i^{(1)}(\alpha) \\
     h_i^{(2)}(\alpha)
 \end{bmatrix} = \begin{bmatrix}
        \|\bar{\mathbf{z}}_i'\|_2\cos(\alpha + \psi_i) \\
        \|\bar{\mathbf{z}}_i'\|_2\sin(\alpha + \psi_i) 
    \end{bmatrix}, \\
    g_i(\beta, \gamma, F) &= \begin{bmatrix}
     g_i^{(1)}(\beta, \gamma, F) \\
     g^{(2)}(\beta, \gamma, F)
 \end{bmatrix} = \begin{bmatrix}
    -\frac{\|\bar{\mathbf{z}}_i\|_2\cos(\gamma+\sigma_i) + \omega}{1 - \omega\|\bar{\mathbf{z}}_i\|_2\cos(\gamma+\sigma_i)k^{2}} \\
    -\frac{\|\bar{\mathbf{z}}_i\|_2\sin(\gamma+\sigma_i)\sqrt{1+\omega^{2}k^{2}}}{1 - \omega\|\bar{\mathbf{z}}_i\|_2\cos(\gamma+\sigma_i)k^{2}} 
    \end{bmatrix},
\end{align*}
where $\psi_i := \arctan(\frac{z_{2i}'}{z_{1i}'})$ and $\sigma_i := \arctan(\frac{z_{2i}}{z_{1i}})$.  
Now, the residual $r_i(\alpha, \beta, \gamma, F) $ here has already followed the general  form in \eqref{eq:general-r_i}.
Note that $\alpha$, $\beta$, and $\gamma$ all belong to $[-\frac{\pi}{2},\frac{\pi}{2}]$, and we denote the range of $F$ by $[0, \overline{F}]$ where $\overline{F}$ can be set by some large enough number. This leads us to the following truncated loss minimization problem to estimate the rotational homography and focal length:
\begin{align}\label{eq:TL-3}
    &\min_{\alpha,\beta,\gamma, F} \sum_{i=1}^M \min\{\|h_i(\alpha) + g_i(\beta, \gamma, F)\|_\infty,\ \xi\} \tag{\textcolor{red}{TL-3}} \\ 
    \text{s.t.} \quad & \alpha\in[-\frac{\pi}{2},\frac{\pi}{2}], \beta\in[-\frac{\pi}{2},\frac{\pi}{2}], \gamma\in[-\frac{\pi}{2},\frac{\pi}{2}], F\in[0,\overline{F}]. \nonumber
\end{align}

Again, $\xi$ is a predefined threshold and the residual here is a special case of   \eqref{eq:general-r_i}. Next, we develop bounding functions for \eqref{eq:TL-3}, based on which we apply \textsf{GTM} to solve this problem globally optimally.

\subsection{Bounding Functions for Solving \eqref{eq:TL-3}}
\label{sec:4d_bound}
To solve the 4-dimensional problem \eqref{eq:TL-3} based on our \textsf{GTM}, we conduct BnB search on the last three variables $\beta$, $\gamma$, and $F$; during the 3D search, given a sub-branch $\mathbb{B}_{2:4} = [\beta^l,\beta^u] \times [\gamma^l,\gamma^u] \times [F^l,F^u] \subseteq [-\frac{\pi}{2},\frac{\pi}{2}] \times [-\frac{\pi}{2},\frac{\pi}{2}] \times [0, \overline{F}]$, we construct the upper and lower bounding functions over the left variable $\alpha\in[-\frac{\pi}{2},\frac{\pi}{2}]$ as follows:

\noindent (\textbf{\textit{Upper Bound}})
Again, we follow \eqref{eq:GTM_UB} and choose the center point in $\mathbb{B}_{2:4}$, i.e., $\dot{\beta} = \frac{\beta
^l + \beta^u}{2}$, $\dot{\gamma} = \frac{\gamma^l + \gamma^u}{2}$, and $\dot{F} = \frac{F^l + F^u}{2}$ to compute the upper bound $U_{\textsf{GTM}-3}$:
\begin{equation}\label{eq:U_app3}
    \begin{aligned}
        U_{\textsf{GTM}-3} &= \min_{\alpha\in[-\frac{\pi}{2},\frac{\pi}{2}]} \sum_{i=1}^M \min\{\|h_i(\alpha) + g_i(\dot{\beta}, \dot{\gamma}, \dot{F})\|_\infty,\ \xi\} 
    \end{aligned}
\end{equation}

\noindent (\textbf{\textit{Lower Bound}})
Given $[\beta, \gamma, F]^{\top} \in \mathbb{B}_{2:4}$, we compute the range of both ${g_i^{(1)}}(\beta, \gamma, F)$ and ${g_i^{(2)}}(\beta, \gamma, F)$ in  constant time~(see \cite[Section~3.2]{Bazin-ECCV2014} for details).
We denote the corresponding bounding vectors by $\mathbf{s}_i^l$ and $\mathbf{s}_i^u$ as in \eqref{eq:LB-general-s}. 
Then, based on the derivation in steps (b) and (c) of Section~\ref{subsubsection:lb-general}, we can easily compute an underestimator $\myunderline{f}_i(\alpha)$ that lower bounds the truncated residual $f_i(\alpha, \beta, \gamma, T) := \min\{\|h_i(\alpha) + g_i(\beta, \gamma, F)\|_\infty,\ \xi\}$. Next, a lower bound of \eqref{eq:TL-3} can be derived as in \eqref{eq:GTM_LB} by
\begin{align}\label{eq:L_app3}
    L_{\textsf{GTM}-3} = \min_{\alpha\in[-\frac{\pi}{2},\frac{\pi}{2}]}\sum_{i=1}^M {\myunderline{f}_i(\alpha)}.
\end{align}

Clearly, both $h_i^{(1)}(\alpha) = \|\bar{\mathbf{z}}_i'\|_2\cos(\alpha + \psi_i)$ and $h_i^{(2)}(\alpha) = \|\bar{\mathbf{z}}_i'\|_2\sin(\alpha + \psi_i)$ here are Lipschitz continuous for all $1\leq i \leq M$. Therefore, Proposition~\ref{prop:Lips_inher} implies that, the objective functions in both \eqref{eq:U_app3} and \eqref{eq:L_app3} are Lipschitz continuous. We again apply the DIRECT method within a BnB framework compute both bounds globally optimally.


\subsection{Evaluation Experiments}
Below, we compare \textsf{GTM} and baseline methods for rotational homography estimation on both simulated and real-world datasets.
The compared methods include: \textsf{RANSAC} and \textsf{MLESAC} with the minimal solver of~\cite{Brown-CVPR2007}; \textsf{DIRECT-4D} and \textsf{ACM}, where \textsf{DIRECT-4D} solves \eqref{eq:TL-3} and \textsf{ACM} solves the CM version of \eqref{eq:TL-3}. Again, the maximum iteration numbers of \textsf{RANSAC} and \textsf{MLESAC} are set to $5k$.

\begin{figure}[!t]
    \centering
    \includegraphics[width=0.45\textwidth]{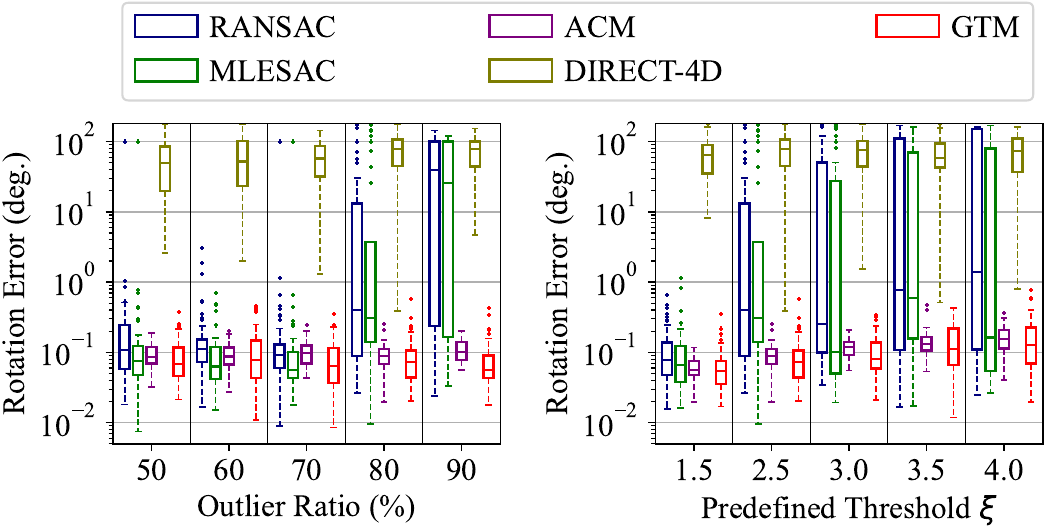}
    \\[-0.5em]
    \makebox[0.24\textwidth]{\footnotesize \quad \quad \quad \ \ (a-1)}
    \makebox[0.24\textwidth]{\footnotesize \quad \quad \ \ (b-1)}
    \\
    \includegraphics[width=0.45\textwidth]{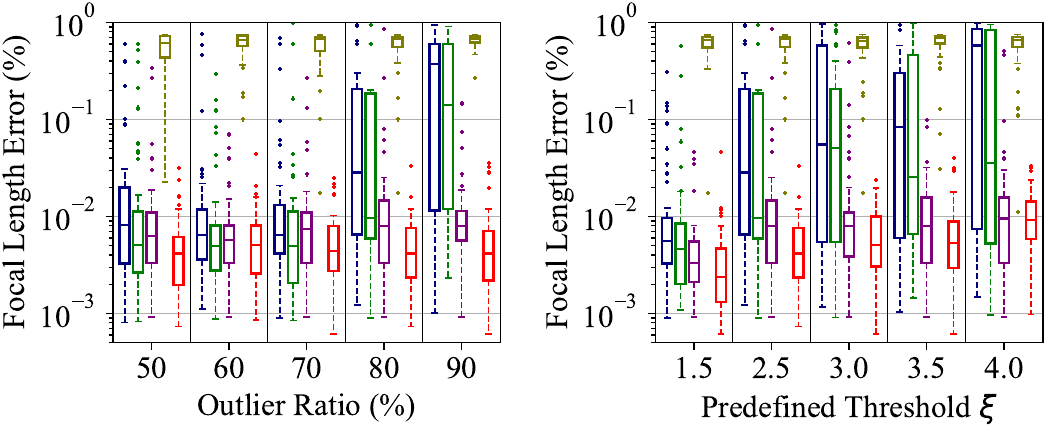}
    \\[-0.4em]
    \makebox[0.24\textwidth]{\footnotesize \quad \quad \quad \ \ (a-2)}
    \makebox[0.24\textwidth]{\footnotesize \quad \quad  \ \ (b-2)}
    \\[0.3em]
    \includegraphics[width=0.45\textwidth]{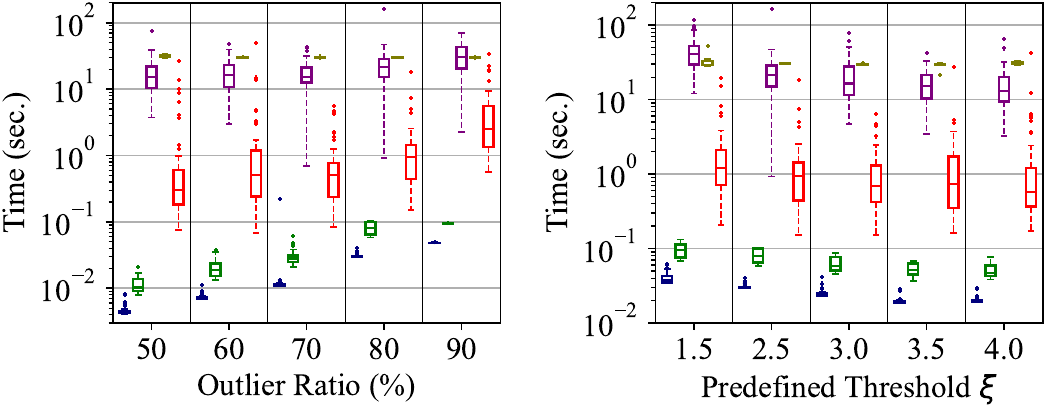}
    \\[-0.4em]
    \makebox[0.24\textwidth]{\footnotesize \quad \quad \quad \ \ (a-3)}
    \makebox[0.24\textwidth]{\footnotesize \quad \quad  \ \ (b-3)}
    \\[0.05em]
    \includegraphics[width=0.45\textwidth]{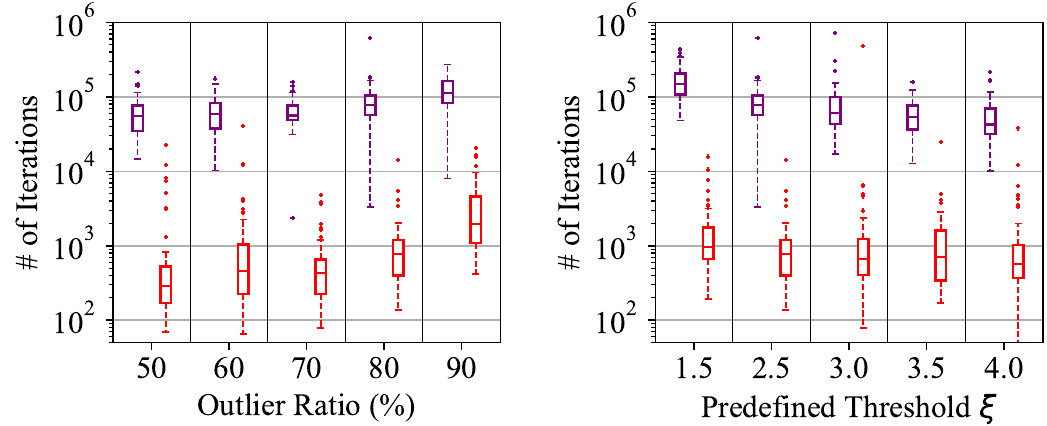}
    \\[-0.5em]
    \makebox[0.24\textwidth]{\footnotesize \quad \quad \quad \ \ (a-4)}
    \makebox[0.24\textwidth]{\footnotesize \quad \quad  \ \ (b-4)}
    \vspace{-2em}
    \caption{Evaluation resluts of application~2 on simulated datasets~(c.f.~Section~\ref{sec:4dest-sim}). (a) Increasing outlier ratio with fixed $\xi = 2.5$; (b) Increasing $\xi$ with fixed $80\%$ outlier ratio. Across various setups, only \textsf{GTM} and \textsf{ACM} maintains high robustness, meanwhile \textsf{GTM} usually runs 10-60 times faster than \textsf{ACM}.~($M = 200$, 50 trials)}
    \label{fig:app3_simu_eval}
\end{figure}

\subsubsection{Experiments on Simulated Data}
\label{sec:4dest-sim}
\noindent \textbf{Data Generation}.
First, we follow the data generation in Section~\ref{sec:2dest-sim} to generate $M = 200$ randomly sampled 3D points and then project them onto two camera frames with the same optical centre. The focal length is uniformly sampled from $[500, 1000]$, and the canvas size is set to be $1480\times2160$ pixels. Between the two frames, the rotation angles are uniformly sampled within $[-\pi/2, \pi/2]$. We also add Gaussian distributed noise with radius 0.5 to the pixel coordinates of all 2D-2D matches. To evaluate the accuracy, we consider the rotation error and the relative focal length error computed by $|\hat{F} - F^*|/F^*$, where $\hat{F}$ and $F^*$ denote the estimated and ground-truth focal length, respectively.

\noindent \textbf{Results}. From Fig.~\ref{fig:app3_simu_eval}, we observe that \textsf{GTM} is consistently the most robust across different setups. While \textsf{ACM} achieves comparable accuracy regarding rotation estimation with \textsf{GTM}, it sometimes produces inaccurate focal lengths; also, \textsf{GTM} is much faster than \textsf{ACM}, usually achieving a $10\times-60\times$ speed-up. 
The efficiency gain is because \textsf{GTM} relies on the TL loss combined with our tight bounding functions, and therefore requires much fewer iterations than \textsf{ACM} to converge~(Figs.~\ref{fig:app3_simu_eval}(a-4) and (b-4)).
In addition, despite their efficiency, \textsf{RANSAC} and \textsf{MLESAC} both fail at high outlier ratios~($\geq 80\%$) or large thresholds~($\geq 2.5$ pixels). 

Interestingly, \textsf{DIRECT-4D} fails at all setups, in contrast to its robustness in the previous two applications~(see Fig.~\ref{fig:app1_simu_eval} and Fig.~\ref{fig:app2_simu_eval}). 
This is not surprising, as DIRECT can only guarantee the global optimality when the objective function is Lipschitz continuous, but the 4-dimensional objective function in \eqref{eq:TL-3} is not, as the term $F\tan(\beta)$ in the denominator of $g_i^{(1)}(\beta, \gamma, F)$ could lead to an infinite gradient.  
In contrast, \textsf{GTM} only conducts the DIRECT method on our developed 1-dimensional bounding functions that are carefully formulated to be Lipschitz continuous, which significantly improves the robustness over \textsf{DIRECT-4D} in rotational homography estimation.

\subsubsection{Experiments on Real-world Data}
\label{sec:4dest-real}
We further conduct evaluation experiments on the real-world PASSTA image stitching dataset~\cite{Meneghetti-SCIA2015}. 

\noindent \textbf{Data Pre-processing}.
From the 72 \textit{Lunch Room} images, we randomly choose 100 image pairs where the index gap in each pair belongs to $[5, 9]$. Next, for each image pair, we extract putative correspondences via SURF descriptors~\cite{Bay-ECCV2006}. The number of matches ranges from 52 to 419 among all image pairs.
Since the ground-truth camera rotation is not available, we directly evaluate the success rates of various methods regarding image stitching results. We consider three thresholds: $\xi = 2$ pixels, $\xi = 4$ pixels, and $\xi = 6$ pixels (most methods perform the best with $\xi = 2$ pixels).

\begin{table}[!t]
\centering
    \caption{Evaluation results of application~3 on the real-world PASSTA dataset~(c.f.~Section~\ref{sec:4dest-real}). \textsf{GTM} is the most robust to thresholds $\xi$.}
    \vspace{-0.8em}
    \label{tab:app3_real_eval}
    \footnotesize
     \renewcommand{\tabcolsep}{4pt} 
    \renewcommand\arraystretch{1.1}
     \begin{tabular}{l|c|c|c|c}
         \Xhline{1pt}
         \multirow{2}{*}{Method} & $\xi = 2$ & $\xi = 4$ & $\xi = 6$ & \multirow{2}{*}{Time~(s)$\downarrow$} \\
         \cline{2-4}
         & \textit{SR}~($\%$)$\uparrow$  & \textit{SR}~($\%$)$\uparrow$ & \textit{SR}~($\%$)$\uparrow$ &  \\
         \Xhline{0.5pt}
         \textsf{RANSAC}~\cite{Fischler-CACM1981} & 91 & 81 & 74 & \textbf{0.03} \\
         \textsf{MLESAC}~\cite{Torr-CVIU2000} & 90 & 83 & 71 & 0.05 \\
         \textsf{ACM}~\cite{Zhang-TPAMI2024} & \textbf{94} & 85 & 80 & 7.24\\ 
         \textsf{DIRECT-4D}~\cite{Jones-JOTA1993} & 0 & 0 & 0 & 10.69\\
         \textsf{GTM} & \textbf{94} & \textbf{91} & \textbf{90} & 1.38 \\
         \Xhline{1pt}
     \end{tabular}
\end{table}

\begin{figure*}
    \centering
    \footnotesize
    \renewcommand{\tabcolsep}{6pt}
    \begin{tabular}{cccccc}
         &  \makecell{\textsf{RANSAC}~\cite{Fischler-CACM1981} \\ [0.5em]} & \makecell{\textsf{MLESAC}~\cite{Torr-CVIU2000} \\ [0.5em]} & \makecell{\textsf{ACM}~\cite{Zhang-TPAMI2024} \\ [0.5em]} & \makecell{\textsf{DIRECT-4D}~\cite{Jones-JOTA1993} \\ [0.5em]} & \makecell{\textsf{GTM} \\ [0.5em]}  \\ 
         \multirow{2}{*}{\makecell{\includegraphics[width=0.17\textwidth, height=0.075\textheight]{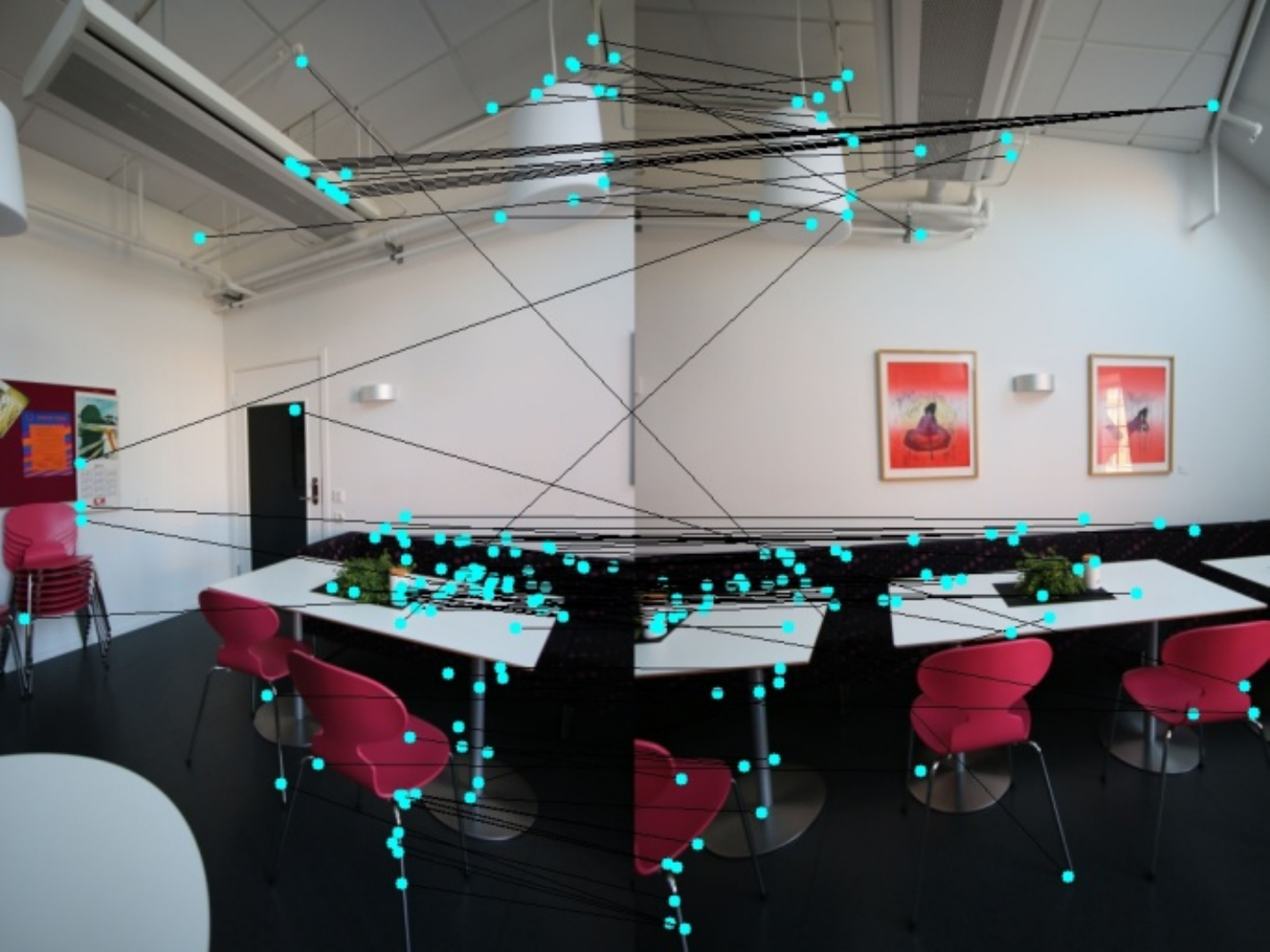} \\ (a) Match Pairs : 114}} & \makecell{\includegraphics[width=0.115\textwidth, height=0.07\textheight]{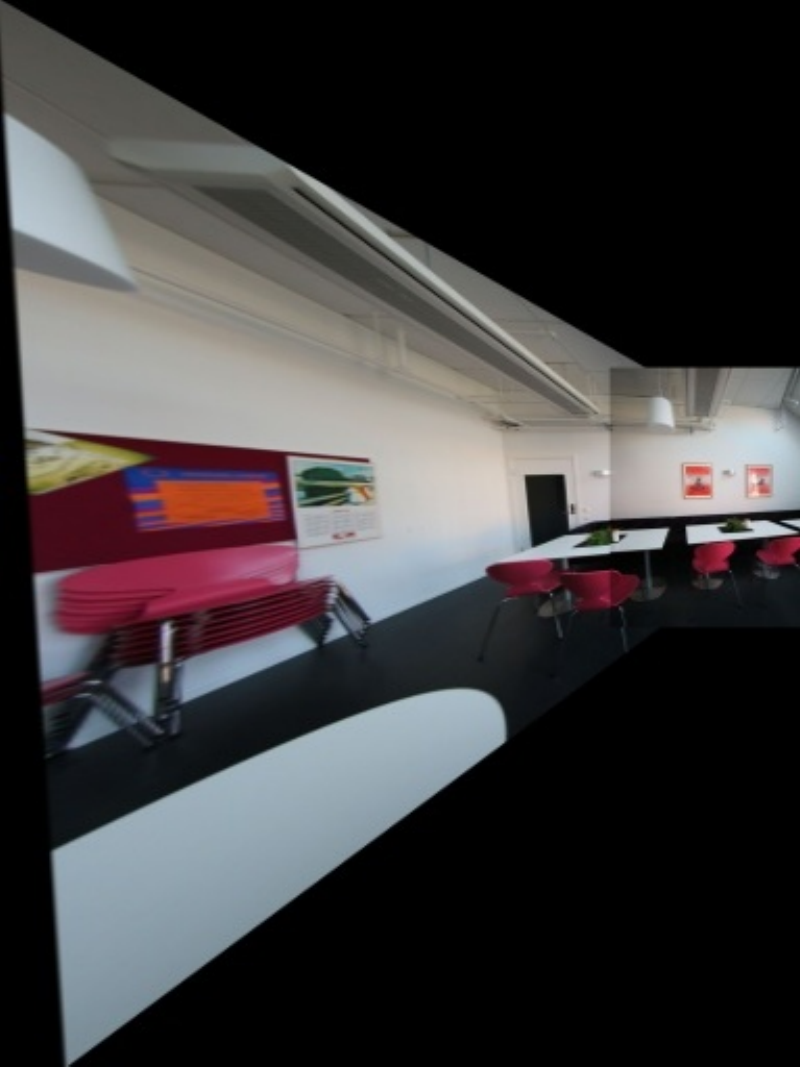} \\ (b-1) $\xi$ = 2, $t$ = 0.02s \\(\textcolor{blue}{Success}) \\ [0.5em]} & \makecell{\includegraphics[width=0.115\textwidth, height=0.07\textheight]{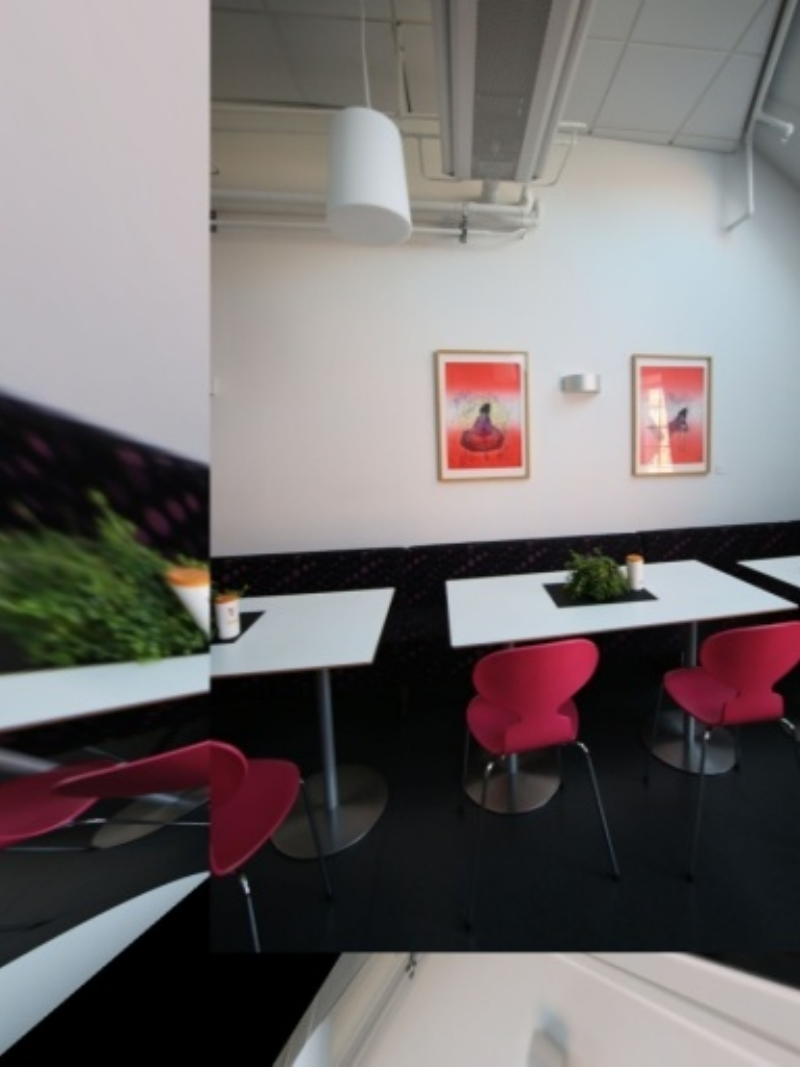} \\ (c-1) $\xi$ = 2, $t$ = 0.03s \\(\textcolor{red}{Fail}) \\ [0.5em]} & \makecell{\includegraphics[width=0.115\textwidth, height=0.07\textheight]{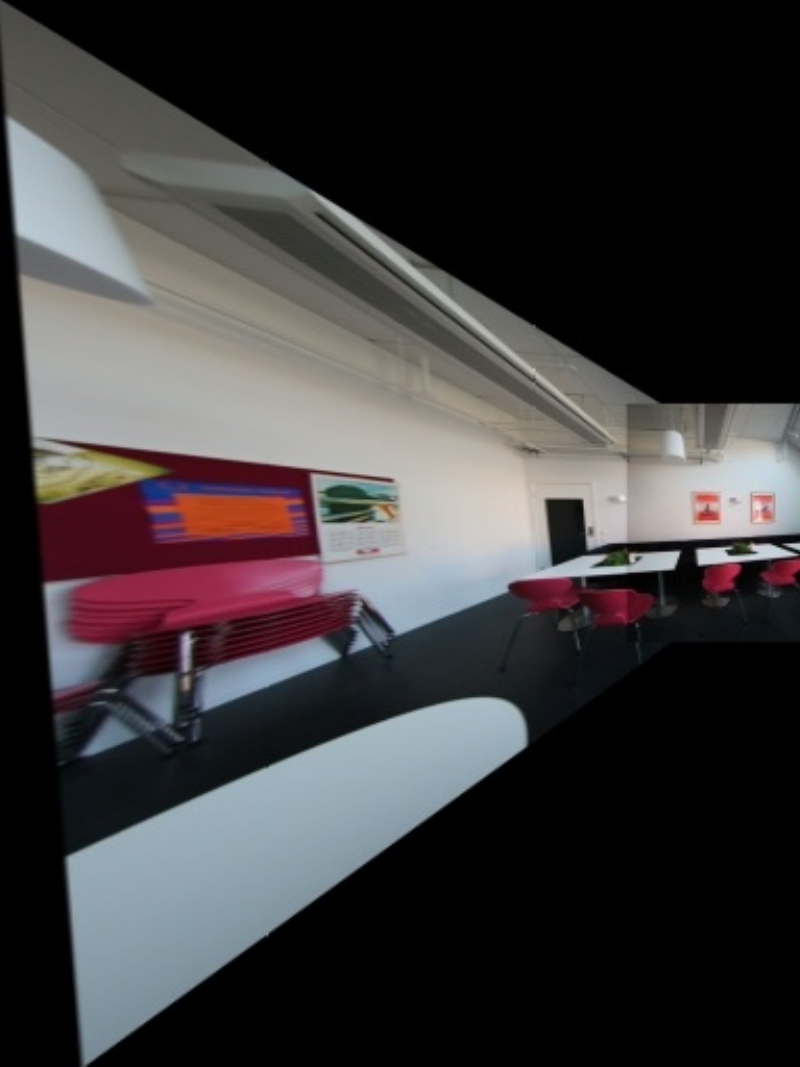}  \\ (d-1) $\xi$ = 2, $t$ = 3.94s \\(\textcolor{blue}{Success}) \\ [0.5em]} & \makecell{\includegraphics[width=0.115\textwidth, height=0.07\textheight]{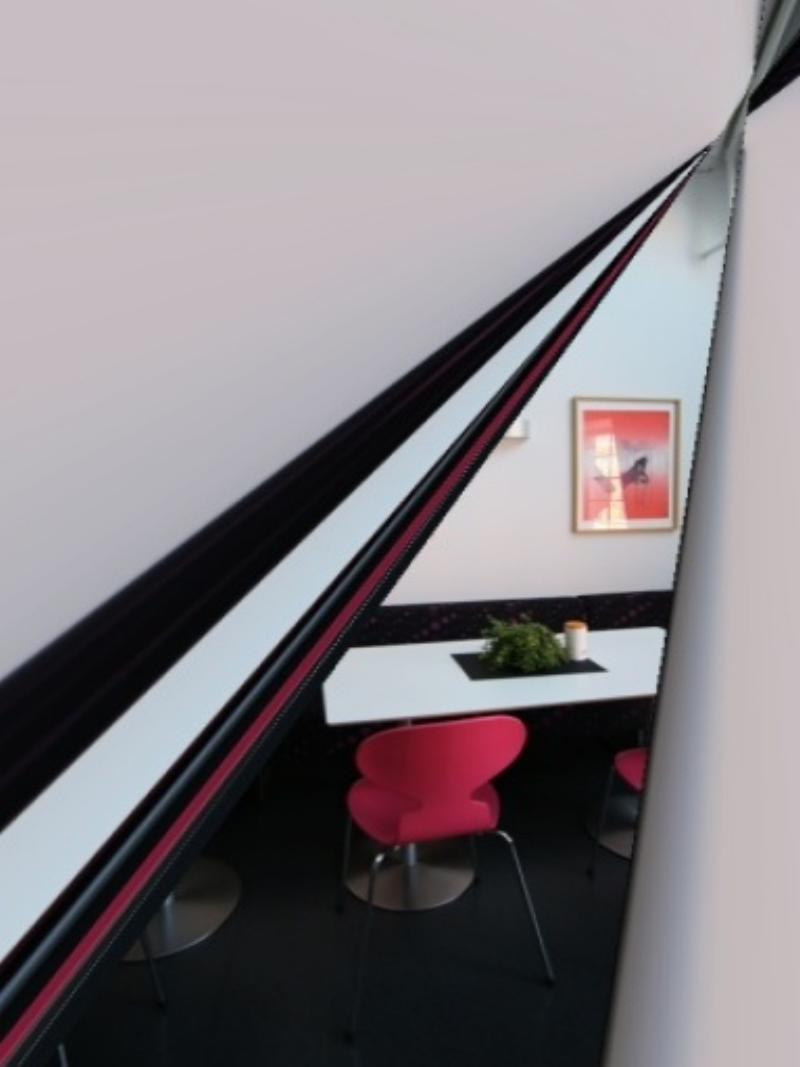} \\ (e-1) $\xi$ = 2, $t$ = 5.76s \\(\textcolor{red}{Fail}) \\ [0.5em]} & \makecell{\includegraphics[width=0.115\textwidth, height=0.07\textheight]{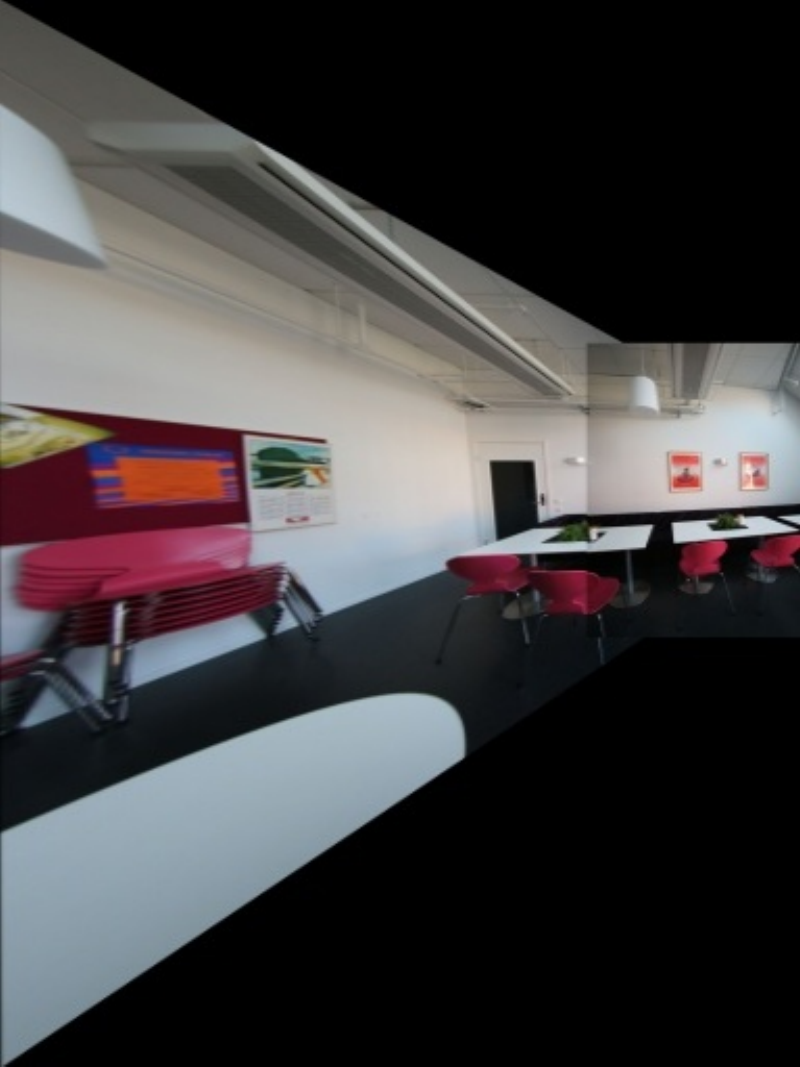} \\ (f-1) $\xi$ = 2, $t$ = 0.66s \\(\textcolor{blue}{Success}) \\ [0.4em]} \\ 
         & \makecell{\includegraphics[width=0.115\textwidth, height=0.07\textheight]{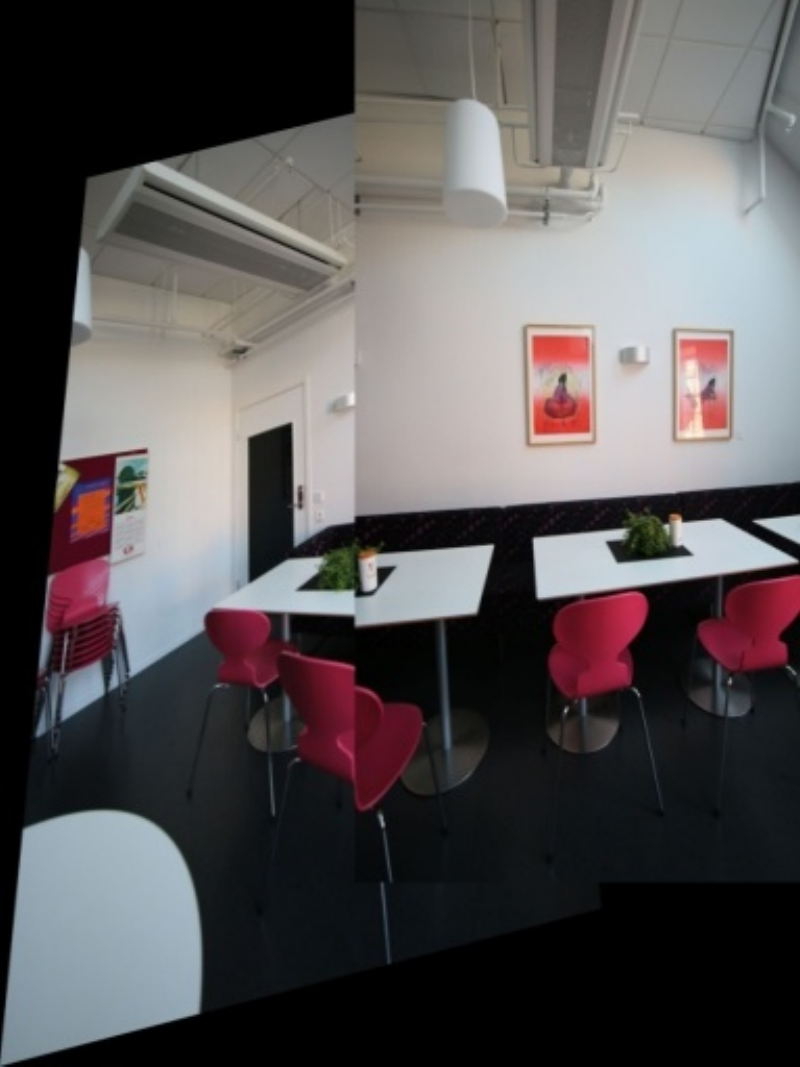} \\ (b-2) $\xi$ = 6, $t$ = 0.02s \\(\textcolor{red}{Fail})} & \makecell{\includegraphics[width=0.115\textwidth, height=0.07\textheight]{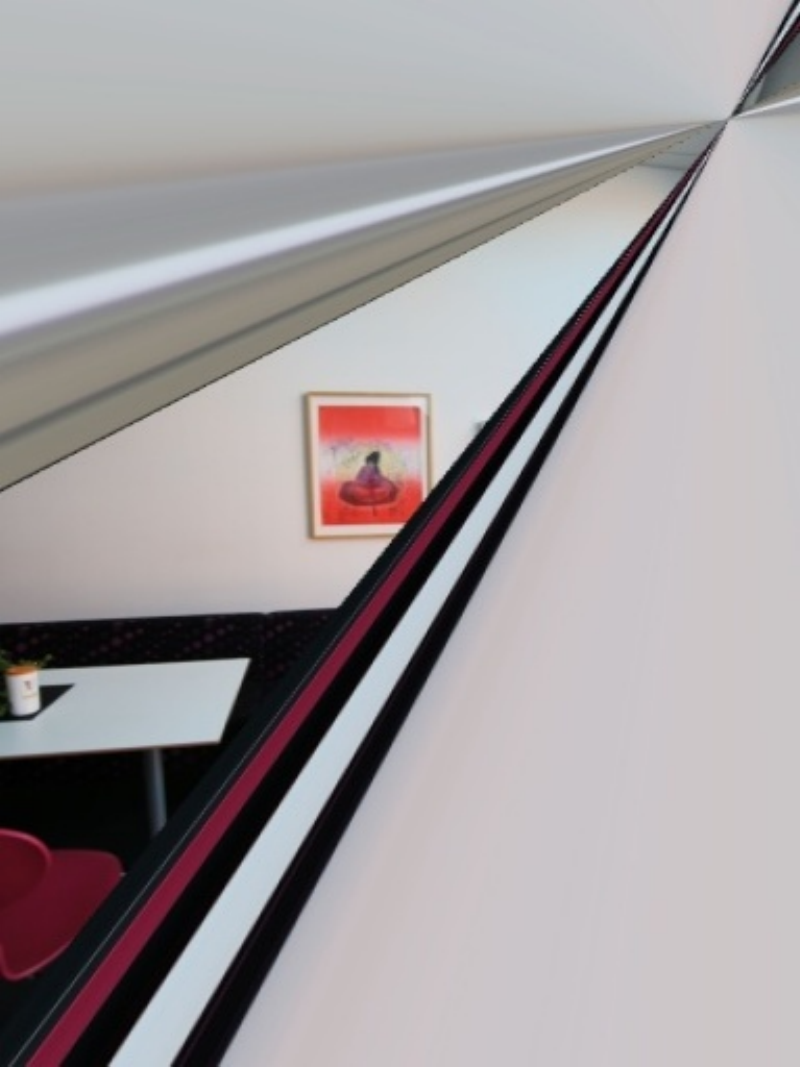} \\ (c-2) $\xi$ = 6, $t$ = 0.02s \\(\textcolor{red}{Fail}) } & \makecell{\includegraphics[width=0.115\textwidth, height=0.07\textheight]{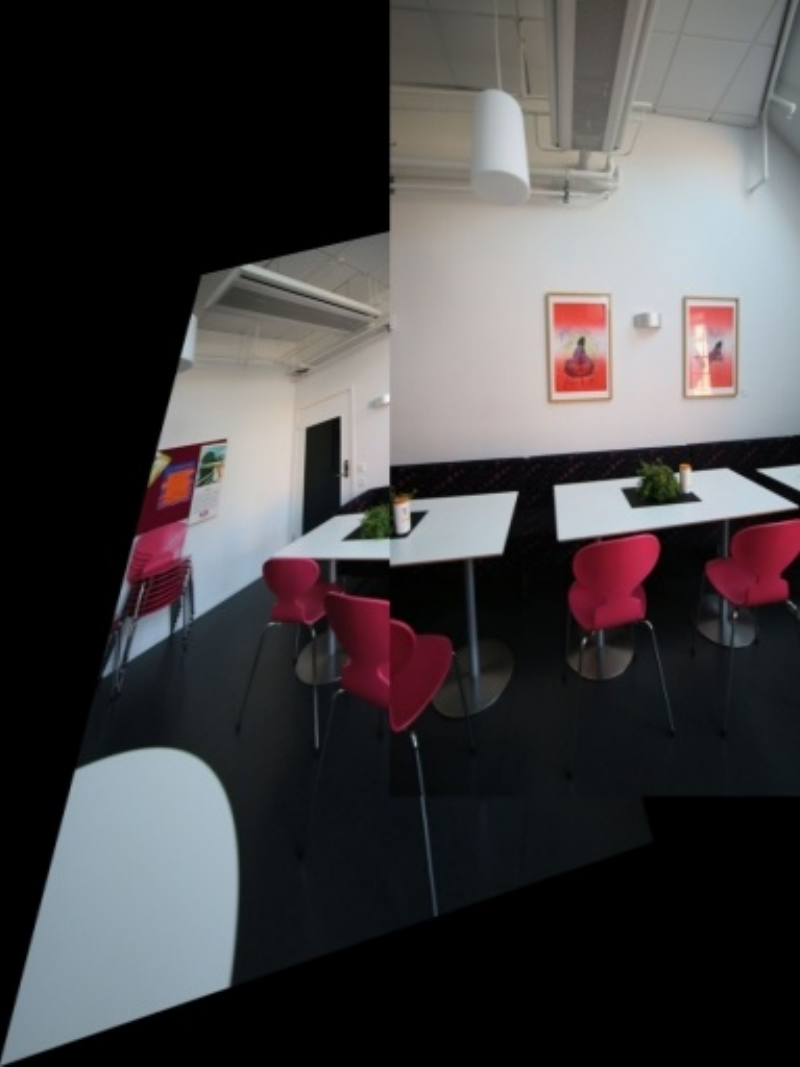}  \\ (d-2) $\xi$ = 6, $t$ = 5.33s \\(\textcolor{red}{Fail})} & \makecell{\includegraphics[width=0.115\textwidth, height=0.07\textheight]{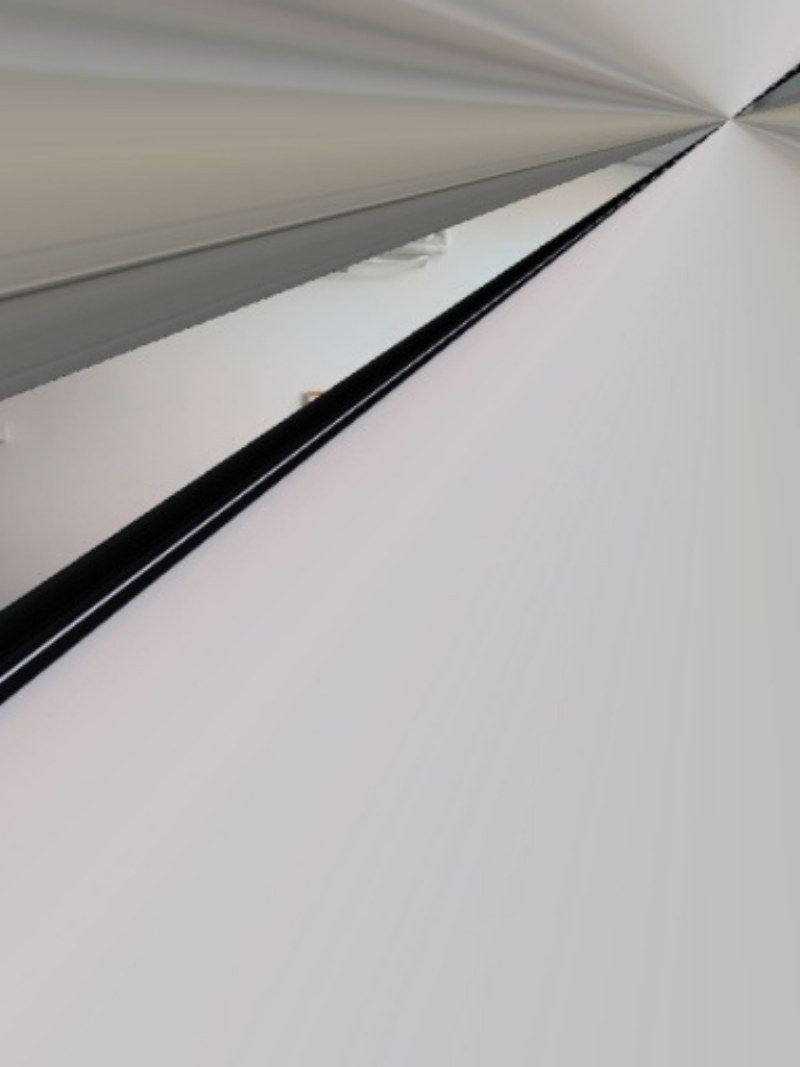} \\ (e-2) $\xi$ = 6, $t$ = 6.17s \\(\textcolor{red}{Fail}) } & \makecell{\includegraphics[width=0.115\textwidth, height=0.07\textheight]{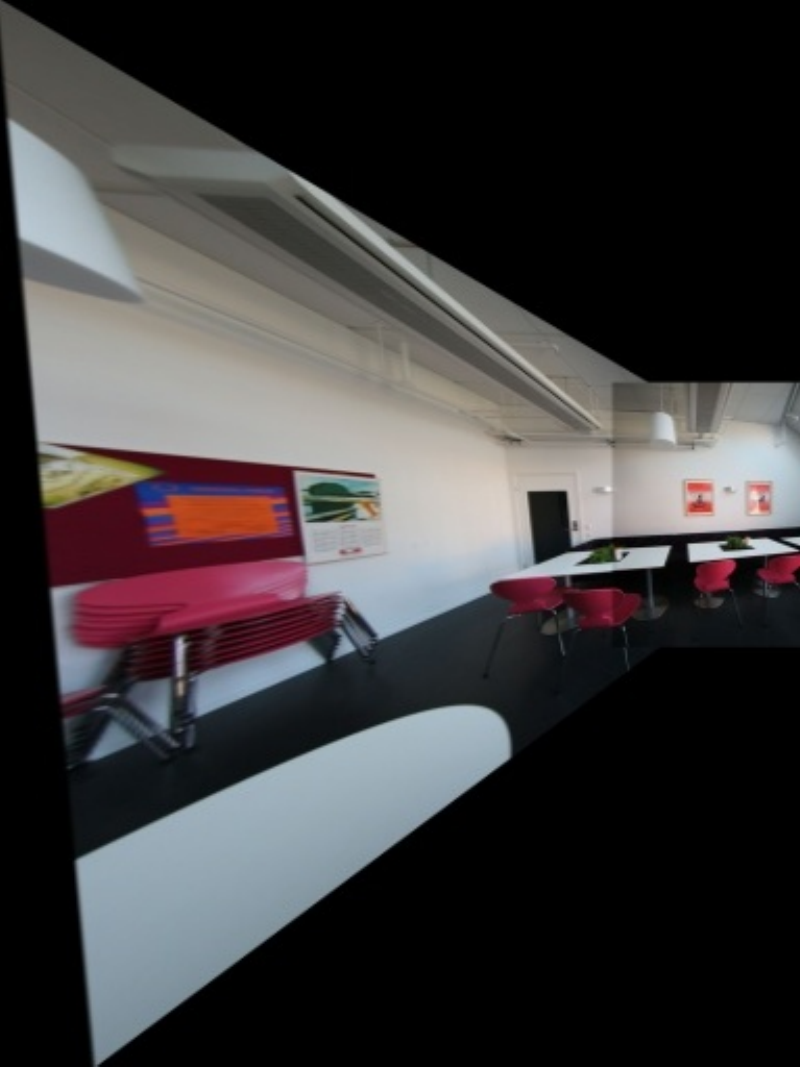} \\ (f-2) $\xi$ = 6, $t$ = 0.74s \\(\textcolor{blue}{Success}) } \\
         [-0.8em]
         \\
         \multirow{2}{*}{\makecell{\includegraphics[width=0.17\textwidth, height=0.075\textheight]{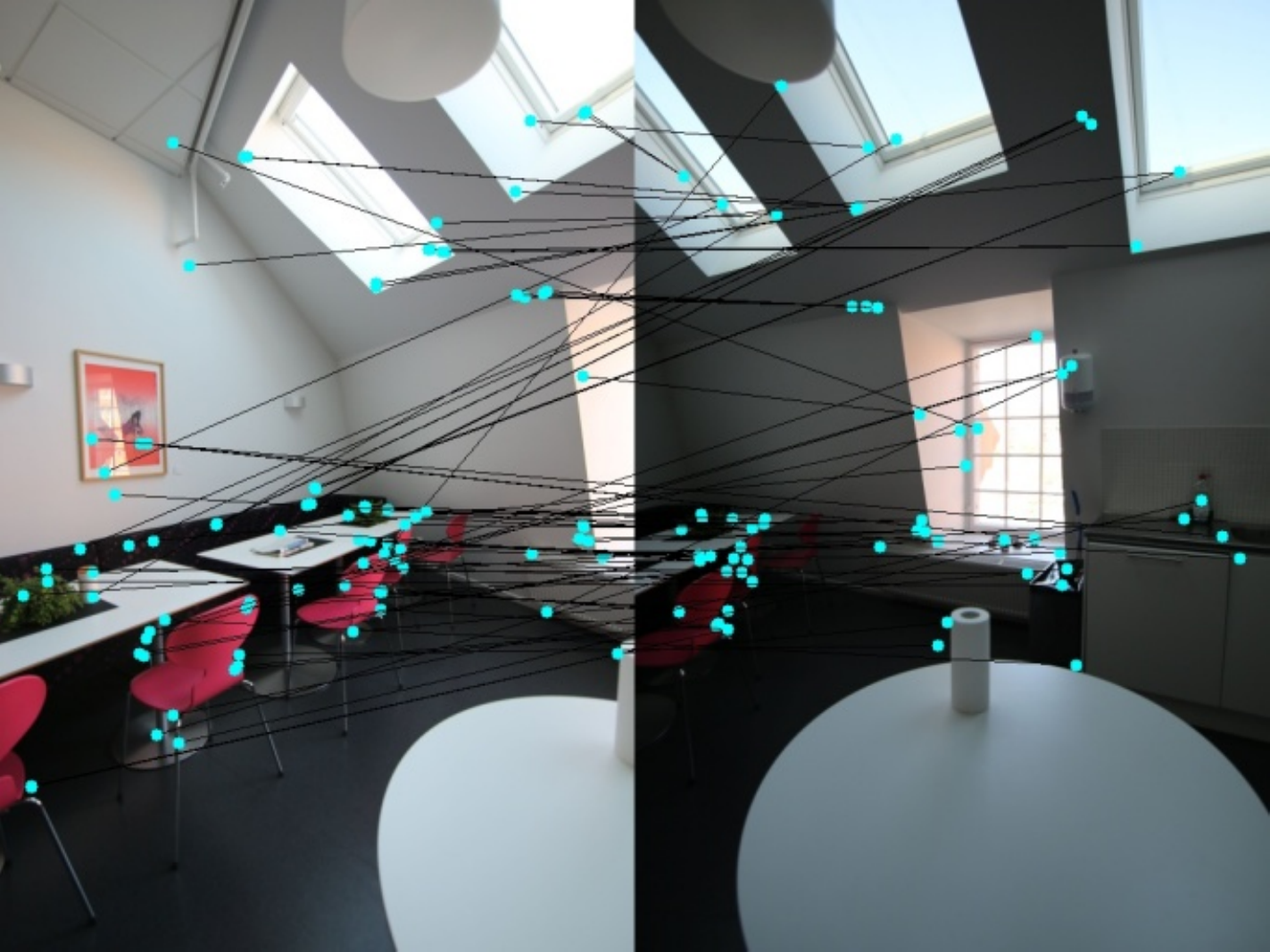} \\ (g) Match Pairs : 87}} & \makecell{\includegraphics[width=0.115\textwidth, height=0.07\textheight]{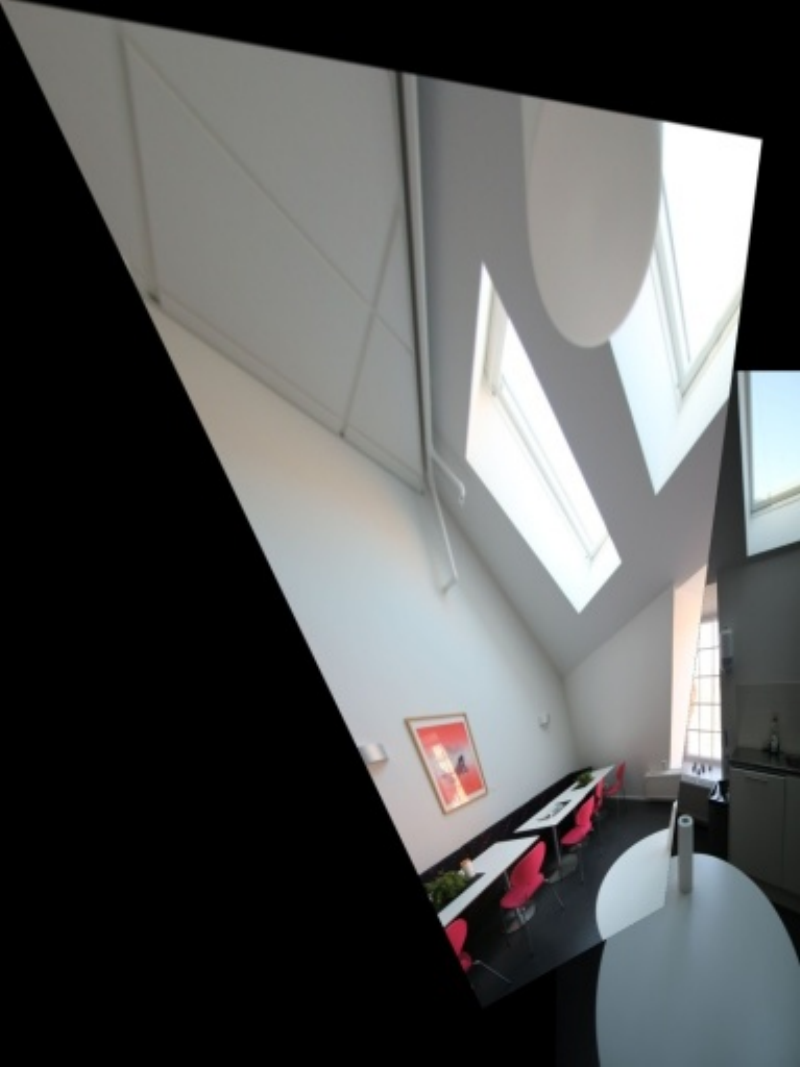} \\ (h-1) $\xi$ = 2, $t$ = 0.01s \\(\textcolor{red}{Fail}) \\ [0.5em]} & \makecell{\includegraphics[width=0.115\textwidth, height=0.07\textheight]{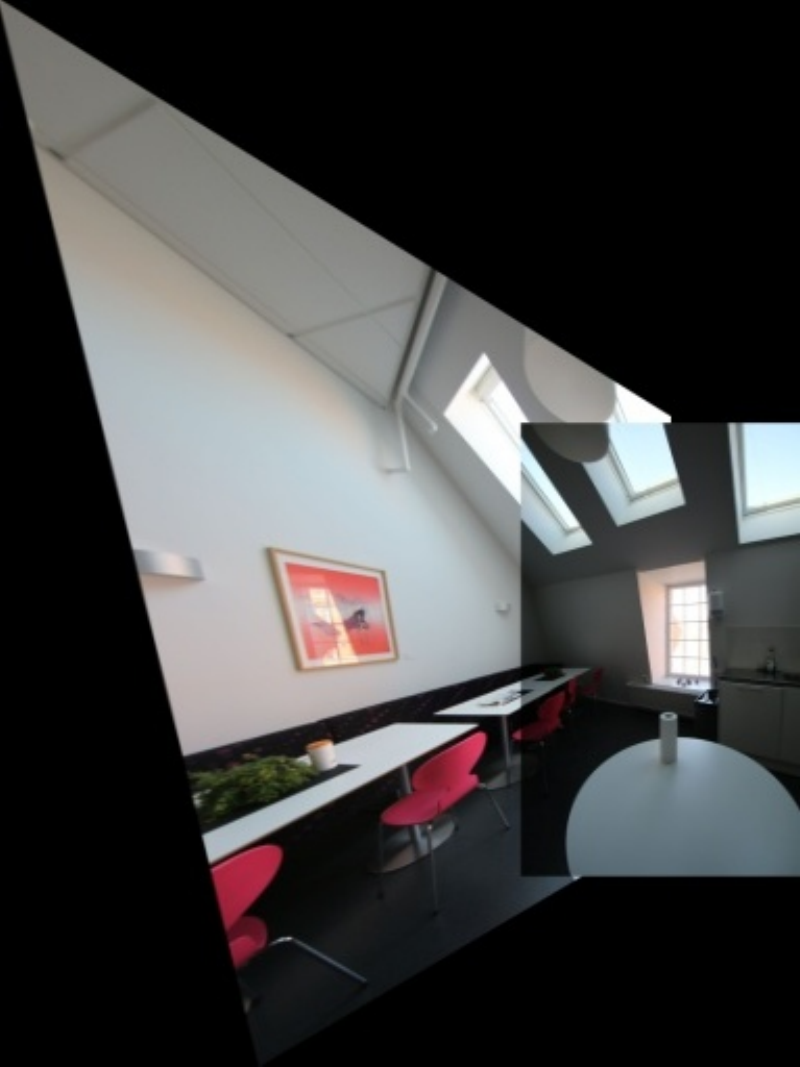} \\ (i-1) $\xi$ = 2, $t$ = 0.02s \\(\textcolor{blue}{Success}) \\ [0.5em]} & \makecell{\includegraphics[width=0.115\textwidth, height=0.07\textheight]{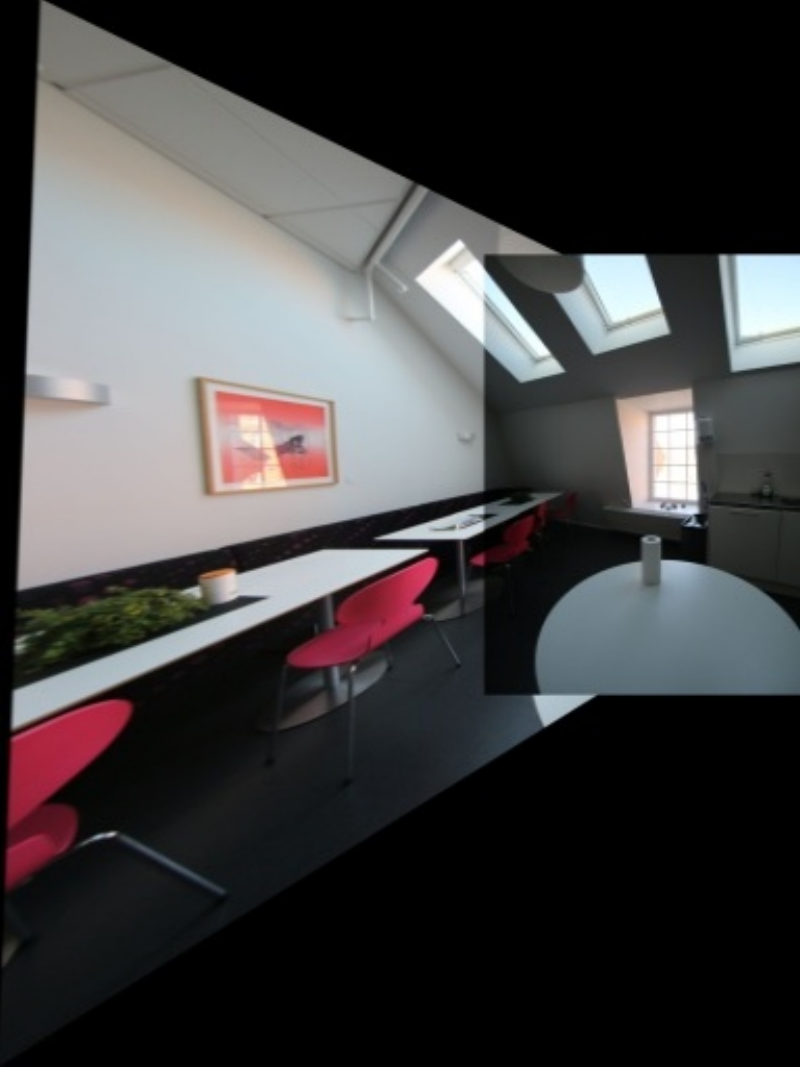}  \\ (j-1) $\xi$ = 2, $t$ = 2.94s \\(\textcolor{blue}{Success}) \\ [0.5em]} & \makecell{\includegraphics[width=0.115\textwidth, height=0.07\textheight]{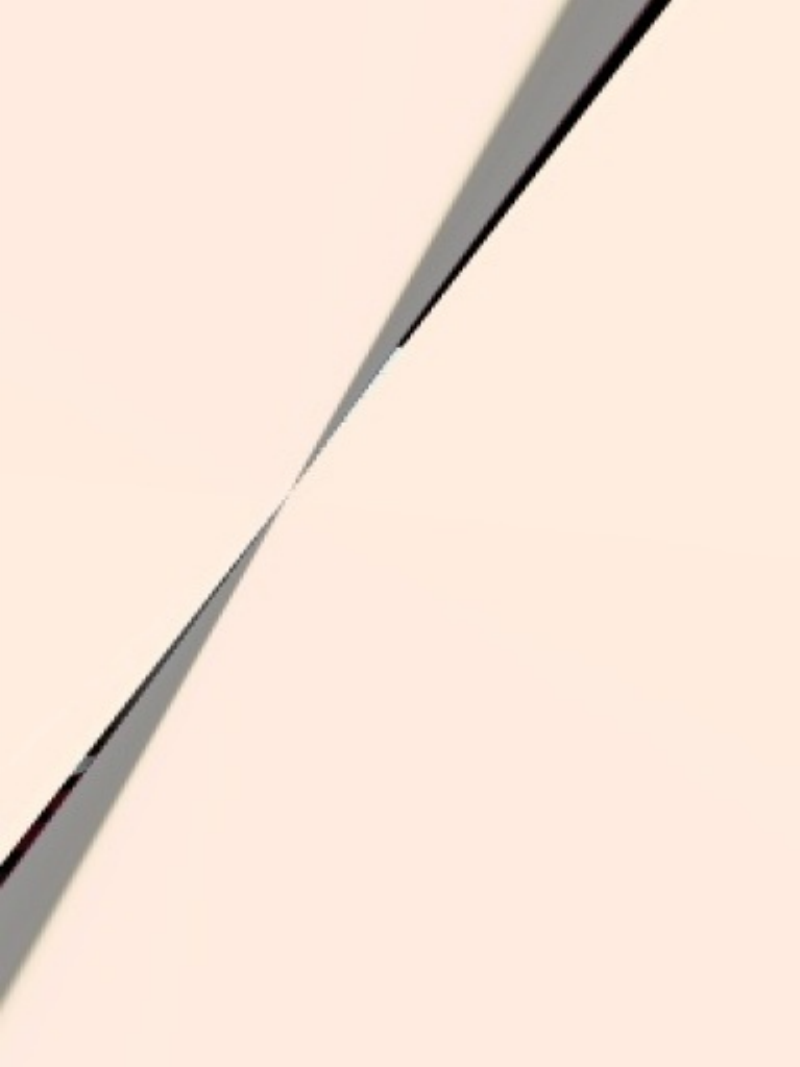} \\ (k-1) $\xi$ = 2, $t$ = 3.38s \\(\textcolor{red}{Fail}) \\ [0.5em]} & \makecell{\includegraphics[width=0.115\textwidth, height=0.07\textheight]{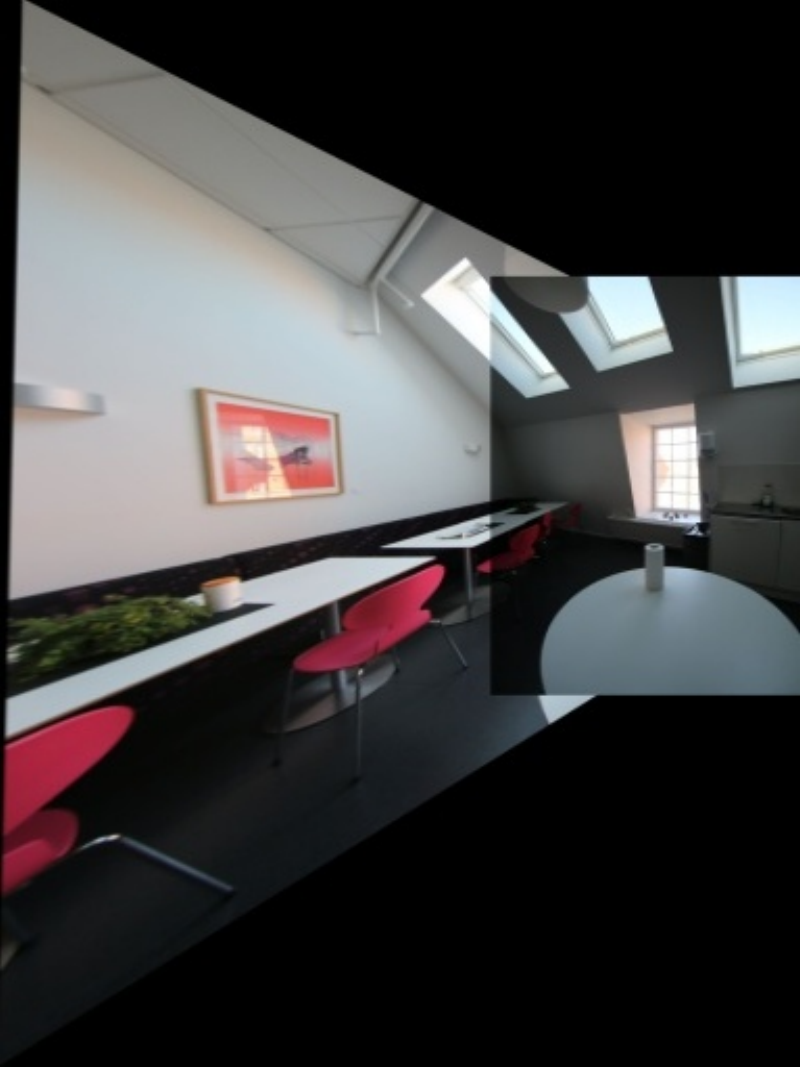} \\ (l-1) $\xi$ = 2, $t$ = 0.58s \\(\textcolor{blue}{Success}) \\ [0.5em]} 
         \\
         & \makecell{\includegraphics[width=0.115\textwidth, height=0.07\textheight]{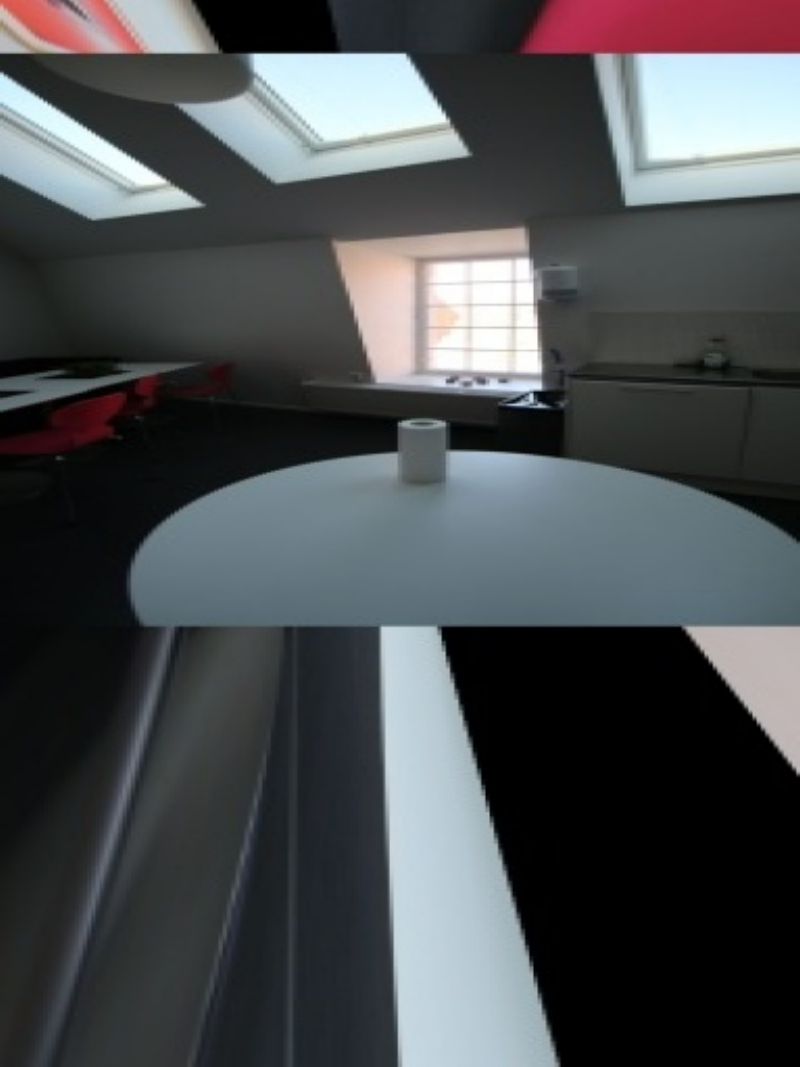} \\ (h-2) $\xi$ = 6, $t$ = 0.01s \\(\textcolor{red}{Fail})} & \makecell{\includegraphics[width=0.115\textwidth, height=0.07\textheight]{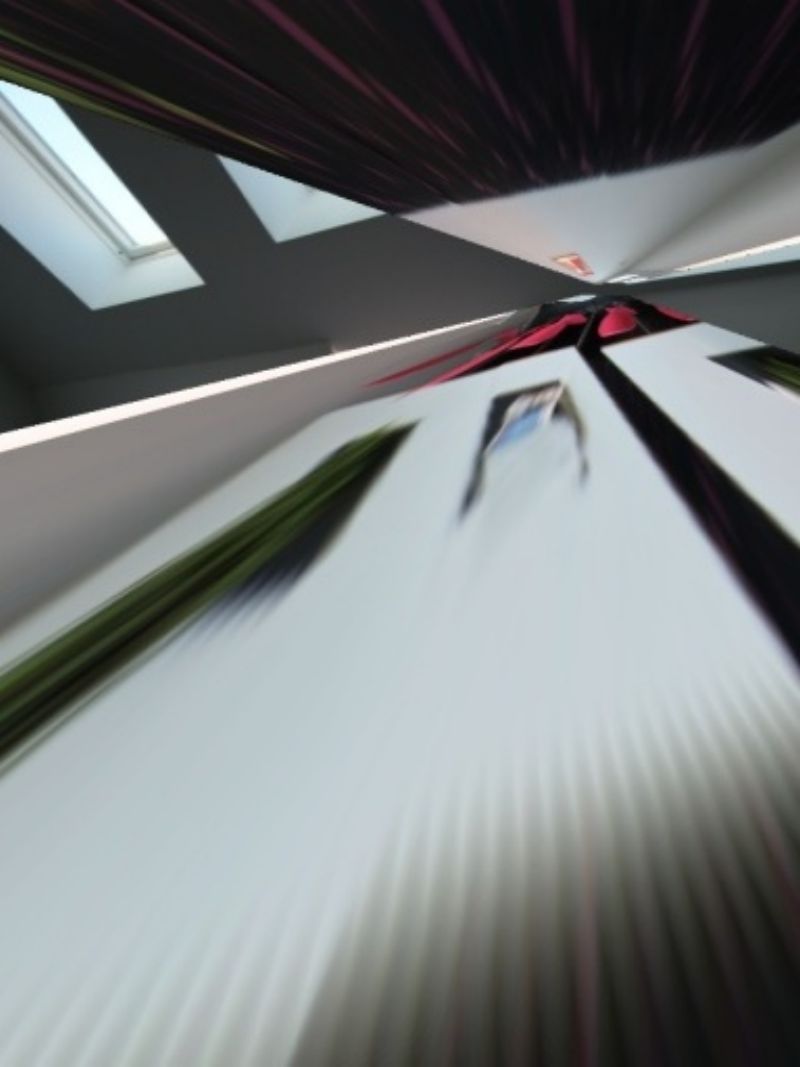} \\ (i-2) $\xi$ = 6, $t$ = 0.01s \\(\textcolor{red}{Fail})} & \makecell{\includegraphics[width=0.115\textwidth, height=0.07\textheight]{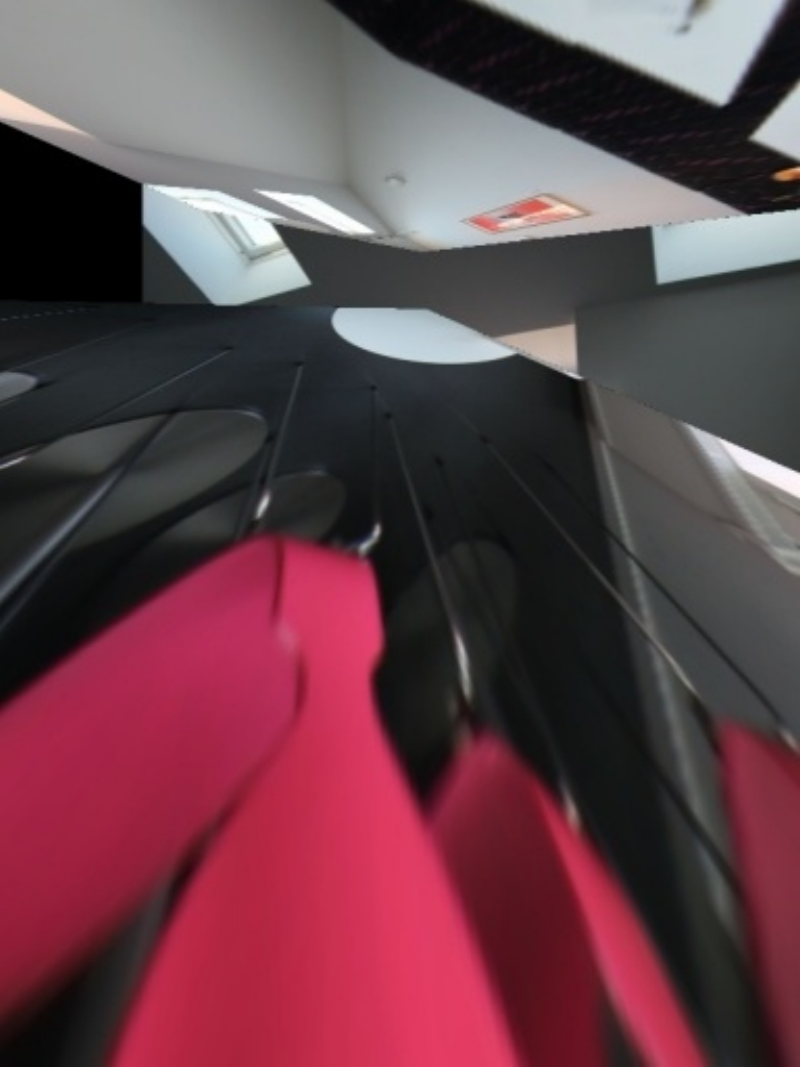}  \\ (j-2) $\xi$ = 6, $t$ = 5.11s \\(\textcolor{red}{Fail})} & \makecell{\includegraphics[width=0.115\textwidth, height=0.07\textheight]{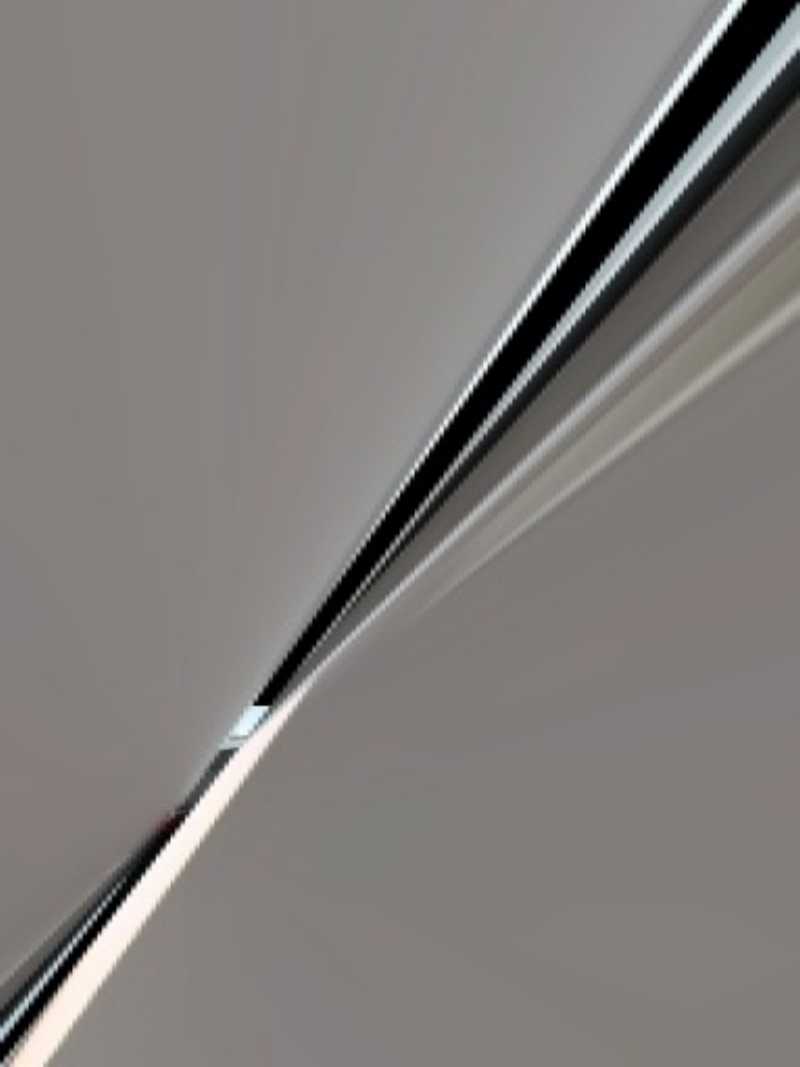} \\ (k-2) $\xi$ = 6, $t$ = 3.61s \\(\textcolor{red}{Fail})} & \makecell{\includegraphics[width=0.115\textwidth, height=0.07\textheight]{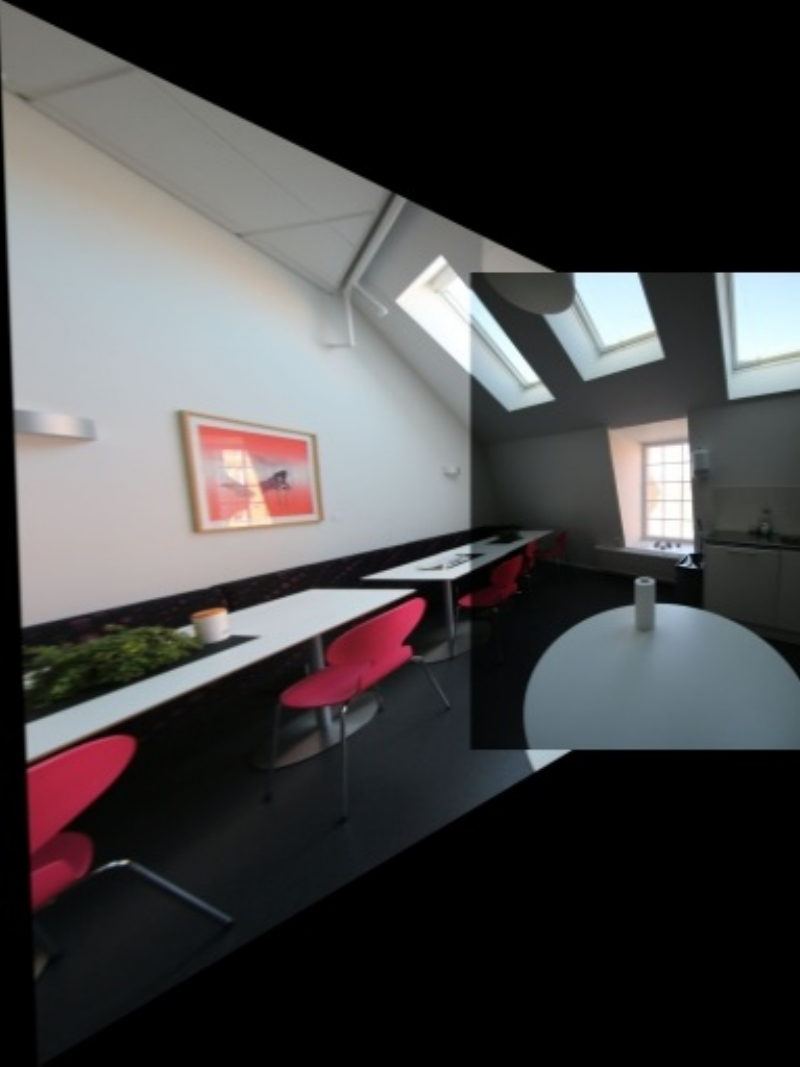} \\ (l-2) $\xi$ = 6, $t$ = 0.45s \\(\textcolor{blue}{Success})} \\
    \end{tabular}
    \vspace{-0.4em}
    \caption{Two representative visual cases of rotational homography estimation on the real-world PASSTA dataset~(c.f.~Section~\ref{sec:4dest-real}). The first column presents the input 2D-2D image matches and the other columns show the outputs of various methods. Only \textsf{GTM} succeeds in all presented cases.}
    \label{fig:app3_real_rep}
\end{figure*}

\noindent \textbf{Results}.
Table~\ref{tab:app3_real_eval} presents the success rates and timing results of various methods under different threshold setups.
Our \textsf{GTM} again achieves the highest accuracy across threshold variations, demonstrating the effectiveness of the proposed framework. 
The heuristic methods, \textsf{RANSAC} and \textsf{MLESAC}, while remarkably efficient, become unreliable at large misspecified thresholds; for example, the number of successful cases of RANSAC drops from 91 to 74 when $\xi$ increases from 2 pixels to 6 pixels. \textsf{ACM}, which is based on CM and BnB, achieves the second highest robustness, but it is on average 5 times slower than \textsf{GTM}.
Meanwhile, \textsf{DIRECT-4D} fails at all cases, as the objective function in \eqref{eq:TL-3} is not Lipschitz continuous, which is in line with the results of prior simulation experiments.
Fig.~\ref{fig:app3_real_rep} shows two visual cases where only \textsf{GTM} delivers fully successful estimation.

\section{Conclusion}
\label{sec:conc}
In this work, we present \textsf{GTM}, the first unified BnB framework for globally optimizing the truncated loss functions across diverse geometric tasks characterized by various residual functions. Behind \textsf{GTM} is an innovative hybrid solving strategy that involves $(n-1)$-dimensional BnB search and 1-dimensional Lipschitz continuous bounding functions that are general, tight, and can be efficiently solved by a global Lipschitz solver. We conduct extensive evaluation experiments and show that \textsf{GTM} achieves the state-of-the-art outlier-robustness and threshold-resilience while holding high efficiency across four different robust estimation applications. We look forward to extending the ideas in \textsf{GTM} to solve other robust M-estimation problems and more geometric estimation problems.

\ifCLASSOPTIONcompsoc
  \section*{Acknowledgments}
\else
  \section*{Acknowledgment}
\fi
This work is supported by the InnoHK initiative of the Innovation and Technology Commission of the Hong Kong Special Administrative Region Government via the Hong Kong Centre for Logistics Robotics.

\ifCLASSOPTIONcaptionsoff
  \newpage
\fi

\bibliographystyle{IEEEtran}
\bibliography{papers}

\appendices

\section{The General BnB Algorithm Pipeline}
In this section, we give a detailed introduction of the general BnB algorithm pipeline.
As presented in Algorithm~\ref{alg:BnB}, the method starts by initializing a priority queue $\mathcal{Q}$~(Line 1) with the first element being the initial branch $\mathcal{C}$ with its corresponding lower bound $L(\mathcal{C})$ as priority~(Lines 3-4). 
Then the search enters a while loop in which we maintain the smallest upper bound $U^*$ found so far and its related variable $\mathbf{v}^{*}$ that achieves this bound~(Lines 5-23). 
At each iteration, the sub-branch $\mathbb{B}$ with the smallest lower bound~(the highest priority) is examines: if the lower bound $L(\mathbb{B})$ and $U^*$ differ by less than $\epsilon$, then $\mathbf{v}^*$ already achieves the minimum value up to error $\epsilon$ and the search terminates~(Lines 7-9); otherwise, the current sub-branch $\mathbb{B}$ is further divided and each sub-branch $\mathbb{B}_i$ is further checked. Particularly, $\mathbb{B}_i$ with $L(\mathbb{B}_i)$ higher than $U^*$ cannot contain a better solution and would be eliminated~(Lines 12-15). 
Otherwise, the upper bound $U(\mathbb{B}_i)$ is computed and the $(U^*, \mathbf{v}^*)$ would be updated if $U(\mathbb{B}_i) < U^*$, meanwhile $\mathbb{B}_i$ is stored into $\mathcal{Q}$~(Lines 16-20).

\section{Proof of Proposition~\ref{prop:Lips_inher}}
\label{sec:proof_prop1}
In this section, we give the proof of Proposition~\ref{prop:Lips_inher} in the main manuscript.

Consider the upper bounding function in \eqref{eq:GTM_UB}: 
\begin{align}
f(v_1, \dot{\mathbf{v}}_{2:n}) &= \sum_{i=1}^M \min\{r_i(v_1, \dot{\mathbf{v}}_{2:n}), \xi\},
\end{align}
where $r_i$ is defined as in \eqref{eq:general-r_i}. 
Since $g_i(\dot{\mathbf{v}}_{2:n})$ is constant and the norm $\|h_i(v_1) - g_i(\dot{\mathbf{v}}_{2:n})\|$ preserves Lipschitz continuity,
each residual $r_i(v_1, \dot{\mathbf{v}}_{2:n})$ is Lipschitz continuous, say with constant $K_i$.
Then, by inspection we have
\begin{align}\label{eq:min_pro-1}
    |\min\{z_1, z_3\} - \min\{z_2, z_3\}| \leq |z_1 - z_2|,
\end{align}
for all $z_1, z_2, z_3 \in \mathbb{R}$. Thus, for any $v_1', v_1'' \in \mathcal{C}_1$, we have:
\begin{align}
     & \min\{r_i(v'_1, \dot{\mathbf{v}}_{2:n}),\ \xi\} - \min\{r_i(v''_1, \dot{\mathbf{v}}_{2:n}),\ \xi\}| \\
        \leq\ &  | r_i(v'_1, \dot{\mathbf{v}}_{2:n}) - r_i(v''_1, \dot{\mathbf{v}}_{2:n}) | \leq\ K_i\cdot|v_1' - v_1''|.
\end{align}


Thus, the residual function $\min\{r_i(v_1, \dot{\mathbf{v}}_{2:n}), \xi\}$ is Lipschitz continuous, and as a finite sum of the residuals, $f(v_1, \dot{\mathbf{v}}_{2:n})$ is  Lipschitz continuous as well.

Consider the lower bounding function in \eqref{eq:GTM_LB}:
\begin{align}
    \myunderline{f}(v_1) = \sum_{i=1}^M\myunderline{f}_i(v_1) = \sum_{i=1}^M \min\{\myunderline{r}_i(v_1), \xi\},
\end{align}
where $\myunderline{r}_i(v_1)$ is the norm of a vector-valued function with each entry $\myunderline{r}_i^{(j)}(v_1)$ defined in \eqref{eq:general-r_ij}. By Lipschitz continuity of $h_i(v_1)$ and the fast that $\myunderline{r}_i^{(j)}(v_1)$ defined in \eqref{eq:general-r_ij} is continuous and a piece-wise function, we conclude that  $\myunderline{r}_i^{(j)}(v_1)$ is also Lipschitz continuous, and so does the norm $\myunderline{r}_i(v_1)=\| \myunderline{r}_i^{(1)}(v_1),\dots, \myunderline{r}_i^{(W)}(v_1) \|$. 
Then, we can again apply \eqref{eq:min_pro-1} to show that each $\myunderline{f}_i$ and furthermore their sum $\myunderline{f}$ are Lipschitz continuous. The proof is complete.

\begin{algorithm}[!t]
\footnotesize
\caption{General Branch-and-bound Algorithm.}
\label{alg:BnB}
\textbf{Input:} Objective function $f(\cdot)$, prescribed error $\epsilon$;

\textbf{Output:} A global minimizer $\mathbf{v}^{*}$;

    \begin{algorithmic}[1]
        \STATE $\mathcal{Q} \gets $ An empty priority queue;
        \STATE $U^{*},\ \mathbf{v}^{*} \gets$ getUpperBound($f(\cdot)$, $\mathcal{C}$);
        \STATE $L(\mathcal{C}) \gets $ getLowerBound($f(\cdot),\ \mathcal{C})$;
        \STATE Insert $\mathcal{C}$ with priority $L(\mathcal{C})$ into $\mathcal{Q}$;
        \WHILE{$\mathcal{Q}$ is not empty}
            \STATE Pop a branch $\mathbb{B}$ with the lowest $L(\mathbb{B})$ from $\mathcal{Q}$;
            \IF{$U^*$ - $L(\mathbb{B}) < \epsilon$}
            \STATE \textbf{end while}
            \ENDIF
            \STATE Divide $\mathbb{B}$ into $2^n$ sub-branches.
            \FOR{each sub-branch $\mathbb{B}_i$}
		\STATE $L(\mathbb{B}_i) \gets$ getLowerBound($f(\cdot),\ \mathbb{B}_i)$;
            \IF{$L(\mathbb{B}_i) > U^*$}
            \STATE \textbf{continue}
            \ELSE
            \STATE $U(\mathbb{B}_i),\ \mathbf{v}_i \gets$ getUpperBound($f(\cdot),\ \mathbb{B}_i)$;
            \IF{$U(\mathbb{B}_i) < U^{*}$}
            \STATE $\mathbf{v}^* \gets \mathbf{v}_i$;\ $U^* \gets U(\mathbb{B}_i)$;
            \ENDIF
            \STATE Insert $\mathbb{B}_i$ with priority $L(\mathbb{B}_i)$ into $\mathcal{Q}$;
            \ENDIF
 		\ENDFOR 
        \ENDWHILE
    \end{algorithmic}
\end{algorithm}

\section{Range of $g_i(\theta_2)$ for Computing $L_{\textsf{GTM}-1}$}
\label{app:app1_range}
Recall that in Section~\ref{sec:2d_bound} of the main manuscript, we need to compute the range of $g_i(\theta_2) = A_{2i}\sin(\theta_2+\phi_{2i})$ with $\theta_2 \in [\theta_2^l, \theta_2^u]$. Here we provide the computation details.

Without loss of generality, we suppose that $A_{2i} \geq 0$. Denote the range of $g_i(\theta_2)$ by $[{s_i^l}, {s_i^u}]$, then four cases are involved to compute the range:
\begin{enumerate}
    \item[$\bullet$] If $\theta_2^u + \phi_{2i} \leq -\frac{\pi}{2}$ or $\frac{\pi}{2} \leq \theta_2^l + \phi_{2i} \leq \theta_2^u + \phi_{2i} \leq \frac{3\pi}{2}$: 
    \begin{align}
        {s_i^l} &= A_{2i}\sin(\theta_2^u + \phi_{2i}), \\
        {s_i^u} &= A_{2i}\sin(\theta_2^l + \phi_{2i});
    \end{align}
    \item[$\bullet$] If $-\frac{\pi}{2} \leq \theta_2^l + \phi_{2i} \leq \theta_2^u + \phi_{2i} \leq \frac{\pi}{2}$ or $\theta_2^l + \phi_{2i} \geq \frac{3\pi}{2}$:
    \begin{align}
        {s_i^l} &= A_{2i}\sin(\theta_2^l + \phi_{2i}), \\
        {s_i^u} &= A_{2i}\sin(\theta_2^u + \phi_{2i});
    \end{align}
    \item[$\bullet$] If $\theta_2^l + \phi_{2i} \leq -\frac{\pi}{2} \leq \theta_2^u + \phi_{2i}$ or $\theta_2^l + \phi_{2i} \leq \frac{3\pi}{2} \leq \theta_2^u + \phi_{2i}$:
    \begin{align}
        {s_i^l} &= -A_{2i}, \\
        {s_i^u} &= \max\{A_{2i}\sin(\theta_2^l + \phi_{2i}), A_{2i}\sin(\theta_2^u + \phi_{2i})\};
    \end{align}
    \item[$\bullet$] If $\theta_2^l + \phi_{2i} \leq \frac{\pi}{2} \leq \theta_2^u + \phi_{2i}$:
    \begin{align}
        {s_i^l} &= \min\{A_{2i}\sin(\theta_2^l + \phi_{2i}), A_{2i}\sin(\theta_2^u + \phi_{2i})\}, \\
        {s_i^u} &= A_{2i}.
    \end{align}
\end{enumerate}
Therefore we can compute the range $[{s_i^l}, {s_i^u}]$ of $g_i(\theta_2)$ related to each sample data with constant time. 

\section{Range of $g_i(t_2, t_3)$ for Computing $L_{\textsf{GTM}-2}$}
\label{app:app2_range}
Recall that in Section~\ref{sec:3d_bound} of the main manuscript, we need to compute the range of ${g_i}(t_2, t_3) = (t_2 + p_{i2})^2 + (t_3 + p_{i3})^2 - \|\mathbf{q}_i\|_2^2$ with $t_2\in[t_2^l,t_2^u]$ and $t_3\in[t_3^l, t_3^u]$. Clearly, the range depends on the two square summands, i.e., $(t_2 + p_{i2})^2$ and $(t_3 + p_{i3})^2$. We can compute the range of $(t_2 + p_{i2})^2$ as follows:
\begin{align}\label{eq:range_2g1}
    (t_2 + p_{i2})^2 \in \begin{cases}
        [(t_2^u + p_{i2})^2,\ (t_2^l + p_{i2})^2],\ \textnormal{if}\ t_2^u \leq -p_{i2}; \\
        [0,\ \max\{\substack{(t_2^u + p_{i2})^2,\\ (t_2^l + p_{i2})^2}\}],\ \textnormal{if}\ t_2^l \leq -p_{i2} \leq t_2^u; \\
        [(t_2^l + p_{i2})^2,\ (t_2^u + p_{i2})^2],\ \textnormal{if}\ t_2^l \geq -p_{i2}; \\
    \end{cases}
\end{align}
Similarly, the range of $(t_3 + p_{i3})^2$ can be also computed as \eqref{eq:range_2g1}. Finally, we can get the range of $g_i(t_2, t_3)$ by simply applying the additive interval arithmetic operation, i.e., \eqref{eq:arith-add}.

\section{More Implementation Details}
We implement the proposed method based on C++. For the other existing methods, we use the provided implementations.
The confidence of \textsf{RANSAC} and \textsf{MLESAC} in all experiments is set to $99.99\%$, so as to ensure enough accuracy.
To assess the threshold robustness of different methods, we need to set multiple thresholds. For each geometric problem, we first determine an approximate threshold range based on its measurement scale. Subsequently, evaluation thres
holds are sampled in descending order across this range.
All experiments are evaluated on a laptop equipped with an Intel Core i7-10875H CPU@2.3 GHz and 32GB of RAM.

\begin{table}[!t]
\centering
    \caption{Detailed information of the processed ETH dataset~\cite{Theiler-ISPRS2014}. $M$ in the table represent the number of correspondences.}
    \vspace{-0.6em}
    \label{tab:eth_info}
    \footnotesize
    \renewcommand{\tabcolsep}{4.2pt} 
    \renewcommand\arraystretch{1.4}
    \begin{tabular}{c|ccccc}
        \Xhline{1pt}
        Scene & Arch & Courtyard & Facade & Office & Trees \\
        \Xhline{0.5pt}
        Overlap & 30$\%$-40$\%$ & 40$\%$-70$\%$ & 60$\%$-70$\%$ & $>$80$\%$ & $\approx$50$\%$ \\
        $\#$ of Pairs & 10 & 28 & 21 & 10 & 15\\
        Average of $M$ & 8892 & 9877 & 1356 & 912 & 14987\\
        \makecell{Average of \\ Outlier Ratio} & 99.44$\%$ & 96.23$\%$ & 95.86$\%$ & 97.92$\%$ & 99.70$\%$\\
        \Xhline{1pt} 
    \end{tabular}
\end{table}

\section{Details of the ETH Dataset}
The ETH dataset~\cite{Theiler-ISPRS2014} is a challenging large-scale point cloud dataset consisting of five distinct scenes: arch, courtyard, facade, office, and trees.
Recall that in Section~\ref{sec:3dest-real} of the main manuscript, we employ the ISS~\cite{Zhong-ICCVW2009} and FPFH~\cite{Rusu-ICRA2009} descriptors to extract the 3D-3D correspondences. The details of the processed data is presented here in Table~\ref{tab:eth_info}.

\end{document}